\documentclass[sigconf,nonacm]{acmart}
 
\AtBeginDocument{%
  }
\setcopyright{none}
\acmConference[KDD Workshop on Agentic AI '26]{KDD Workshop on Evaluation
  and Trustworthiness of Agentic AI}{August 9--13, 2026}{Jeju, Republic of Korea}
\acmISBN{}
\acmDOI{}

\usepackage{booktabs}
\usepackage{multirow}
\usepackage{array}
\usepackage{graphicx}
\usepackage{amsmath}
\usepackage[table]{xcolor}
\usepackage{longtable}
\usepackage{subcaption}
\usepackage{makecell}
\usepackage{enumitem}
\usepackage{url}
\usepackage{algorithm}
\usepackage{algpseudocode}
\usepackage[capitalize]{cleveref}
\crefname{figure}{Figure}{Figures}
\crefname{subfigure}{Figure}{Figures}

\makeatletter
\AtBeginDocument{%
  \fancypagestyle{standardpagestyle}{%
    \fancyhf{}%
    \renewcommand{\headrulewidth}{\z@}%
    \renewcommand{\footrulewidth}{\z@}%
    \fancyfoot[C]{\if@ACM@printfolios\footnotesize\thepage\fi}%
    \fancyhead[LO]{\ACM@linecountL\@headfootfont\shorttitle}%
    \fancyhead[RE]{\@headfootfont\@shortauthors\ACM@linecountR}%
    \fancyhead[LE]{\ACM@linecountL\@headfootfont\footnotesize
      \acmConference@shortname,
      \acmConference@date, \acmConference@venue}%
    \fancyhead[RO]{\@headfootfont
      \acmConference@shortname,
      \acmConference@date, \acmConference@venue\ACM@linecountR}%
  }%
  \pagestyle{standardpagestyle}%
}
\makeatother
\usepackage{comment}

\usepackage[most]{tcolorbox}
\tcbuselibrary{listings, breakable, skins}
\tcbset{promptbox/.style={
    width=\linewidth,
    colback=gray!5,
    colframe=black,
    breakable,
    enhanced,
    pad at break*=2mm,
    before skip=\medskipamount,
    after skip=\medskipamount,
    listing only,
    listing options={
        basicstyle=\footnotesize\ttfamily,
        breaklines=true,
        breakatwhitespace=false,
        columns=fullflexible,
        keepspaces=true,
        aboveskip=0pt,
        belowskip=0pt
    },
    boxsep=1mm,
    left=1mm,
    right=1mm,
    top=1mm,
    bottom=1mm
}}
\usepackage{adjustbox}
\usepackage{rotating}

\usepackage{listings}
\usepackage{pifont} 

\let\origtexttt\texttt
\makeatletter
\begingroup
  \catcode`\_=\active \catcode`\.=\active \catcode`\/=\active
  \gdef\BT@activate{%
    \catcode`\_=\active \catcode`\.=\active \catcode`\/=\active
    \def_{\char`\_\allowbreak}%
    \def.{\char`\.\allowbreak}%
    \def/{\char`\/\allowbreak}%
    \def\_{\char`\_\allowbreak}%
  }
\endgroup
\protected\def\texttt{%
  \begingroup
  \BT@activate
  \BT@texttt}
\newcommand\BT@texttt[1]{\origtexttt{#1}\endgroup}
\makeatother

\newcommand{\msg}[1]{}
\newcommand{\sketch}[1]{}

\newlength{\reservedtmp}

\newlength{\phfigtmp}

\DeclareRobustCommand{\ours}{\textsc{PCBWorld}}
\DeclareRobustCommand{\oursbench}{\textsc{PCBWorld-Bench}}
\DeclareRobustCommand{\kicad}{\textsc{KiCad}}

\title{\ours: A Benchmark Environment\texorpdfstring{\\}{ }for Engine-Grounded PCB Design Automation}

\author{Hyungseok Song}
\authornote{Equal contribution.\hspace{1.5em}$^{\dagger}$Corresponding author.}
\affiliation{%
  \institution{LG AI Research}
  \city{Seoul}
  \country{Republic of Korea}
}
\email{hyungseok.song@lgresearch.ai}

\author{Junseok Park}
\authornotemark[1]
\affiliation{%
  \institution{LG AI Research}
  \city{Seoul}
  \country{Republic of Korea}
}
\email{frank.park227@lgresearch.ai}

\author{Won-Seok Choi}
\authornotemark[1]
\affiliation{%
  \institution{LG AI Research}
  \city{Seoul}
  \country{Republic of Korea}
}
\email{wonseok.choi@lgresearch.ai}

\author{Seohui Bae}
\affiliation{%
  \institution{LG AI Research}
  \city{Seoul}
  \country{Republic of Korea}
}
\email{seohui.bae@lgresearch.ai}

\author{Han-Seul Jeong}
\affiliation{%
  \institution{LG AI Research}
  \city{Seoul}
  \country{Republic of Korea}
}
\email{hanseul.jeong@lgresearch.ai}

\author{Youngjoon Park}
\authornotemark[2]
\affiliation{%
  \institution{LG AI Research}
  \city{Seoul}
  \country{Republic of Korea}
}
\email{yj.park@lgresearch.ai}

\author{Soonyoung Lee}
\authornotemark[2]
\affiliation{%
  \institution{LG AI Research}
  \city{Seoul}
  \country{Republic of Korea}
}
\email{soonyoung.lee@lgresearch.ai}

\renewcommand\shortauthors{Song et al.}

\renewcommand\footnotetextcopyrightpermission[1]{}

\begin{document}

\begin{abstract}
{
PCB routing is the task of connecting the nets of a board with copper traces under strict design rules, yet learning-based methods still lag behind rule-based routers.
We introduce \ours{}, an open-source engine-grounded PCB routing environment built on \kicad{}, an electronic design automation (EDA) engine.
As a human engineer does, agents in \ours{} interactively route a board through the engine's native operations, guided by its Design Rule Check (DRC) feedback.
The environment supports both RL and tool-using LLM agents.
Alongside the environment, \oursbench{} provides three board datasets in the native \texttt{.kicad\_pcb} format, two controllable synthetic generators and 679 real open-source boards.
It scores any completed board with eight engine-checked evaluation metrics, regardless of the routing method.
In our experiments, agents in \ours{} consistently outperformed grid-action RL policies and open-loop LLM baselines, and an RL policy trained only on synthetic boards transferred zero-shot to real boards, approaching rule-based routers.
\ours{} and \oursbench{} are available at \url{https://github.com/LGAI-Research/PCBWorld}.
}
\end{abstract}

\maketitle

\section{Introduction}\label{sec:intro}

{
Printed circuit boards (PCBs) are the physical boards that mount and interconnect electronic components, forming the backbone of nearly every electronic product~\citep{ipc2221a,khandpur2005printed,coombs2016printed}.
Turning a circuit design into a manufacturable board centers on \emph{routing}, the task of drawing copper traces that connect the pads of each net while keeping distinct nets electrically isolated under strict design rules~\citep{sherwani1999vlsi}.
Yet industry still relies mainly on rule-based routers rooted in decades-old heuristics~\citep{lee1961maze,linsker1984ripup,freerouting}, which automate much of the routing process but rarely complete complex production boards end-to-end.
Learning-based methods have been studied extensively to overcome this limit~\citep{jumanjibench,sable}, but unlike in language, vision, and games, they are not yet competitive even with rule-based routers.
}

{
We attribute this limitation to how the routing problem is modeled.
Existing RL formulations either cast routing as cell-by-cell movement on a grid, where the search space grows rapidly with grid size~\citep{jumanjibench,sable}, or delegate routing to an off-the-shelf router, steering it through auxiliary decisions such as net ordering~\citep{wang2020attention,lin2023xroute,netordering2025transformer}.
Recent LLM-based methods generate electronic design automation (EDA) scripts~\citep{he2024chateda}, Verilog modules~\citep{thakur2024autochip}, or schematics~\citep{pcbschemagen2026llm} as whole artifacts, revising them at most on per-artifact verdicts. None targets board routing, where the strict design rules bind.
The workflow that reliably completes production boards remains that of a human engineer, who neither hands the board to an auto-router nor draws every trace by hand.
Instead, the engineer repeatedly reads the current board, judges where the next connection can feasibly run, and draws it with the engine's native operations, which keep the trace within the design rules.
}

\begin{figure}[t]
  \centering

  \includegraphics[width=0.92\linewidth]{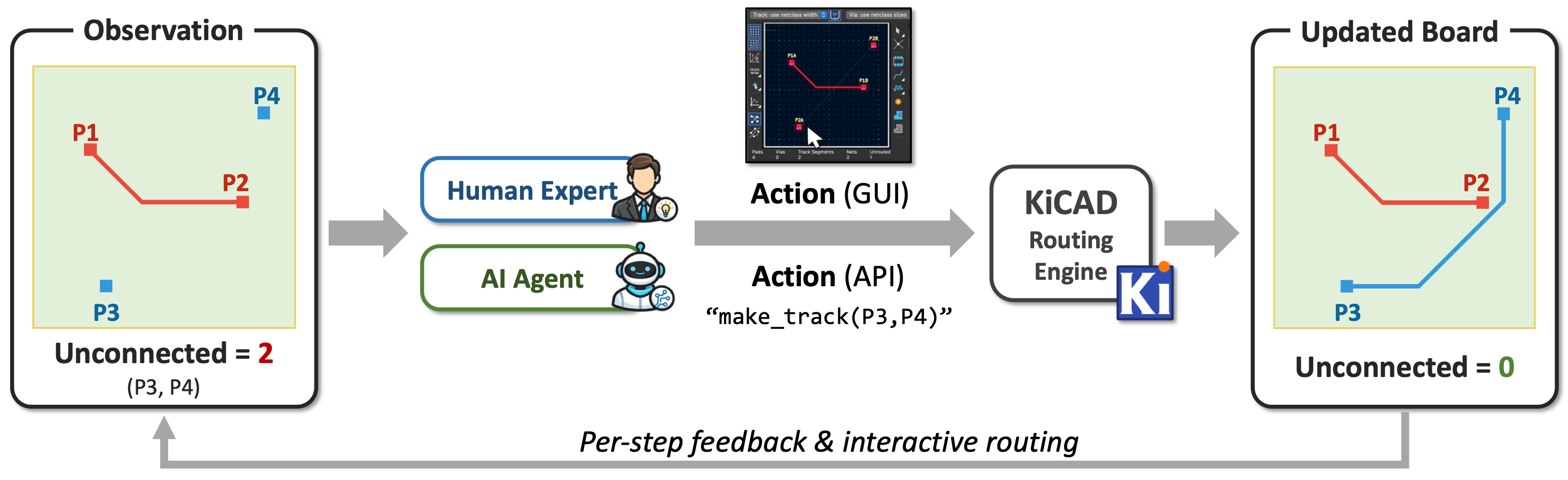}

  \footnotesize
  \setlength{\tabcolsep}{4pt}
  \renewcommand{\arraystretch}{0.95}
  \begin{adjustbox}{max width=\linewidth}
  \begin{tabular}{lcccccc}
  \toprule
   & \textbf{Task} & \textbf{Geom.} & \textbf{DRC} & \textbf{API} & \textbf{RL} & \textbf{LLM} \\
  \midrule

  \ours{} (ours)                        & PCB      & Gridless & \checkmark & \checkmark & \checkmark & \checkmark \\
  TREND~\citep{netordering2025transformer} & Chip     & Grid     & \checkmark & --         & \checkmark & --         \\
  XRoute-Env~\citep{lin2023xroute}         & Chip     & Grid     & \checkmark & \checkmark & \checkmark & --         \\
  Jumanji~\citep{jumanjibench}             & Abstract & Grid     & --         & --         & \checkmark & --         \\
  PCB-Bench~\citep{pcbbench2026placement}  & PCB      & --       & --         & --         & --         & \checkmark \\

  \bottomrule
  \end{tabular}
  \end{adjustbox}

  \caption{
\textbf{Overview of \ours{} and \oursbench{}.} Agents invoke the \kicad{} engine's native API and condition each step on engine-computed feedback.
  }
  \label{fig:intro-overview}
  \Description{Overview of \ours{} and \oursbench{}. Agents invoke the \kicad{} engine's native API and condition each step on engine-computed feedback, and rewards and metrics are computed from the same engine-checked outcomes.}
\end{figure}

{
To place agents in this same loop, we introduce \ours{}, an \emph{engine-grounded} PCB environment in which agents complete boards by directly invoking an EDA engine's native routing operations.
Built on the open-source \kicad{} EDA engine~\citep{linuxfoundation_kicad,kicad}, \ours{} exposes 58 Python APIs over the engine, from step-level routing operations to its Design Rule Check (DRC).
The agent routes the board itself in a closed loop, receiving the updated board state and DRC feedback after every operation.
On top of a standard Gym interface, \ours{} provides tailored wrappers for RL policies and tool-using LLM agents, so both families train and act in the same environment.
Unlike GUI-driven agent environments~\citep{xie2024osworld}, \ours{} also runs headless and vectorized, sustaining the throughput that large-scale RL training requires.
}

{
Alongside \ours{}, we release \oursbench{}, a benchmark that evaluates \emph{geometric feasibility reasoning}, the capability to produce a routed board in which every net is connected and every trace satisfies the strict design rules.
This capability remains largely untested in existing LLM benchmarks, which focus on software engineering~\citep{swebench}, board interpretation~\citep{pcbbench2026placement}, or 3D-CAD generation~\citep{wu2021deepcad,text2cadbench2026}.
\oursbench{} combines two synthetic board generators, one grid-based (D1) and one gridless (D2), with a curated set of 679 real open-source boards (D3).
Every instance is provided in \kicad{}'s native board format (\texttt{.kicad\_pcb}) with its own design rules, and loads directly into \ours{} for training and evaluation.
The evaluation is method-agnostic, scoring any completed board with the same engine whether or not the method routes through \ours{}.
}
 
{
On \oursbench{}, we evaluated three families of routing methods under the same protocol: rule-based routers, RL agents, and LLM agents.
Three findings emerge.
(i) Agents acting through \ours{} outperform their counterparts, with PPO maintaining performance at grid resolutions where grid-action RL collapses and interactive LLM routing beating the open-loop baselines.
(ii) Trained only on synthetic boards, PPO transfers zero-shot to real boards, outperforming all baselines on small boards while trailing the strongest rule-based router on larger ones.
(iii) Large-board routing remains an open challenge for both RL and LLM agents.
We release \ours{} and \oursbench{} as open source, as a shared foundation for evaluating and advancing learning-based PCB routing.
}


\subsection{PCB Routing Problem}\label{sec:bg-formal}
\begin{figure}[t]
  \centering
  \includegraphics[width=0.7\linewidth]{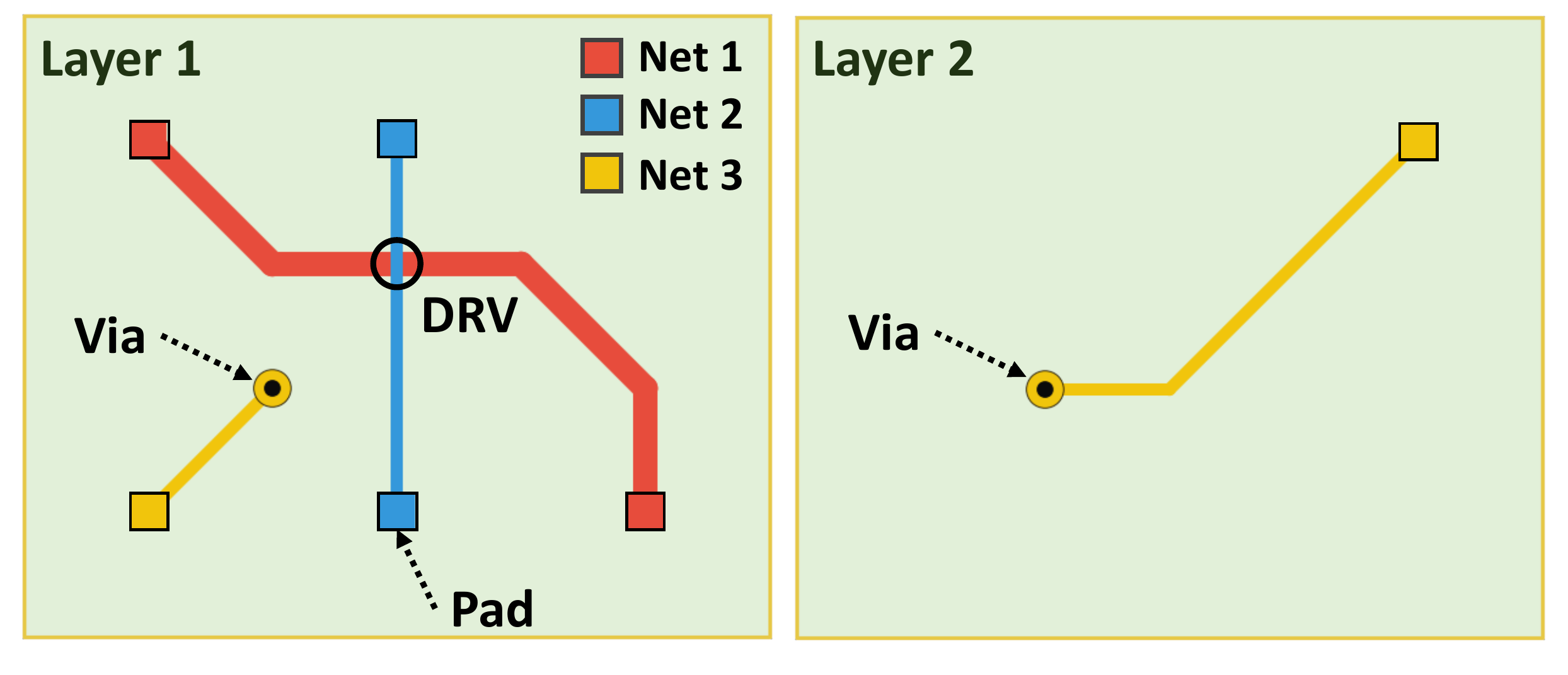}
  \caption{\textbf{Core concepts of PCB routing.} Pads of the same net must be electrically connected. Traces of different nets that come too close incur a design rule violation (DRV). Vias let traces switch layers to avoid such conflicts.}
  \label{fig:pcb-concepts}
  \Description{\textbf{Core concepts of PCB routing.} Pads of the same net must be electrically connected. Traces of different nets that come too close incur a design rule violation (DRV). Vias let traces switch layers to avoid such conflicts.}
\end{figure}

A printed circuit board (PCB) is a multi-layer substrate whose components expose metal contacts called \emph{pads} (\cref{fig:pcb-concepts}).
A \emph{net} is a set of pads that must be electrically connected to share the same signal, and PCB routing is the task of drawing copper \emph{traces} that connect all pads of every net while keeping distinct nets electrically isolated.
A trace is represented as a sequence of straight \emph{track} segments, each lying on a single layer.
A routing scheme is \emph{grid-based} if track endpoints must lie on a fixed lattice, and \emph{gridless} if they may take arbitrary positions on the canvas.
A trace may switch layers through a \emph{via}, a plated hole that joins tracks across layers.
A routed board is valid only if it passes a \emph{Design Rule Check} (DRC). Any \emph{design rule violation} (DRV), for example a \emph{clearance} violation between distinct nets, blocks fabrication.
We call a DRV-free routing \emph{geometrically feasible}. A single violation invalidates an otherwise plausible board.

\section{Background}\label{sec:background}

Among DRV-free routings, quality is measured by total \emph{wirelength} and \emph{via count}, which correlate with signal delay and fabrication cost.
For the routing state $s$, let $n_{\mathrm{drv}}(s)$, $\ell(s)$, and $n_{\mathrm{via}}(s)$ denote the number of DRVs, the total wirelength, and the via count, respectively.
The PCB routing task is then formulated as a constrained optimization problem:
\begin{equation}
  s^{\star} = \arg\min_{s\in\mathcal{R}} \lambda_w\,\ell(s) + \lambda_v\,n_{\mathrm{via}}(s)
  \;\;\text{subject to}\;\; n_{\mathrm{drv}}(s) = 0,
  \label{eq:opt}
\end{equation}
where $\mathcal{R}$ is the set of board states and $\lambda_w, \lambda_v \ge 0$ are user-specified weights balancing wirelength against via count.

\subsection{\kicad{}: An Open, Programmable EDA Suite}\label{sec:bg-kicad}
\kicad{}~\citep{kicad} is an open-source EDA suite maintained under the Linux Foundation~\citep{linuxfoundation_kicad}.
It provides the end-to-end PCB design flow within a unified engine, from component placement and trace routing to design rule checking.
The C++ engine supports exposing its native operations to Python through extensible bindings~\citep{kicad_ipc} and stores boards in an open, human-readable format.
The broader ecosystem includes open routing resources such as PCBench~\citep{pcbench_github} and the routers Freerouting~\citep{freerouting}, OrthoRoute~\citep{orthoroute}, and KiCadRoutingTools~\citep{kicadroutingtools}.
These properties make \kicad{} a production-grade open-source alternative to commercial EDA suites.
     

\section{\ours{}: An Engine-Grounded\texorpdfstring{\\}{ }PCB Routing Environment}\label{sec:env}

\begin{figure*}[t]
  \centering
  \includegraphics[width=0.9\linewidth]{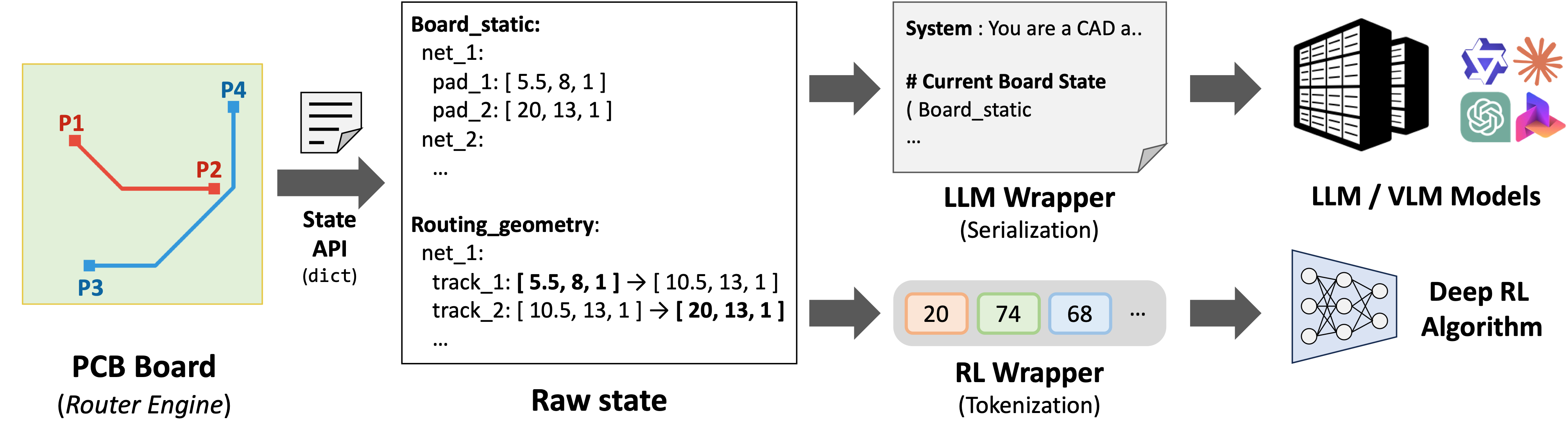}
  \caption{\textbf{Gym Wrappers.} Two Gym wrappers communicate with the shared engine through a unified state dictionary, which they re-encode into agent-specific formats: tokenized sequences for the RL and serialized tool-call schemas for the LLM.}
  \label{fig:state-repr}
  \Description{\textbf{Two Gym wrappers communicate with the shared engine through a unified state dictionary,} re-encoding it into agent-specific formats: a tokenized sequence for the RL wrapper and an s-expression with a tool-call schema for the LLM wrapper.}
\end{figure*}

\ours{} wraps \kicad{} engine as a Gym environment that faithfully reproduces a real-world PCB routing workflow.
The environment is organized as three layers: the C++ \kicad{} engine performing the actual routing and design rule checks (\S\ref{sec:env-engine}), the Gym environment defining an MDP over the engine's native operations (\S\ref{sec:env-base}), and two tailored wrappers re-encoding this MDP for RL and LLM agents (\S\ref{sec:env-wrapper}).

\subsection{Python API via Bindings to the \kicad{} Engine}\label{sec:env-engine}

The bottom layer provides 58 \textit{Python APIs} built on bindings to the \kicad{} engine (\S\ref{sec:bg-kicad}): 14 reproduce the core routing API of \texttt{PNS::ROUTER}, 28 directly wrap \kicad{}'s DRC and board-state extraction APIs, and 16 provide auxiliary utilities.\footnote{The full list is provided in Appendix~\ref{sec:action-apilist}.
\ours{} is built against \kicad{}~9.0.8 with \texttt{kicad-python}~0.6.0.} All higher layers reach the engine only through these headless APIs, so every agent action executes as a native \kicad{} engine operation, as in a human engineer's workflow (\cref{fig:intro-overview}).
Because each environment holds its own engine instance, \ours{} also runs vectorized, stepping many boards in parallel.

\subsection{MDP Formulation}\label{sec:env-base}
The Gym environment formalizes the routing workflow as an MDP.
An episode loads a real PCB file (\texttt{.kicad\_pcb}) as its initial state, a \emph{bare board} whose components are fixed and whose nets are unrouted. The agent then routes the board through successive API calls, observing the result of each, until every net is routed or a step limit is reached.

\paragraph{State.}\label{sec:env-state}
At each time step $t \in \{0, 1, \ldots, T\}$, the state $s_t$ captures a snapshot of the board, where $t=0$ is the bare board and $t=T$ the terminal state.
A \kicad{} board is naturally described as a composition of heterogeneous geometric objects organized in a hierarchical structure (e.g., \texttt{net} $\rightarrow$ \texttt{pad} $\rightarrow$ \texttt{coordinate}).
We introduce a \textbf{nested dictionary} that faithfully represents the exact numeric values and hierarchy of \kicad{} (\cref{fig:state-repr}).
Unlike abstracted representations such as a grid or a rendered image, this lets the agent observe the board's exact state, as the EDA domain's strict design rules demand.
The dictionary splits by mutability during routing. \textbf{Board\_static} holds immutable objects such as pad geometries, internal obstacles, and design rules. \textbf{Routing\_geometry} holds objects added during routing, such as tracks, vias, and the per-pad connectivity status.
Further details are provided in Appendix~\ref{sec:state-schema}.

{
\paragraph{Action.}\label{sec:env-action}
\begin{table}[t]
\centering
\caption{\textbf{Routing actions.}}
\label{tab:action-description}
\footnotesize
    \begin{tabular}{p{0.15\linewidth} p{0.21\linewidth} p{0.48\linewidth}}
    \toprule
    \textbf{Action} & \textbf{Arguments} & \textbf{Description} \\
    \midrule
    \texttt{net\_select} & \texttt{net\_id} &
    Select \texttt{net\_id} for subsequent tracks and vias.\\
    \midrule
    \texttt{net\_end} & -- &
    Release the current net.\\
    \midrule
    \texttt{start\_route} & \texttt{p\_start} &
    Begin a route from \texttt{p\_start}, a pad center or existing track/via endpoint of the net. \\
    \midrule
    \texttt{make\_line} &
    \makecell[tl]{\texttt{p\_end},\\ \texttt{routing\_mode}} &
    Extend the route to \texttt{p\_end} on the current layer.
    \texttt{routing\_mode} sets how it interacts with existing traces.\\
    \midrule
    \texttt{make\_via} &
    \makecell[tl]{\texttt{p\_end},\\ \texttt{routing\_mode}} &
    Execute \texttt{make\_line}  to \texttt{p\_end}, then place a via at \texttt{p\_end}.\\
    \midrule
    \texttt{finish} &
    \makecell[tl]{\texttt{routing\_mode}} &
    Auto-complete the route to the Euclidean-nearest unconnected pad on the current layer.\\
    \bottomrule
    \end{tabular}
\end{table}
\begin{figure*}[t]
  \centering
  \includegraphics[width=0.8\textwidth]{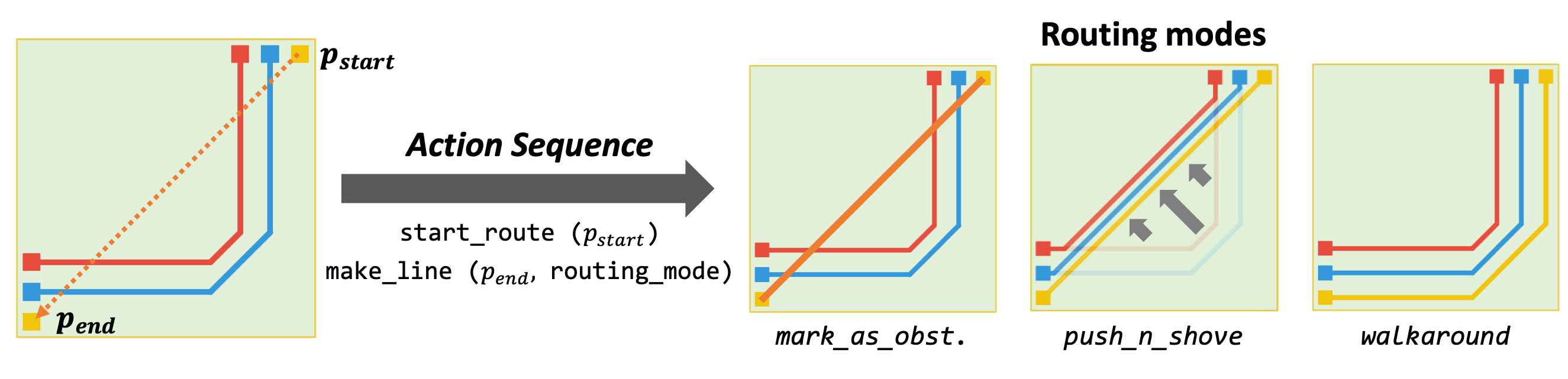}
  \caption{\textbf{Example Action of \ours{}.} An example of \texttt{make\_line}, where the \texttt{routing\_mode} argument alters routing behavior through the \kicad{} routing engine, yielding different trace geometries across the three modes.}
  \label{fig:act-repr}
  \Description{\textbf{Example Action of \ours{}.} An example of \texttt{make\_line}, where the \texttt{routing\_mode} argument alters routing behavior through the \kicad{} routing engine, yielding different trace geometries across the three modes.}
\end{figure*}
An action is a pair of an action type and its arguments, composed from the APIs exposed in \S\ref{sec:env-engine}.
\cref{tab:action-description} summarizes the six action types.
In the example of \cref{fig:act-repr}, \texttt{make\_line} takes arguments (\texttt{p\_end}, \texttt{routing\_mode}) and draws a track to \texttt{p\_end}.
The \texttt{routing\_mode} argument controls how the engine realizes this track, either pushing blocking wires aside (\texttt{push\_n\_shove}) or detouring around them (\texttt{walkaround}), while satisfying strict design rules such as clearance.
The \kicad{} engine exposes a finite set of candidate points for \texttt{p\_end}, such as pad centers and the endpoints of existing tracks and vias. This discretizes the continuous plane into a small candidate set, as an engineer's editor does, and the agent selects a \texttt{p\_end} and a \texttt{routing\_mode} rather than drawing raw segments itself.
The RL agent selects \texttt{p\_end} from this candidate set.
The LLM agent is prompted toward the same points but may emit free coordinates for detour waypoints.

\paragraph{Reward.}\label{sec:env-reward}
We relax Equation~\eqref{eq:opt}'s hard DRV constraint into a penalty and define a potential function $\Phi(s)$ that measures the board quality,
\begin{equation*}
\Phi(s) \;=\; -\big(f_{\text{drv}}(s) + \lambda_w\, \ell(s) + \lambda_v\, n_{\text{via}}(s)\big).
\end{equation*}
The DRV count $n_{\text{drv}}(s)$ comes from \kicad{}'s DRC API, which evaluates violations across 38 checks.
The two weights $\lambda_w, \lambda_v \ge 0$ reuse those of Equation~\eqref{eq:opt}, and $f_{\text{drv}}(s)$ is a concave penalty that increases with every violation and is steepest near zero DRVs. This penalty dominates the wirelength and via terms, so maximizing $\Phi$ drives the policy toward DRV-free routings that approximate the optimum of Equation~\eqref{eq:opt}.
The full check catalog is given in Appendix~\ref{sec:drc-catalog}, and the exact form of $f_{\text{drv}}$ in Appendix~\ref{sec:reward-detail}.

The MDP's reward is the \emph{terminal} reward, with $r_t = 0$ for $t < T$ and $r_T = \Phi(s_T) - \Phi(s_0)$ at termination, scoring the completed board against the bare board.
For dense training we also expose a \emph{per-step} form, $r_t = \Phi(s_{t+1}) - \Phi(s_t)$.
With an undiscounted return ($\gamma = 1$), the per-step rewards telescope to $\sum_{t=0}^{T-1} r_t = \Phi(s_T) - \Phi(s_0)$, so the per-step form is a potential-based shaping of the terminal reward~\citep{ng1999policy} and shares its optimal policy.
Since GRPO~\citep{shao2024deepseekmath} uses only the total episode return, the two reward forms coincide for it.
For PPO~\citep{schulman2017ppo}, which estimates advantages with a value function, the per-step form gives a denser credit-assignment signal.
Because the equivalence holds for any potential, \ours{} exposes $\Phi$ as a pluggable training objective.
Further details are provided in Appendix~\ref{sec:hp-rl-main}.

\subsection{Environment Wrappers}\label{sec:env-wrapper}
We implement separate wrappers that expose the nested-dictionary raw state and structured actions through interfaces tailored to RL and LLM tasks.
The RL wrapper tokenizes each geometric object (pad, track, via, point) into an individual token by combining a Fourier feature map of its coordinates with an entity-type embedding, yielding a flat sequence that preserves the native object hierarchy.
The action is decoded autoregressively: the head emits an action type, then only the parameter tokens that type requires. Depending on the type, these are a point selected by a pointer network from the engine-provided candidate set, a routing mode, or both.
This keeps every emitted action syntactically complete and within the engine's valid-action set by construction.

For the LLM wrapper, we adopt an S-expression representation that preserves the nested hierarchy and key–value relations while minimizing token count, retaining only routing-relevant fields (coordinates, connectivity, layers, effective DRC constraints).
The prompt distills the task into a concise set of routing guidelines.
Finally, the wrapper supplies a compact action history that flags the most recent rejected or no-effect action, so the agent avoids repeating it.
Full details of both wrappers are provided in Appendix~\ref{sec:impl-detail}.
\section{\oursbench{}}\label{sec:bench}

We introduce \oursbench{}, a benchmark for geometric feasibility reasoning in PCB routing.
It provides a method-agnostic evaluation protocol, three board datasets, and eight evaluation metrics.

\subsection{Task and Evaluation Protocol}\label{sec:bench-protocol}

Each instance is a \kicad{} board in the native \texttt{.kicad\_pcb} format with its routes removed.
Given the component placement and design rules, the agent completes the routing in the same file.
The \oursbench{} evaluator scores the completed board with the \kicad{} engine.
Since the evaluation depends only on the board, the protocol is method-agnostic.
Routing methods that do not use \ours{} and the reference solutions are scored the same way as our agents.

\begin{table}[!tb]
  \centering
  \caption{\textbf{Dataset statistics.} The table reports the number of boards and the per-board ranges of net, pad, and layer counts.}
  \label{tab:bench-stats}
  \small
  \begin{adjustbox}{max width=\linewidth}
  \begin{tabular}{lcccc}
    \toprule
    Dataset & \# inst. (train/valid/test) & \# nets & \# pads & \# layers \\
    \midrule
    D1 (Grid) & 10{,}000 / 128 / 128 & 5 & 10 & 1 \\
    D2 (Gridless) & 10{,}000 / 128 / 128 & 4--6 & 8--21 & 2 \\
    D3-A (Small) & -- / -- / 100 & 2--13 & 6--31 & 2--4 \\
    D3-B (Medium) & -- / -- / 287 & 5--42 & 31--100 & 2--4 \\
    D3-C (Large) & -- / -- / 292 & 6--451 & 101--2{,}103 & 2--8 \\
    \bottomrule
  \end{tabular}
  \end{adjustbox}
\end{table}

 \begin{figure}[!tb]
    \centering
    \begin{subfigure}{0.36\linewidth}
        \centering
        \begin{minipage}[c][\linewidth][c]{\linewidth}
            \centering
            \includegraphics[width=\linewidth,height=\linewidth,keepaspectratio]{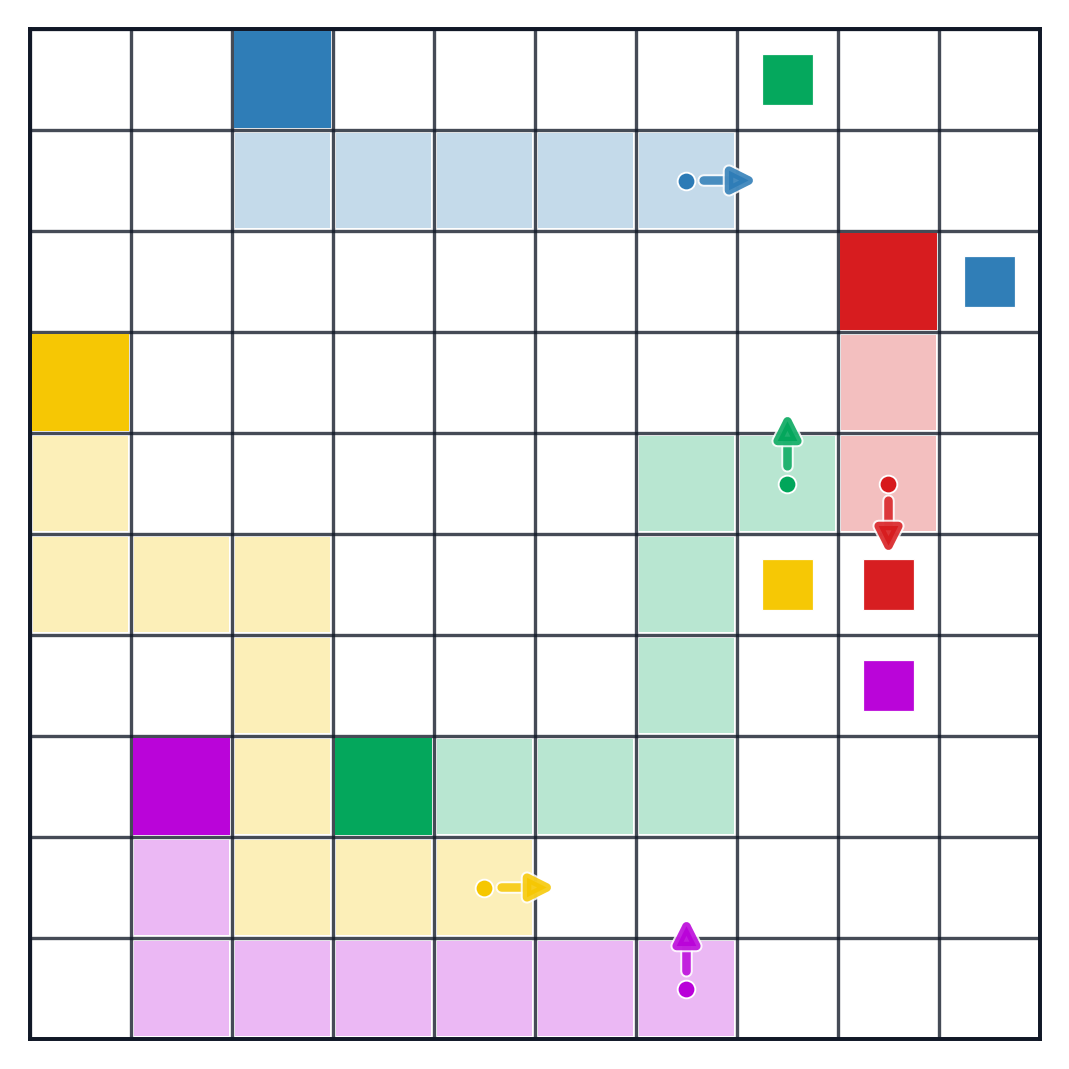}
        \end{minipage}
        \caption{D1-10}
        \label{fig:rq1-instances-grid10}
    \end{subfigure}
    \hskip 0.1\linewidth
    \begin{subfigure}{0.36\linewidth}
        \centering
        \begin{minipage}[c][\linewidth][c]{\linewidth}
            \centering
            \includegraphics[width=\linewidth,height=\linewidth,keepaspectratio]{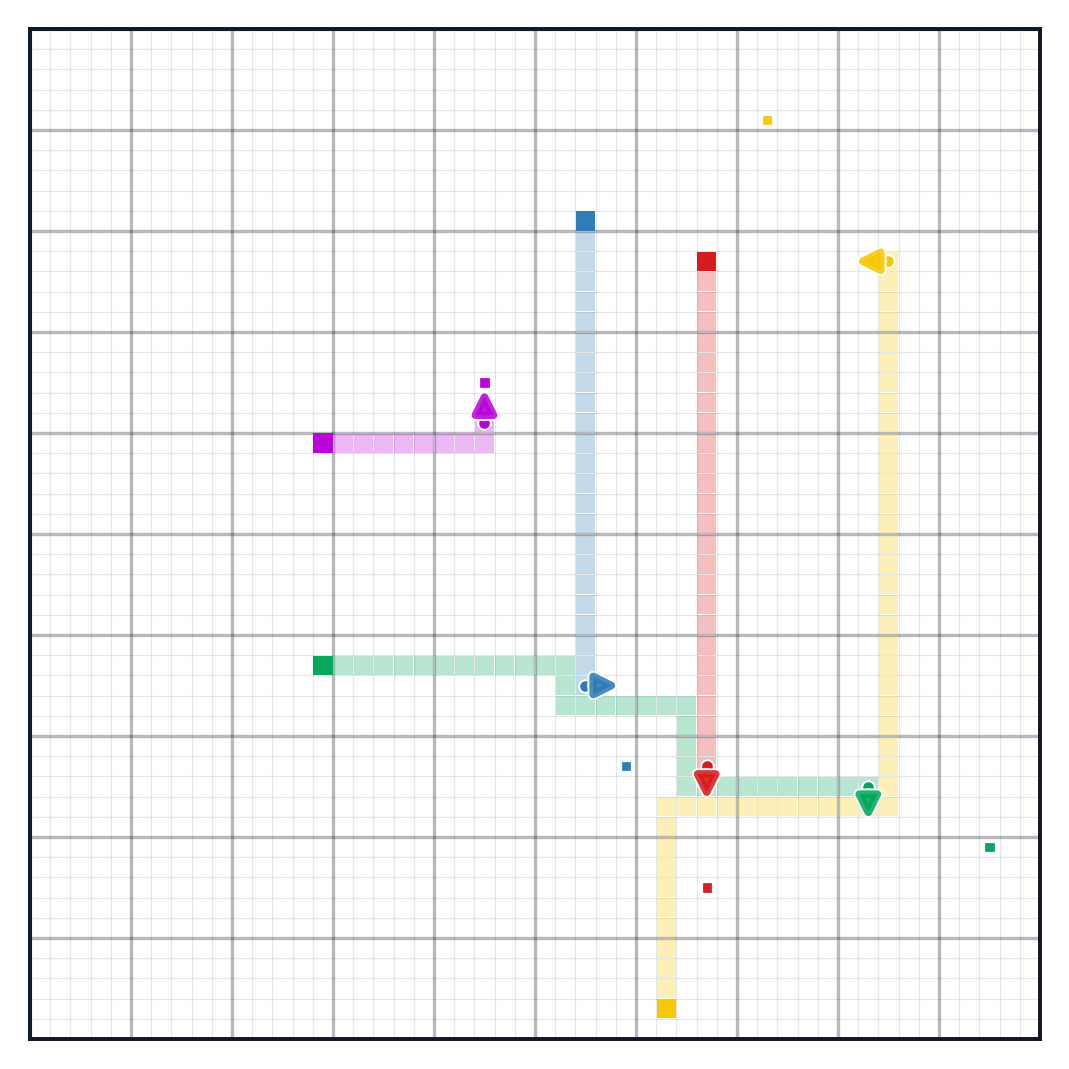}
        \end{minipage}
        \caption{D1-50}
        \label{fig:rq1-instances-grid50}
    \end{subfigure}

    \begin{subfigure}{0.37\linewidth}
        \centering
        \begin{minipage}[c][\linewidth][c]{\linewidth}
            \centering
            \includegraphics[width=\linewidth,height=\linewidth,keepaspectratio]{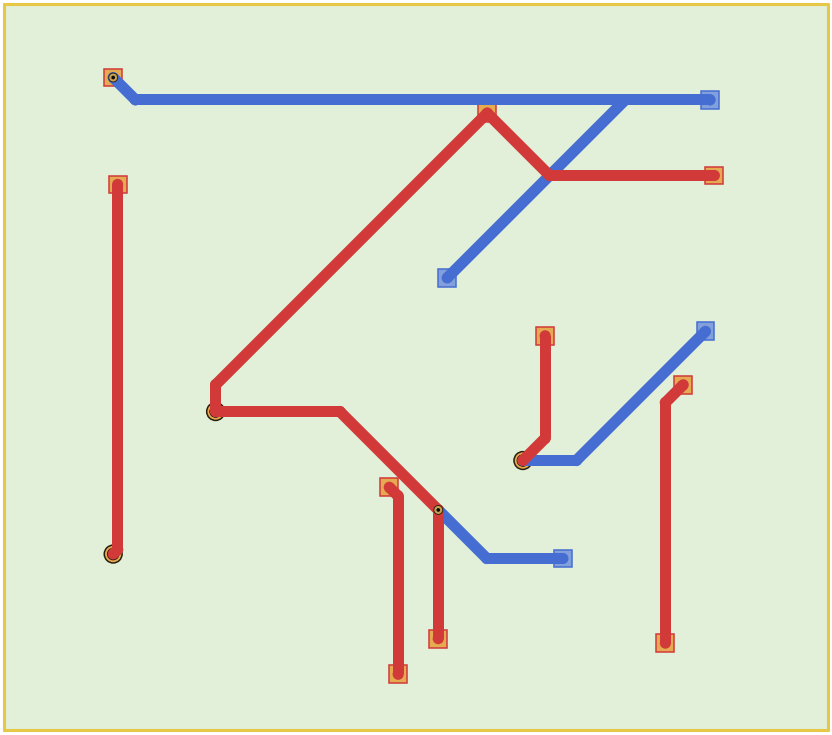}
        \end{minipage}
        \caption{D2}
        \label{fig:t2}
    \end{subfigure}
    \hskip 0.05\linewidth
    \begin{subfigure}{0.39\linewidth}
        \centering
        \includegraphics[width=\linewidth,height=\linewidth,keepaspectratio]{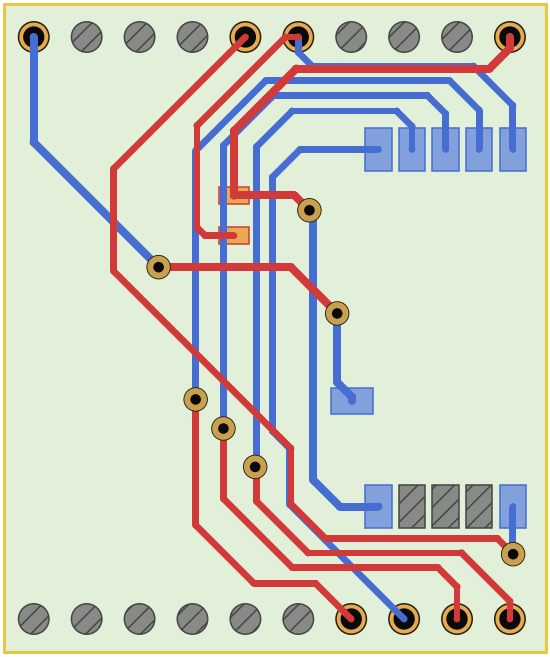}
        \caption{D3-A}
        \label{fig:t3a}
    \end{subfigure}

    \begin{subfigure}{0.41\linewidth}
        \centering
        \begin{minipage}[c][0.7\linewidth][c]{\linewidth}
            \centering
            \includegraphics[width=\linewidth,height=0.7\linewidth,keepaspectratio]{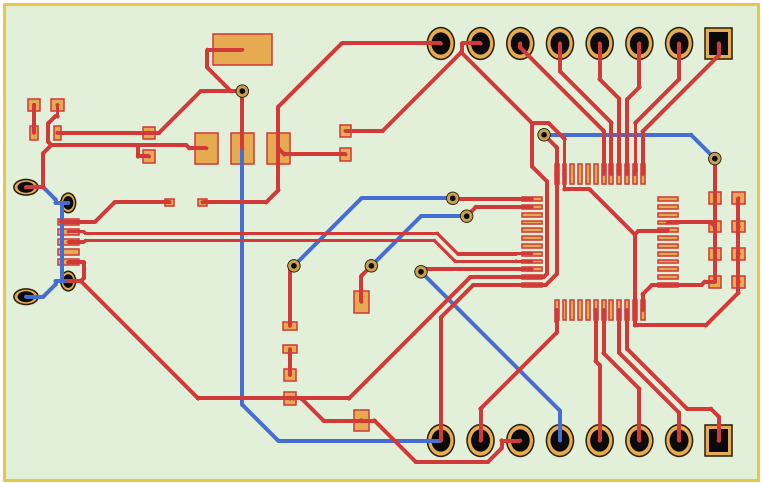}
        \end{minipage}
        \caption{D3-B}
        \label{fig:t3b}
    \end{subfigure}
    \hfill
    \begin{subfigure}{0.44\linewidth}
        \centering
        \begin{minipage}[c][0.7\linewidth][c]{\linewidth}
            \centering
            \includegraphics[width=\linewidth,height=0.7\linewidth,keepaspectratio]{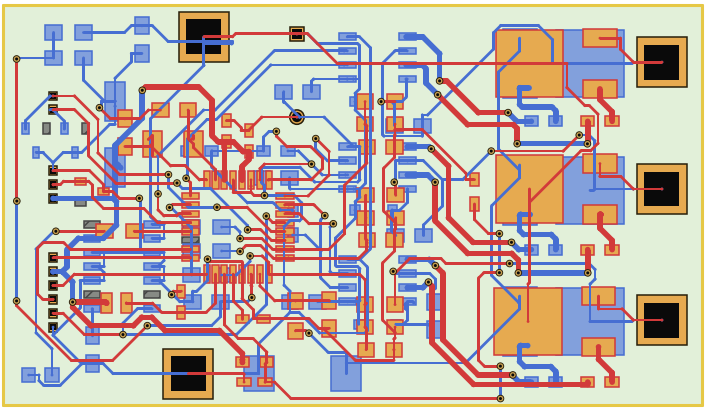}
        \end{minipage}
        \caption{D3-C}
        \label{fig:t3c}
    \end{subfigure}

    \caption{
\textbf{Example boards from \oursbench{}.} Synthetic grid-based boards (D1), synthetic gridless boards (D2), and real-world open-source boards (D3).
    }
    \label{fig:dataset-examples}
    \Description{\textbf{Example boards from \oursbench{}.} Synthetic grid-based boards (D1), synthetic gridless boards (D2), and real-world open-source boards (D3).}
\end{figure}

\subsection{Board Datasets}\label{sec:bench-tasks}

\cref{tab:bench-stats} summarizes the instance counts and key statistics of the three board datasets, and \cref{fig:dataset-examples} presents representative examples from each.
The datasheet and license information are provided in Appendices~\ref{sec:datasheet} and~\ref{sec:license}.

\paragraph{(D1) Synthetic Grid-based Boards.}
D1 replicates the grid-routing task of Jumanji's \emph{Connector} environment~\citep{jumanjibench} in \ours{}.
Boards are generated in \kicad{} format from the Connector instance distribution, and agents route on the same grid, one wire per cell.
Five grid sizes are provided, $G \in \{10, 50, 100, 200, 500\}$, denoted D1-10 through D1-500.
D1 enables direct comparison with state-of-the-art methods developed on Connector, such as Sable~\citep{sable}.

\paragraph{(D2) Synthetic Gridless Boards.}
D2 is a synthetic gridless routing dataset for training and in-distribution evaluation.
We release its board generator, which controls the number of nets and pads per net and thus adjusts routing difficulty directly.

\paragraph{(D3) Open-Source Boards.}
D3 is a curated set of 679 boards derived from PCBench's open-source PCB corpus~\citep{pcbench_github}.
We convert these boards to the \kicad{}~9 format, preserving their design-rule settings, and retain only boards that remain fully connected and DRV-free after conversion.
Each board retains its original routed traces, which serve as reference solutions for evaluation.
These boards are partitioned by pad count into D3-A/B/C, from smallest to largest.
For zero-shot evaluation, all methods share a common set of 99 D3-A boards and 10 D3-B boards.
Owing to the high per-board cost of LLM evaluation, the D3-B subset is capped at ten boards, sampled to represent the split's size range (Appendix~\ref{sec:d3-splits}).
We reserve D3-C, whose boards reach hundreds of nets, as a challenge split for future research.

\subsection{Evaluation Metrics}\label{sec:bench-metrics}

We use \textbf{Clean Pass} (CP) as the primary metric.
CP counts a board as successful only if the selected rollout achieves full connectivity with zero DRVs.
We further report \textbf{Potential Gain} (Pot.) as a routing-quality metric, defined as the change in board potential from the bare-board state to the completed routed state.
Additional diagnostics include \textbf{Routability} (Rout.), the fraction of target connections successfully routed; \textbf{wallclock time} (Time); \textbf{DRV count} (DRV); and physical cost metrics such as \textbf{total wirelength} (WL) and \textbf{via count} (Via).
For LLM agents we also report \textbf{Parse-fail}, the fraction of boards for which no rollout loads as a valid \texttt{.kicad\_pcb}, i.e., whose outputs break the \kicad{} file format.

For each board we draw five rollouts and select the single rollout with the largest potential gain.
All metrics are computed on this selected rollout, except Time, which is averaged over all five rollouts, and Parse-fail, which is computed from all five.
We write @5 for this best-of-five protocol, the default for all reported metrics, and @1 when a metric scores a single rollout without selection.
Deterministic methods are run once, as selection does not apply.
Formal definitions and implementation details are provided in Appendix~\ref{sec:metrics}.




\section{Experiments}\label{sec:exp}

We evaluate on the three task suites of \S\ref{sec:bench} (D1, D2, and D3).
Through D1, we show that the \kicad{}-API actions of \ours{} scale to fine grid resolutions where grid-action methods break down.
Through D2 and D3, we evaluate the geometric feasibility reasoning of the LLM and RL agents in \ours{} against diverse baselines. The RL agents are trained on the D2-train split, while the LLM agents are used off-the-shelf without task-specific fine-tuning.

\subsection{Experimental Setup}\label{sec:exp-setup}
\paragraph{\ours{} agents.}
We build LLM and RL agents that act in a closed loop over the full \ours{} MDP, the \emph{interactive} mode throughout.
For \textbf{LLM agents}, we examine how routing performance varies with model capacity across GPT-5.4, GPT-5.4-mini, GPT-5.4-nano~\citep{openai2026gpt54,openai2026gpt54mininano}, and Qwen3.5-397B~\citep{qwen2026qwen35}.
For \textbf{RL agents}, we train a small Transformer policy from scratch in \ours{} and report three variants, PPO (per-step), PPO (terminal), and GRPO, using the per-step and terminal rewards defined in \S\ref{sec:env-reward}.
Since PPO (per-step) is our default policy, we write it as plain PPO unless noted otherwise.
We additionally train PPO (w/o \texttt{finish}), removing \texttt{finish} from the action space (\S\ref{sec:exp-rl}).
Further training details are provided in Appendix~\ref{sec:hp-rl-main}.

\begin{figure}[!tb]
  \centering
  \includegraphics[width=0.85\linewidth]{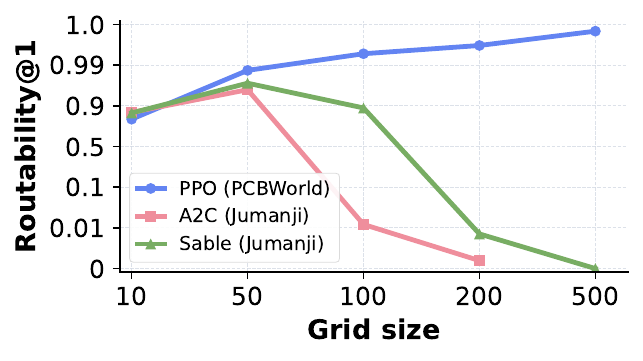}
  \caption{\textbf{Simulation Results of Synthetic Grid-based Boards (D1).} Single-rollout routability (Rout.@1$\uparrow$) across grid sizes. The vertical axis is warped symmetrically about 0.5 to make both ends of the range legible. A2C (Jumanji) exhausts memory at D1-500, so its curve stops at D1-200.}
  \label{fig:rq1-instances}
  \label{fig:rq1-instances-rout}
  \Description{\textbf{Simulation Results of Synthetic Grid-based Boards (D1).} Single-rollout routability (Rout.@1$\uparrow$) across grid sizes. The vertical axis is warped symmetrically about 0.5 to make both ends of the range legible. A2C (Jumanji) exhausts memory at D1-500, so its curve stops at D1-200.}
\end{figure}

\begin{table*}[!tb]
\centering
\caption{\textbf{Routing quality on D2, D3-A, and D3-B.} Results follow the @5 protocol of \S\ref{sec:bench-metrics}.
We report CP$\uparrow$, Pot.$\uparrow$, Rout.$\uparrow$, and Time$\downarrow$ in seconds.
Best CP and Pot. per split (excluding the Reference row) are in \textbf{bold}. Freerouting and the RL agents report the mean over 4 seeds. Per-metric breakdowns and @1 results are provided in Appendix~\ref{sec:add-results}.}
\label{tab:rq2}
\footnotesize
\setlength{\tabcolsep}{4pt}
\resizebox{0.85\textwidth}{!}{
\begin{tabular}{l!{\hskip 10pt}cccc!{\hskip 12pt}cccc!{\hskip 12pt}cccc}
\toprule
 & \multicolumn{4}{c}{D2} & \multicolumn{4}{c}{D3-A (99 Boards)} & \multicolumn{4}{c}{D3-B (10 Boards)} \\
\cmidrule(lr{\dimexpr 0.5em+12pt\relax}){2-5}\cmidrule(lr{\dimexpr 0.5em+12pt\relax}){6-9}\cmidrule(lr){10-13}
Method
& CP$\uparrow$ & Pot.$\uparrow$ & Rout.$\uparrow$ & Time$\downarrow$
& CP$\uparrow$ & Pot.$\uparrow$ & Rout.$\uparrow$ & Time$\downarrow$
& CP$\uparrow$ & Pot.$\uparrow$ & Rout.$\uparrow$ & Time$\downarrow$ \\
\midrule
Reference
& - & - & - & -
& 1.00 & 23.16 & 1.00 & -
& 1.00 & 63.96 & 1.00 & - \\
\midrule
Freerouting
& \textbf{1.00} & 15.47 & 1.00 & 2.69
& 0.80 & \textbf{22.71} & 0.91 & 7.09
& \textbf{0.78} & \textbf{61.06} & 1.00 & 9.94 \\
OrthoRoute
& 0.01 & 0.38 & 0.34 & 2.54
& 0.02 & -6.16 & 0.53 & 2.20
& 0.00 & -10.25 & 0.44 & 9.30 \\
KiCadRoutingTools
& \textbf{1.00} & 15.52 & 1.00 & 0.82
& 0.74 & 20.00 & 0.94 & 0.65
& 0.20 & 40.94 & 0.86 & 3.27 \\
\midrule
\multicolumn{13}{l}{\scriptsize\textit{\ours: Inference on different LLM backbones}}\\
GPT-5.4
& 0.96 & 16.05 & 1.00 & 94.04
& 0.65 & 19.42 & 0.91 & 231.15
& 0.00 & 34.72 & 0.62 & 865.83 \\
GPT-5.4-mini
& 0.58 & 12.45 & 0.86 & 33.93
& 0.28 & 14.22 & 0.72 & 56.85
& 0.00 & 30.84 & 0.61 & 422.06 \\
GPT-5.4-nano
& 0.55 & 12.23 & 0.85 & 63.39
& 0.30 & 14.16 & 0.73 & 100.76
& 0.00 & 30.66 & 0.59 & 510.25 \\
Qwen3.5-397B
& 0.33 & 9.77 & 0.74 & 47.58
& 0.34 & 14.88 & 0.75 & 81.82
& 0.00 & 25.99 & 0.51 & 222.29 \\
\midrule
\multicolumn{13}{l}{\scriptsize\textit{\ours: RL trained on Transformer}}\\
PPO
& \textbf{1.00} & \textbf{16.24} & 1.00 & 0.43
& 0.86 & 21.46 & 0.95 & 1.83
& 0.45 & 50.20 & 0.85 & 10.77 \\
GRPO
& \textbf{1.00} & 14.03 & 1.00 & 1.16
& 0.85 & 18.83 & 0.98 & 3.54
& 0.10 & 30.00 & 0.69 & 11.20 \\
PPO (terminal)
& 0.92 & 15.61 & 0.98 & 1.37
& 0.82 & 21.02 & 0.95 & 2.58
& 0.38 & 44.33 & 0.81 & 12.60 \\
\midrule
PPO (w/o \texttt{finish})
& \textbf{1.00} & 16.21 & 1.00 & 0.88
& \textbf{0.94} & 21.78 & 0.99 & 3.05
& 0.42 & 46.45 & 0.87 & 14.44 \\
\bottomrule
\end{tabular}
}
\end{table*}

\paragraph{Baselines.}
We evaluate three baseline families and score the \texttt{.kicad\_pcb} files they produce under the same metrics.
The \textbf{rule-based routers} are Freerouting~\citep{freerouting}, OrthoRoute~\citep{orthoroute}, and KiCadRoutingTools (KRT)~\citep{kicadroutingtools}.
The \textbf{grid-action RL baselines}, A2C (Jumanji)~\citep{jumanjibench} and Sable~\citep{sable}, are trained on Jumanji-Connector v2 and route each net cell by cell on the grid.
The two \textbf{open-loop LLM baselines} produce their full routing in a single pass, with no feedback on any intermediate board state and no revision on the engine's verdict: \emph{plan-and-execute} emits the full sequence of \ours{} routing actions and has it executed in \ours{}, mirroring command-sequence generation in 3D CAD~\citep{cadgpt2025spatial,qi2026pointer}, and \emph{engine-free generation} directly emits the routing as segments and vias in \texttt{.kicad\_pcb} syntax, analogous to one-shot code generation~\citep{liu2023verilogeval,query2cad2024nl,llm4cad2025multimodal}.



\subsection{Scalability Limits of Grid-Action Routing}\label{sec:exp-scale}

On D1, we trained and evaluated each method independently at each grid resolution, training PPO to maximize Rout.@1 to match the objective of A2C (Jumanji) and Sable (\cref{fig:rq1-instances-rout}).
On D1-10, PPO is marginally below the grid-action baselines. 
It reaches near-perfect routability from D1-50 onward, while A2C collapses by D1-100 and Sable by D1-200.
As the grid grows finer (\cref{fig:rq1-instances-grid10,fig:rq1-instances-grid50}), grid-action baselines that route one cell at a time inevitably face a long horizon that weakens credit assignment.
By contrast, PPO issues segment-level \kicad{}-API actions, so its decision horizon is tied to routed segments and does not grow with grid resolution.
The full grid-size sweep numbers are tabulated in Appendix~\ref{sec:add-results}.
  


\subsection{Comparing Routing Agents in \ours{}}\label{sec:exp-overall}
\begin{figure*}[!tb]
    \centering
    \includegraphics[width=0.88\linewidth]{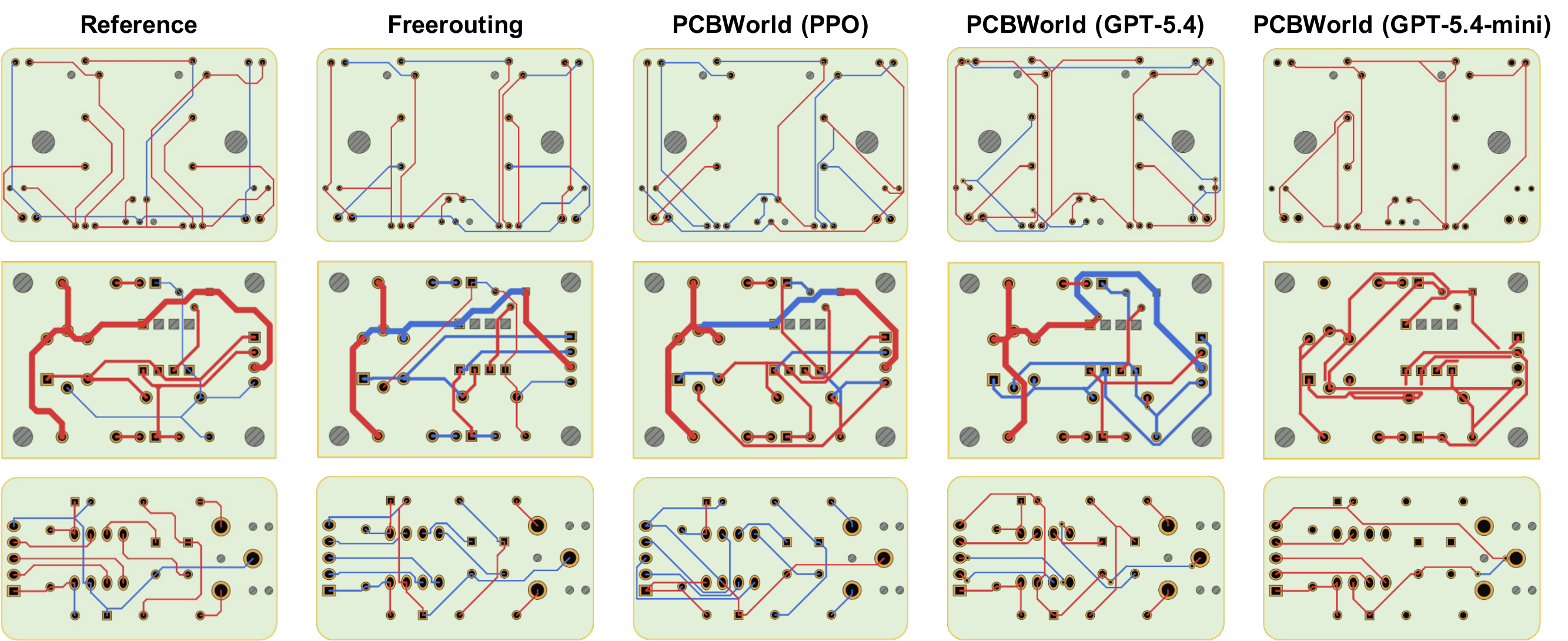}
    \caption{
\textbf{Routed boards on three D3-A instances.} Each column shows one method's output for the same boards. Red and blue traces mark wire segments on the two copper layers. Additional boards are shown in \cref{fig:qualitative-full} (Appendix~\ref{sec:gallery}).
    }
    \label{fig:qualitative-main}
  \Description{\textbf{Routed boards on three D3-A instances.} Each column shows one method's output for the same boards. Red and blue traces mark wire segments on the two copper layers.}
\end{figure*}

We compare all methods in \cref{tab:rq2} using CP as the primary metric, with Pot., Rout., and Time reported alongside.
\cref{fig:qualitative-main} shows qualitative results produced by each method.

Trained from scratch on synthetic D2-train, a compact PPO Transformer outperforms all LLM agents and the remaining rule-based routers, with only Freerouting, the strongest baseline, remaining competitive.
On in-distribution D2-test, PPO matches Freerouting's perfect CP and Rout. (both $1.00$) and leads on Pot. ($16.24$ vs.\ $15.47$) and Time ($0.43$s vs.\ $2.69$s).
On out-of-distribution D3-A, it still attains higher CP ($0.86$ vs.\ $0.80$), indicating generalization.
On the larger D3-B split, Freerouting outperforms PPO on both CP ($0.78$ vs.\ $0.45$) and Pot. ($61.06$ vs.\ $50.20$). This generalization thus reaches its limit on boards far beyond the D2-train size range (\cref{tab:bench-stats}).

LLM agents leverage the engine's native operations without any routing-specific training, and their routing ability scales with model capacity.
GPT-5.4 reaches a CP of $0.96$ on D2, ahead of its mini ($0.58$) and nano ($0.55$) variants.
All LLM agents nonetheless obtain $0.00$ CP on D3-B, leaving large-board routing in \oursbench{} an open challenge for current LLMs.
The LLM agents are also far slower than the routers and PPO ($94.04$s vs.\ $0.43$s on D2 for GPT-5.4), as every action requires a full model inference.
The strong performance of a compact PPO policy suggests RL fine-tuning of LLMs as a direction toward expert routing agents.

\subsection{Impact of Engine Engagement in LLM Agents}\label{sec:exp-llm}

\cref{tab:llm-ablation} compares three engine-engagement levels of the LLM agent on D3-A, \emph{interactive}, \emph{plan-and-execute}, and \emph{engine-free}. On our primary metrics CP and Pot., the three levels rank in this order consistently across all GPT models.

Leveraging the native API is essential for design-rule-clean routing.
The engine-free agent records far worse DRV than the interactive and plan-and-execute agents, since generating raw wire segments while keeping every track within the design rules is beyond current models.
Smaller models degrade further. Under engine-free generation, for $10\%$ of D3-A boards none of GPT-5.4-nano's five rollouts even loads as a valid \kicad{} board (Parse-fail, \cref{tab:llm-ablation}).

Although the plan-and-execute and interactive agents share the same \ours{} action space and both form connections, only the interactive agent keeps them design-rule-clean, so DRV and CP diverge sharply on D3-A (\cref{tab:llm-ablation}). As in an engineer's routing workflow, both acting through the native API and accurate observation of the board state are essential for clean PCB routing.
\cref{fig:llm_ablation} extends the D3-A comparison of \cref{tab:llm-ablation} to D2.
The interactive $>$ plan-and-execute $>$ engine-free ordering on CP and Pot.\ holds consistently across all GPT models on D2 as well.

\begin{figure*}[!tb]
  \centering
  \begin{subfigure}[t]{\linewidth}
    \centering
    \includegraphics[width=0.9\linewidth]{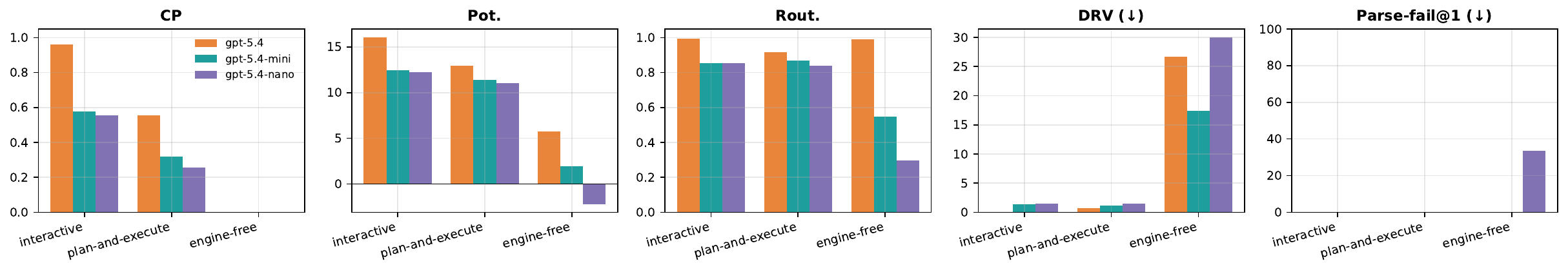}
    \caption{D2 (Synthetic Gridless Board).}
    \label{fig:llm_ablation_t2}
  \end{subfigure}
  \\[2pt]
  \begin{subfigure}[t]{\linewidth}
    \centering
    \includegraphics[width=0.9\linewidth]{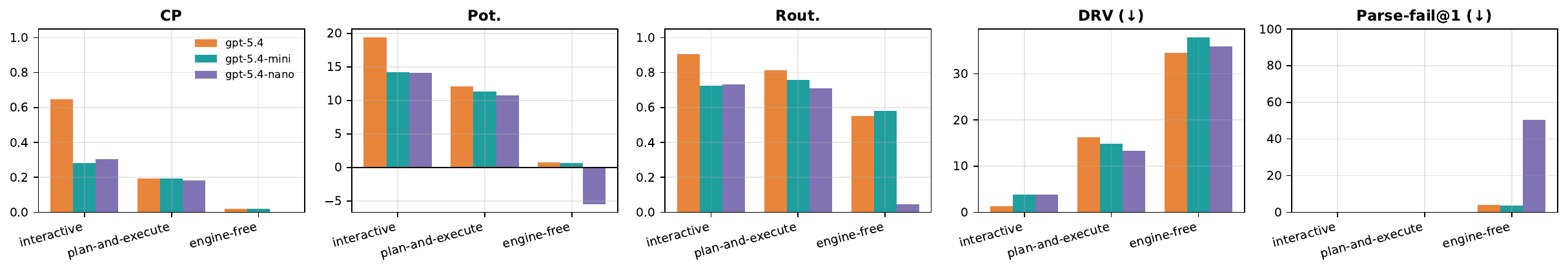}
    \caption{D3-A (Open-source Board).}
    \label{fig:llm_ablation_t3a}
  \end{subfigure}
  \caption{\textbf{Engine engagement in the LLM agent on D2 and D3-A} across interactive, plan-and-execute, and engine-free modes.
Panels report CP, Pot., Rout., and DRV under the @5 protocol and Parse-fail@1, the fraction of rollouts that fail to load, plotted in percent; \cref{tab:llm-ablation} reports Parse-fail under the default @5 protocol.}
  \label{fig:llm_ablation}
  \Description{\textbf{Engine engagement in the LLM agent on D2 and D3-A} across interactive, plan-and-execute, and engine-free modes.
Panels report CP, Pot., Rout., and DRV under the @5 protocol and Parse-fail@1, the fraction of rollouts that fail to load, plotted in percent; \cref{tab:llm-ablation} reports Parse-fail under the default @5 protocol.}
\end{figure*}

\begin{table}[!tb]
\centering
\caption{\textbf{Engine engagement in the LLM agent on D3-A.} We compare the interactive, plan-and-execute, and engine-free modes.}
\label{tab:llm-ablation}
\footnotesize
\setlength{\tabcolsep}{3.5pt}
\begin{adjustbox}{max width=\linewidth}
\begin{tabular}{llccccc}
\toprule
Model & Mode & CP$\uparrow$ & Pot.$\uparrow$ & Rout.$\uparrow$ & DRV$\downarrow$ & Parse-fail$\downarrow$ \\
\midrule
Reference & --          & 1.00 & 23.16 & 1.00 & 0.00 & -- \\
\midrule
GPT-5.4 & interactive & \textbf{0.65} & \textbf{19.42} & 0.91 & 1.31 & 0.00 \\
        & plan-and-execute & 0.19 & 12.14 & 0.81 & 16.28 & 0.00 \\
        & engine-free & 0.02 & 0.76  & 0.55 & 34.58 & 0.00 \\
\midrule
GPT-5.4-mini & interactive & \textbf{0.28} & \textbf{14.22} & 0.72 & 3.88  & 0.00 \\
             & plan-and-execute & 0.19 & 11.32 & 0.76 & 14.85 & 0.00 \\
             & engine-free & 0.02 & 0.70  & 0.58 & 37.80 & 0.00 \\
\midrule
GPT-5.4-nano & interactive & \textbf{0.30} & \textbf{14.16} & 0.73 & 3.84  & 0.00 \\
             & plan-and-execute & 0.18 & 10.80 & 0.71 & 13.26 & 0.00 \\
             & engine-free & 0.00 & -5.45 & 0.04 & 35.89 & 0.10 \\
\bottomrule
\end{tabular}
\end{adjustbox}
\end{table}

\subsection{Impact of Reward and Action Design in RL}\label{sec:exp-rl}

\paragraph{Finer credit assignment improves the learned policy.}
We compare PPO (per-step) against the terminal-reward variants PPO (terminal) and GRPO (\cref{tab:rq2}).
All three train to competent policies, with PPO (per-step) the strongest overall, showing the highest Pot.\ on all three splits and CP that matches or exceeds the other two.
PPO (terminal) and GRPO both learn from the completed board alone.
They perform comparably on D2 and D3-A, while PPO (terminal) holds up better on the larger D3-B boards (CP $0.38$ vs.\ $0.10$, Pot.\ $44.33$ vs.\ $30.00$).
Since PPO (terminal) differs from PPO only in reward density, the gain traces to the denser per-step shaping signal (\S\ref{sec:env-reward}).

\paragraph{The policy does not rely on engine auto-completion.}
We trained PPO (w/o \texttt{finish}), removing the strongest form of engine assistance from its action space (\cref{tab:action-description}), to test whether the agent can lay every segment on its own.
Routing quality remains comparable on all three splits, with CP on D3-A improving to $0.94$ (\cref{tab:rq2}).
The geometrically feasible routing thus comes from the learned policy rather than the engine's auto-completion.
We keep \texttt{finish} in the default action space because it is a native operation of the \kicad{} engine, which \ours{} exposes as is.

\begin{figure}[!tb]
  \centering
  \includegraphics[width=\linewidth]{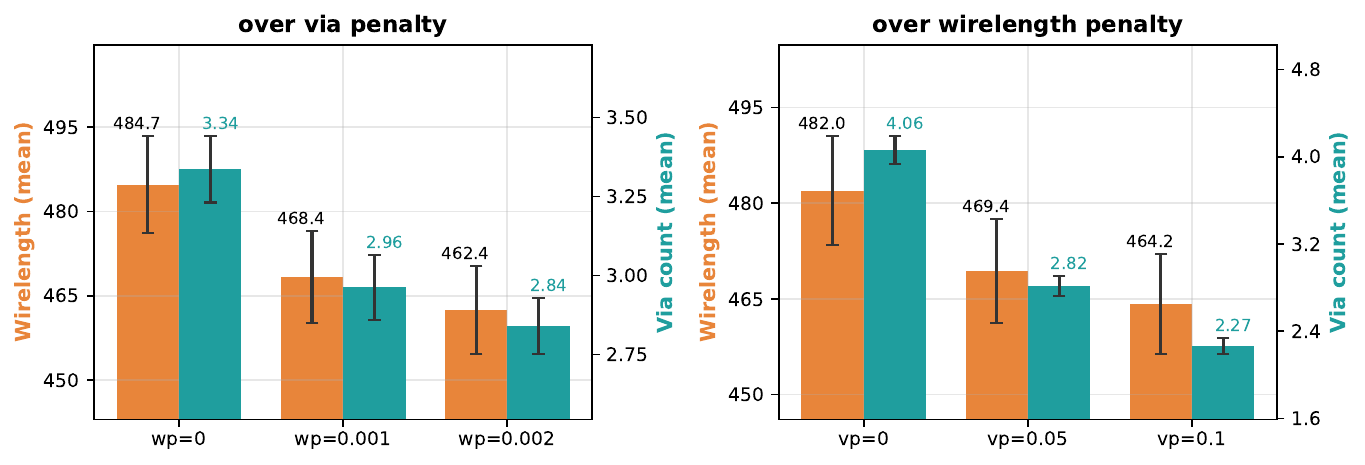}
  \caption{\textbf{Reward weight sweep ($3{\times}3$ factorial).} Bars show marginal means with error bars.
Wirelength: left axis (orange); via count: right axis (teal).}
  \label{fig:rq4-factorial}
  \Description{\textbf{Reward weight sweep ($3{\times}3$ factorial).} Bars show marginal means with error bars.
Wirelength: left axis (orange); via count: right axis (teal).}
\end{figure}

\paragraph{Controllability through penalty weights.}
We test whether the policy responds to fine-grained routing objectives by sweeping the wirelength weight $\lambda_w \in \{0, 0.001, 0.002\}$ and the via weight $\lambda_v \in \{0, 0.05, 0.1\}$ in a $3{\times}3$ factorial design (\cref{fig:rq4-factorial}).
We analyze the full factorial results by aggregating them along each axis.
Averaged over all via-weight settings, increasing $\lambda_w$ consistently reduces wirelength from $484.8$ mm to $468.4$ mm and $462.4$ mm.
Likewise, averaged over all wirelength-weight settings, increasing $\lambda_v$ reduces the via count from $4.06$ to $2.82$ and $2.27$.
Although wirelength and via count are coupled in PCB routing, these marginal trends show that each penalty acts as a consistent optimization signal for its associated cost.
Overall, the policy adapts its routing behavior to the specified objective rather than collapsing to a single operating point.

\section{Related Work}\label{sec:related}

\paragraph{Rule-based PCB routers.}
Automatic PCB routing has a long algorithmic history, spanning grid-based maze search since Lee's algorithm~\cite{lee1961maze} and gridless techniques~\cite{sherwani1999vlsi}, with rip-up and reroute as the central iterative heuristic~\cite{dees1982ripup,linsker1984ripup}.
Academic refinements continue this line~\cite{lin2021pcb,chen2023triangular,he2024padrouting}, and rule-based routers remain central in practice, from the open-source routers evaluated here~\cite{orthoroute,freerouting,kicadroutingtools} to commercial EDA suites~\cite{cadence_allegro,altium_designer}.
Since their prior knowledge is encoded in manually designed rules and heuristics, performance depends on objective- and board-specific tuning.

\paragraph{RL for PCB routing.}
Existing RL work narrows the agent's action interface in one of two ways.
The first family casts routing as cell-by-cell movement on a discretized grid~\cite{liao2020globalrouting,circuitrouting2022mcts,jumanjibench,sable}; at the grid resolution production boards require, the search space grows impractically large.
The second family routes only indirectly: the agent selects high-level options for a fixed auto-router, net order~\cite{wang2020attention,lin2023xroute,netordering2025transformer,chiang2026eswa}, routing pattern~\cite{detailedrouting2023ispd}, or fanout locations~\cite{fanoutnet2023drl}, while the router draws all geometry, capping quality at the router's ceiling.
In neither family does the agent draw board geometry through the engine's native operations.
\ours{} instead exposes the engine's native routing operations as the agent's action space~\cite{kicad,kicad_ipc}, and derives the learning signal from the engine's design-rule check rather than from a hand-crafted geometric proxy.

\paragraph{LLM agents for EDA.}
LLM agents that pair reasoning with tool calls~\cite{react} are beginning to reach EDA. Within PCB, PCB-Bench~\cite{pcbbench2026placement} evaluates an LLM's board interpretation but not its routing ability.
Other work either proceeds in a single open-loop pass or uses the engine as a checker.
The former generates EDA scripts or hardware description language (HDL) code~\cite{liu2023chipnemo,liu2023verilogeval,he2024chateda,openroad2024assistant}.
The latter revises the HDL on compiler and simulation errors~\cite{tsai2023rtlfixer,meic2024debug,thakur2024autochip,ho2024verilogcoder} or a schematic on electrical-rule violations~\cite{pcbschemagen2026llm}.
In \ours{}, the agent acts through the engine's native operations in a closed loop, and the engine checks every intermediate board rather than only the completed artifact.

\paragraph{Agent benchmarks beyond EDA.}
Outside EDA, LLM-based design generation has been studied most actively in 3D CAD.
These methods generate CAD operation sequences or parametric code from text and images~\cite{wu2021deepcad,llm4cad2025multimodal,cadgpt2025spatial,cadcoder2025cot}.
Benchmarks score reconstruction of a labeled target 3D shape, a softer standard than EDA's design rules~\cite{text2cadbench2026}.
Agent benchmarks in general software domains score code patches and tool-use episodes, with little geometry at stake~\cite{swebench,yang2024sweagent,toolbench,patil2025bfcl,yao2024taubench}.
Computer-use environments and GUI agents operate software through the screen as a human does~\cite{xie2024osworld,cheng2024seeclick,hong2023cogagent}.
If applied to an EDA tool, they would come closest to an engineer's workflow.
Acting through the screen costs them threefold: (i) pixels only approximate the exact board geometry, turning routing into a POMDP; (ii) precise cursor control remains unreliable for current agents; (iii) rendering every step slows both training and evaluation.
\ours{} bypasses the screen altogether, operating on the exact board state.


\section{Conclusion}\label{sec:conclusion}

We present \ours{}, a Gym environment that wraps the open-source \kicad{} EDA engine and formulates PCB routing as an MDP without surrogate abstraction. Two wrappers let RL and LLM agents train and act in the same environment.
Built on \ours{}, \oursbench{} pairs two synthetic board generators (D1, D2) with 679 curated open-source boards reserved for zero-shot evaluation (D3), benchmarking geometric feasibility reasoning under a method-agnostic evaluation protocol.
Empirically, the \kicad{}-API action interface outperforms both grid-action RL and engine-free LLM generation. A compact PPO policy trained on synthetic boards transfers zero-shot to open-source boards, approaching a rule-based router refined over decades.
We open-source \ours{} and \oursbench{} as an engine-grounded foundation for learning-based EDA research.


\bibliographystyle{ACM-Reference-Format}
\bibliography{references}

\onecolumn
\appendix
\section{Notation Glossary}\label{sec:notations}
    
    
\begin{table}[htbp]
\centering
\small
\setlength{\tabcolsep}{6pt}
\renewcommand{\arraystretch}{1.08}
\caption{Notation.}
\label{tab:notation}
\begin{adjustbox}{max width=\linewidth}
\begin{tabular}{ll}
\toprule
Symbol & Description \\
\midrule
$n_{\mathrm{drv}}(\cdot)$ & Number of design rule violations (DRVs) \\
$n_{\mathrm{via}}(\cdot)$ & Number of vias \\
$\ell(\cdot)$ & Total wirelength \\
$\lambda_w, \lambda_v$ & Wirelength and via weights in Equation~\eqref{eq:opt} \\
$f_{\mathrm{drv}}(\cdot)$ & Concave DRV penalty in the potential $\Phi$, computed from the per-net violation decomposition of $s$ (Equation~\eqref{eq:fd-default}) \\
$\mathcal{R}$ & Set of board states \\
$s$ & Board state, represented as a nested dictionary \\
$s_t$ & Board state at step $t$; $t=T$ is the terminal step \\
$\Phi(\cdot)$ & Potential function \\
$r_T$ & Terminal reward, $\Phi(s_T)-\Phi(s_0)$ (the MDP reward) \\
$r_t$ & Per-step reward, $\Phi(s_{t+1})-\Phi(s_t)$ (shaping form) \\
$G$ & Grid size \\
\bottomrule
\end{tabular}
\end{adjustbox}
\end{table}

\section{System Architecture (\ours{})}\label{sec:pcbworld-arch}

This appendix distinguishes carefully among the routing method's decisions, Gym-env mediation, the wrapper that surfaces the 58 step-level APIs, and the C++ binding underneath.
Confusing these four layers is the most common source of misreading the state, action, and reward decomposition in Appendix~\ref{sec:action-apilist}, so we fix the layered picture here.
\cref{fig:pcbworld-arch} gives the agent--environment view of the stack and the RL signals connecting the two sides.

\begin{figure}[!h]
\centering
\includegraphics[width=\linewidth]{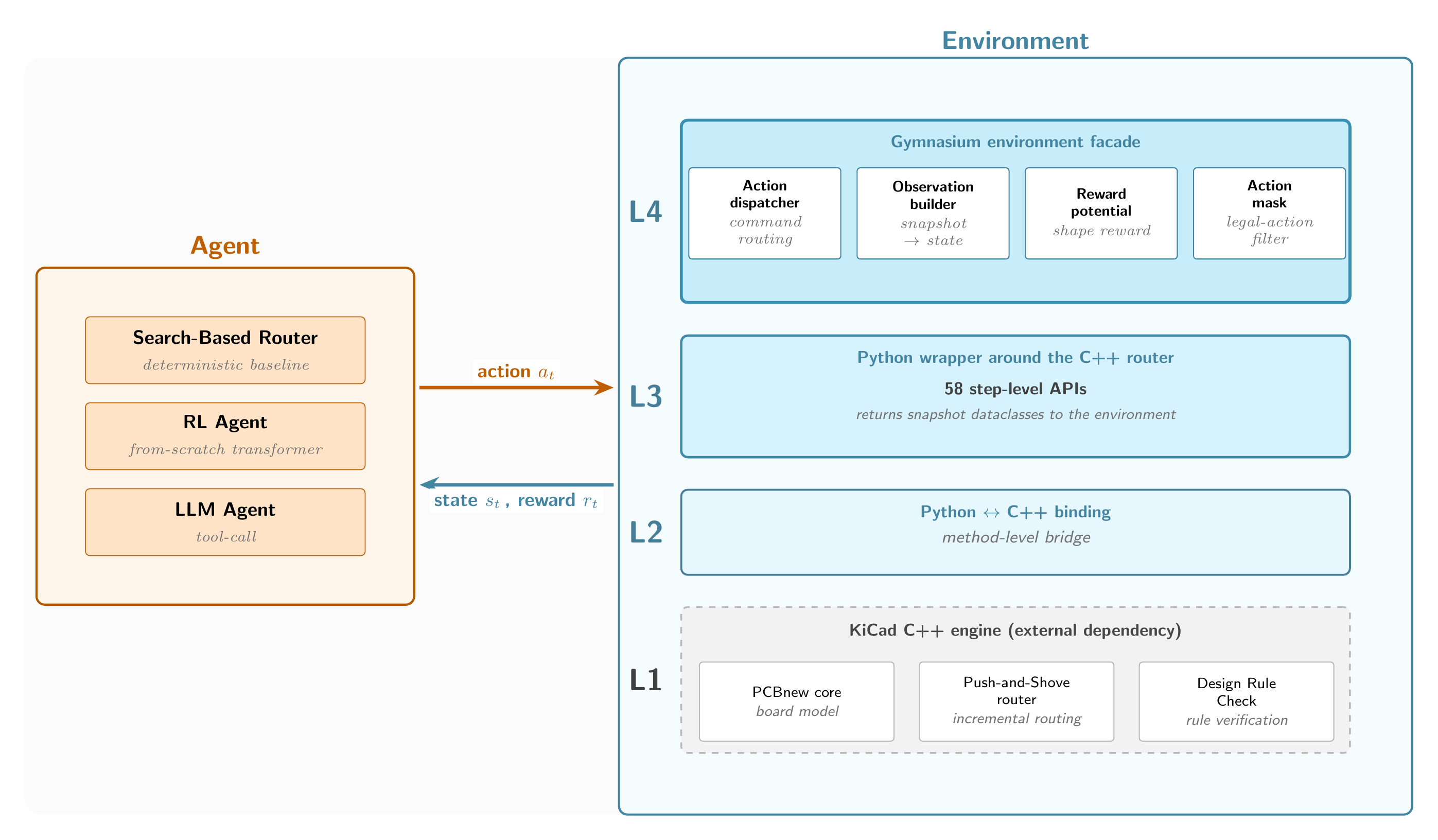}
\caption{\textbf{\ours{} architecture.} A reinforcement-learning view of the proposed environment.
Three classes of routing methods, a scripted rule-based baseline, an RL agent built on a from-scratch Transformer, and an LLM agent that issues tool calls, share a single Gymnasium environment through the standard action $a_t$ and the state $s_t$, reward $r_t$ signals.
Inside the environment, our contribution spans L4 (the Gymnasium environment facade with four sub-modules: action dispatcher, observation builder, reward potential, and action mask), L3 (the Python wrapper exposing 58 step-level APIs), and L2 (the Python$\leftrightarrow$C++ binding); layer-box heights encode relative scope, while sub-boxes detail the internal components.
L1 is \kicad{}'s unmodified C++ engine, namely PCBnew core, Push-and-Shove router, and Design Rule Check, drawn with a dashed border as an external dependency.}
\label{fig:pcbworld-arch}
  \Description{\textbf{\ours{} architecture.} A reinforcement-learning view of the proposed environment.
Three classes of routing methods, a scripted rule-based baseline, an RL agent built on a from-scratch Transformer, and an LLM agent that issues tool calls, share a single Gymnasium environment through the standard action $a_t$ and the state $s_t$, reward $r_t$ signals.
Inside the environment, our contribution spans L4 (the Gymnasium environment facade with four sub-modules: action dispatcher, observation builder, reward potential, and action mask), L3 (the Python wrapper exposing 58 step-level APIs), and L2 (the Python$\leftrightarrow$C++ binding); layer-box heights encode relative scope, while sub-boxes detail the internal components.
L1 is \kicad{}'s unmodified C++ engine, namely PCBnew core, Push-and-Shove router, and Design Rule Check, drawn with a dashed border as an external dependency.}
\end{figure}

\subsection{Layering the stack}
At the top, three classes of routing methods share an identical observation and action interface: a deterministic scripted rule-based baseline, an RL agent built on a from-scratch Transformer, and an LLM agent that issues tool calls.
They all enter the stack through \texttt{KiCadHLEnv} at L4, the Gymnasium environment facade that owns four sub-modules: an action dispatcher for command routing, an observation builder, a reward potential composer, and an action mask.
The dispatcher is the only L4 sub-module that mutates state; the other three are read paths.
L4 holds a single \texttt{KiCadEngine} instance at L3, the Python wrapper around the C++ routing engine that exposes the 58 step-level APIs the rest of the appendices catalog.
L3 is the only Python call site that reaches the Python $\leftrightarrow$ C++ binding at L2, behind which sits the unmodified \kicad{} C++ engine at L1.

\subsection{Building observations and rewards}
This subsection and the next together specify the L3$\leftrightarrow$L4 signal exchange that synthesizes an MDP interface from the unmodified engine, which is what makes the stack an RL environment rather than a Python SDK over the 58 APIs.
Every read crossing back from L3 into Python is packaged into one of three dataclasses.
\texttt{BoardSnapshot} is the geometry bundle that the observation builder consumes.
\texttt{RewardSnapshot} is the unrouted, wirelength, and DRC bundle that the reward potential composer folds into $\Phi(s)$.
\texttt{RoutingSessionState} is the routing FSM bundle that lets the action mask and the action dispatcher decide which actions are admissible.
Because these three are the only read channels, the L3$\leftrightarrow$L4 coupling reduces to which engine method populates which dataclass field, which is exactly the reading the dual-tag analysis in Appendix~\ref{sec:apilist-misc} formalizes.

\subsection{Dispatching actions}
The three method classes do not call the 58 step-level APIs directly.
They emit a single high-level action through the action dispatcher at L4, which translates that action into a small subset of L3 mutators.
Appendix~\ref{sec:action-apilist} catalogs the 58 APIs and their MDP components, and the high-level action surface exposed to the methods is documented in the main text under \S\ref{sec:env-action}.
The data contract above fixes the state, reward, and admissibility channels that a method reads, while this high-level action surface fixes the action space it emits, completing the MDP interface that L4 exposes upward.

\section{Mapping the API to the MDP}\label{sec:action-apilist}

The agent reaches \kicad{}'s PNS router through a thin Python wrapper whose public surface exposes 58 step-level methods, partitioned by the wrapper source into 14 source-code groups and mapped to 25 Action, 25 State, 3 Reward, and 5 Utility components of the MDP.
These role totals reconcile with the main-text partition of \S\ref{sec:env-engine}: the 25 State and 3 Reward methods are the 28 DRC and board-state wrappers, and the 25 Action and 5 Utility methods are the 14 core routing APIs together with the 16 auxiliary utilities.
\cref{tab:api-rolegroup} catalogs every method by MDP component and source-code group, and is the central artifact of this appendix.
Four wrapper-only helpers are excluded from the count: \texttt{close} is episode teardown, \texttt{get\_native} returns a raw C++ handle for renderers, and \texttt{get\_drc\_result}, \texttt{get\_board\_snapshot} are Python-side dataclass aggregators whose underlying getters surface individually below.
\texttt{get\_reward\_snapshot} remains counted because it performs its own engine pass rather than re-bundling listed getters, as detailed in Appendix~\ref{sec:apilist-misc}.
Appendix~\ref{sec:pcbworld-arch} situates this wrapper as the L3 layer of the overall \ours{} stack.

\begin{table*}[!t]
\centering\small
\setlength{\tabcolsep}{5pt}
\renewcommand{\arraystretch}{1.10}
\begin{tabular}{@{}p{2.4cm}>{\raggedleft\arraybackslash}p{0.35cm}>{\footnotesize\raggedright\arraybackslash}p{9.4cm}@{}}
\toprule
Group & $n$ & Methods \\
\midrule
\multicolumn{3}{@{}l@{}}{\textit{\textbf{Action}: drive PCB routing through copper placement, parameter changes, and rework (25 methods).}} \\
\midrule
Configuration              & 6 & \texttt{set\_routing\_mode()}, \texttt{set\_corner\_mode()}, \texttt{set\_track\_width()}, \texttt{set\_via\_diameter()}, \texttt{set\_via\_drill()}, \texttt{reset\_via\_mode()} \\
Routing                    & 9 & \texttt{start\_route()}, \texttt{move()}, \texttt{fix\_route()}, \texttt{cancel\_route()}, \texttt{finish()}, \texttt{undo\_last\_segment()}, \texttt{flip\_posture()}, \texttt{toggle\_via()}, \texttt{switch\_layer()} \\
Dragging                   & 3 & \texttt{start\_drag()}, \texttt{fix\_drag()}, \texttt{cancel\_drag()} \\
Rework                     & 4 & \texttt{delete\_track\_by\_index()}, \texttt{delete\_track\_near()}, \texttt{delete\_via\_by\_index()}, \texttt{delete\_via\_near()} \\
Refresh, \emph{a/r}        & 1 & \texttt{run\_drc()} \\
Refresh, \emph{a/u}        & 2 & \texttt{build\_connectivity()}, \texttt{set\_design\_rules()} \\
\midrule
\multicolumn{3}{@{}l@{}}{\textit{\textbf{State}: read the current PCB layout, routing session FSM, and design rules (25 methods).}} \\
\midrule
Board meta                 & 3 & \texttt{get\_board\_bbox()}, \texttt{get\_board\_net\_count()}, \texttt{get\_copper\_layer\_count()} \\
Elements                   & 7 & \texttt{get\_tracks()}, \texttt{get\_vias()}, \texttt{get\_pads()}, \texttt{get\_points()}, \texttt{get\_ratsnest()}, \texttt{get\_board\_outline()}, \texttt{get\_net\_names()} \\
Elements, \emph{s/r}       & 3 & \texttt{get\_track\_count()}, \texttt{get\_via\_count()}, \texttt{get\_unrouted\_count()} \\
Session                    & 8 & \texttt{get\_router\_state\_code()}, \texttt{is\_routing()}, \texttt{is\_dragging()}, \texttt{is\_placing\_via()}, \texttt{get\_current\_layer()}, \texttt{get\_route\_head()}, \texttt{get\_current\_net\_code()}, \texttt{get\_routing\_target()} \\
Design rules               & 2 & \texttt{get\_design\_rules()}, \texttt{get\_netclass\_for\_net()} \\
Snapshots                  & 2 & \texttt{get\_board\_meta()}, \texttt{get\_routing\_session\_state()} \\
\midrule
\multicolumn{3}{@{}l@{}}{\textit{\textbf{Reward}: score routing quality through DRC violations and routing progress (3 methods).}} \\
\midrule
Diagnostics                & 2 & \texttt{get\_drc\_violation\_count()}, \texttt{get\_drc\_violations()} \\
Diagnostics, \emph{s/r}    & 1 & \texttt{get\_reward\_snapshot()} \\
\midrule
\multicolumn{3}{@{}l@{}}{\textit{\textbf{Utility}: save the routed PCB and reset engine caches between episodes (5 methods).}} \\
\midrule
DRC cache                  & 1 & \texttt{clear\_drc\_cache()} \\
I/O                        & 1 & \texttt{save()} \\
Provenance                 & 3 & \texttt{get\_project\_path()}, \texttt{was\_project\_loaded\_from\_file()}, \texttt{was\_legacy\_design\_settings\_loaded()} \\
\bottomrule
\end{tabular}
\caption{\textbf{Step-level API inventory.} The 58 APIs map to 25 Action, 25 State, 3 Reward, and 5 Utility methods, each further refined into one of 14 source-code groups.
Dual-component rows append the secondary tag in italic after the group name, for example \emph{a/r}; these four rows cover the seven methods whose component overlap is forced by the underlying C++ binding.
The Refresh group covers engine-cache refresh operations called at step or episode boundaries, which is why its rows are dual-tagged rather than purely Action.}
\label{tab:api-rolegroup}
\end{table*}

For a step-level method $m$ at L3, component membership is fixed by which L4 call-sites reach $m$, not by what the method returns.
The two subsections below organize the 58 APIs of \cref{tab:api-rolegroup}: the Routing MDP collects the per-step State, Reward, Action, and transition components, while the miscellaneous group covers episode-level Utility methods and seven dual-tagged overlaps.

\subsection{Routing MDP}\label{sec:apilist-routing-mdp}
\paragraph{State.} Methods that feed the observation builder, either directly or through the \texttt{BoardSnapshot} aggregator.
The aggregator bundles seven read-only getters spanning board geometry, ratsnest, and metadata into a single dataclass, but is excluded from the 58-method inventory because its underlying getters appear individually.
The Board meta, Elements, Session, Design rules, and Snapshots groups are State.

\paragraph{Reward.} Methods that populate fields of the \texttt{RewardSnapshot} dataclass that the potential composer folds into $\Phi(s)$ in \S\ref{sec:env-reward}.
The Diagnostics group exhausts this tag, with the \texttt{get\_reward\_snapshot} aggregator joining the two raw DRC counters.

\paragraph{Action.} Methods that the per-step dispatcher invokes to mutate board or router state, covering reset cleanup, per-step configuration, and within-episode action handling.
The Configuration, Routing, Dragging, and Rework groups fall here.

\paragraph{State transition.} The transition kernel $P(s_{t+1} \mid s_t, a_t)$ is realized entirely by the unmodified C++ engine.
An Action method submits the requested mutation, the engine applies, modifies, or rejects it according to its routing and DRC logic, and the L4 facade then harvests $s_{t+1}$ from the resulting \texttt{BoardSnapshot} and \texttt{RoutingSessionState} while \texttt{RewardSnapshot} supplies $r_t$ on the same C++ pass.
Inadmissible actions are pre-filtered by the action mask reading \texttt{RoutingSessionState}, so the engine's reject path is defensive rather than routine, and no Python code computes the transition itself.

\subsection{Miscellaneous}\label{sec:apilist-misc}
\paragraph{Utility.} Methods reached only from reset or end-of-episode bookkeeping, never from the per-step decision path.
The DRC cache, I/O, and Provenance groups are pure Utility.

\paragraph{Dual-tagged exceptions.} The component assignment above admits seven dual-tagged exceptions, all forced by the underlying C++ binding rather than by taxonomic ambiguity.
The three element-count getters and \texttt{get\_reward\_snapshot} share an s/r tag because one C++ pass populates both a State and a Reward field.
\texttt{run\_drc} carries a/r because its single invocation mutates the DRC cache and also refreshes the violation list the reward consumes.
\texttt{build\_connectivity} and \texttt{set\_design\_rules} carry a/u because they mutate engine caches only at episode boundaries and never commit copper.
With these seven exceptions, 51 of 58 methods carry a single tag, so the per-component totals of 25 Action, 26 State, 7 Reward, and 7 Utility sum to 65 and exceed the unique count of 58 by exactly the dual-tagged rows.

\section{Visual Demonstration for Routing APIs}\label{sec:action-api_example}

This appendix complements the API table of Appendix~\ref{sec:action-apilist} with a visual reference for the 22 routing-related entries.
Every figure shares a single template: a \emph{scenario panel} on the left fixes the starting board and overlays a dashed plan for the intended route, an \emph{action sequence} in the middle prints the literal \texttt{kicad\_engine.py} call sequence, and \emph{variant panels} on the right show the resulting copper under different parameter choices.
Net colors are computed deterministically from the engine's \texttt{net\_code} so each net keeps a consistent color across figures: NET1 stays red, NET2 stays blue, and NET3 is rendered as the dashed plan.
Layer is color-coded with \emph{Top} in red and \emph{Btm} in blue, following the \kicad{} GUI convention.
Coordinates use the \kicad{} board frame, with the origin at the top-left and the $y$-axis pointing downward, and all dimensions are given in millimeters.
The visualization covers twenty-one of the twenty-two APIs in five thematic groups.
The remaining one, \texttt{undo\_last\_segment}, only mutates the in-flight head and never modifies committed copper, so a static figure cannot show its effect.
Its semantics are verified by the unit test \texttt{test\_undo\_does\_not\_commit\_tracks}.

\paragraph{Routing primitives.}
A routing session is bracketed by \texttt{start\_route} and one of three completion calls: \texttt{fix\_route}, \texttt{cancel\_route}, or \texttt{finish}.
Within the session, \texttt{move} re-positions the head while only \texttt{fix\_route} commits a segment to copper.
\cref{fig:appx-routing-session} visualizes this commit-versus-no-commit boundary across a four-step session, and \cref{fig:appx-routing-basic} contrasts the three completion calls on a three-pad scenario.
Two further session-time controls shape the geometry of the committed trace: \texttt{flip\_posture} in \cref{fig:appx-routing-shape} mirrors the L-shape orientation chosen for asymmetric endpoints, and the layer-transition pair \texttt{toggle\_via}/\texttt{switch\_layer} in \cref{fig:appx-via-layer} either drops a stand-alone via or inserts one automatically when the active layer changes.

\begin{figure}[!htbp]
  \centering
  \includegraphics[height=3.1cm]{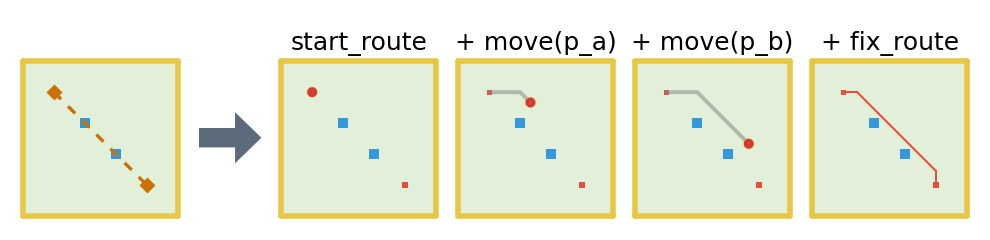}
  \caption{\textbf{Virtual move versus physical commit.} Within a routing session, \texttt{move} relocates the virtual head that reflects the designer's intent and draws only a transient hint-line preview without modifying the board.
\texttt{fix\_route} commits the current virtual path as a physical copper object in the database and converts the preview into a permanent trace.}
  \label{fig:appx-routing-session}
  \Description{\textbf{Virtual move versus physical commit.} Within a routing session, \texttt{move} relocates the virtual head that reflects the designer's intent and draws only a transient hint-line preview without modifying the board.
\texttt{fix\_route} commits the current virtual path as a physical copper object in the database and converts the preview into a permanent trace.}
\end{figure}

\begin{figure}[!htbp]
  \centering
  \includegraphics[height=3.1cm]{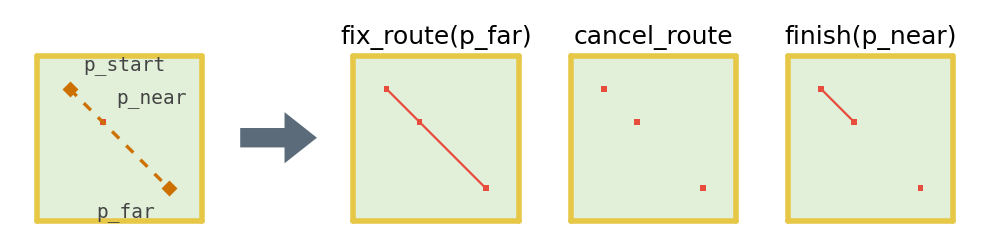}
  \caption{\textbf{Session termination modes.} Three distinct primitives close an open routing session.
\texttt{fix\_route} commits the trace at the designer-specified coordinate and ends the session.
\texttt{cancel\_route} discards every virtual change in the current session and restores the board to its pre-session state.
\texttt{finish} invokes the PNS router to complete the route to the Euclidean-nearest unconnected pad on the current layer and ends the session.}
  \label{fig:appx-routing-basic}
  \Description{\textbf{Session termination modes.} Three distinct primitives close an open routing session.
\texttt{fix\_route} commits the trace at the designer-specified coordinate and ends the session.
\texttt{cancel\_route} discards every virtual change in the current session and restores the board to its pre-session state.
\texttt{finish} invokes the PNS router to complete the route to the Euclidean-nearest unconnected pad on the current layer and ends the session.}
\end{figure}

\begin{figure}[!htbp]
  \centering
  \includegraphics[height=3.1cm]{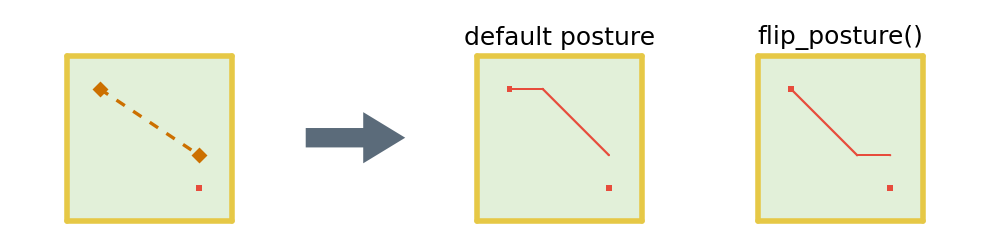}
  \caption{\textbf{Corner posture control.} When two endpoints are joined by an L-shape, the PNS router must choose between leaving the start pad horizontally first or vertically first.
\texttt{flip\_posture} reverses this corner direction and allows the routing efficiency to be tuned dynamically within the available design space.}
  \label{fig:appx-routing-shape}
  \Description{\textbf{Corner posture control.} When two endpoints are joined by an L-shape, the PNS router must choose between leaving the start pad horizontally first or vertically first.
\texttt{flip\_posture} reverses this corner direction and allows the routing efficiency to be tuned dynamically within the available design space.}
\end{figure}

\begin{figure}[!htbp]
  \centering
  \includegraphics[height=3.1cm]{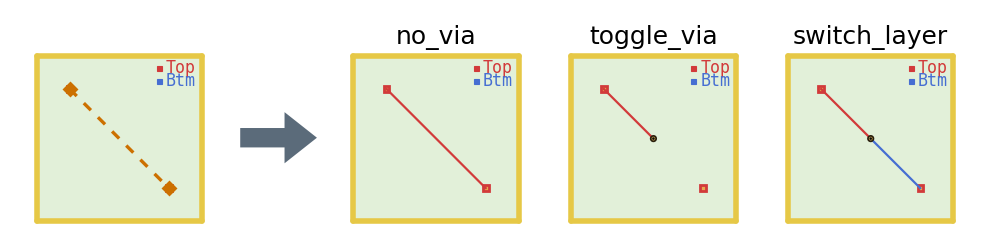}
  \caption{\textbf{Via placement and layer transition.} In a multilayer board, electrical connections between different layers are mediated by vias.
\texttt{toggle\_via} drops a vertical connection at the current location, while \texttt{switch\_layer} changes the active routing layer and automatically inserts a transition via to connect the two layers.}
  \label{fig:appx-via-layer}
  \Description{\textbf{Via placement and layer transition.} In a multilayer board, electrical connections between different layers are mediated by vias.
\texttt{toggle\_via} drops a vertical connection at the current location, while \texttt{switch\_layer} changes the active routing layer and automatically inserts a transition via to connect the two layers.}
\end{figure}

\paragraph{Routing-mode controls.}
Two session-level switches choose how the PNS router treats existing copper and how it draws corners.
\cref{fig:appx-routing-modes} contrasts the three strategies of \texttt{set\_routing\_mode}, namely \texttt{mark\_as\_obstacles}, \texttt{push\_n\_shove}, and \texttt{walkaround}, on a board pre-routed with two obstacle nets.
\cref{fig:appx-corner-modes} contrasts the two settings of \texttt{set\_corner\_mode}, $45^\circ$ mitering against $90^\circ$ rectilinear corners, on identical diagonal endpoints.
The default \texttt{MITERED\_45} mode is the recommended PCB practice, while the \texttt{MITERED\_90} mode is exposed for compatibility with legacy EDA tools but is generally avoided in modern design.

\begin{figure}[!htbp]
  \centering
  \includegraphics[height=3.1cm]{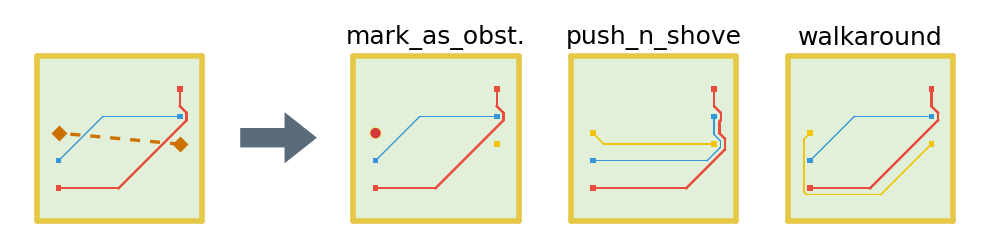}
  \caption{\textbf{Obstacle interaction policies.} \texttt{set\_routing\_mode} chooses one of three policies for interacting with existing copper.
\texttt{mark\_as\_obstacles} performs collision detection and aborts trace generation when a clearance constraint would be violated.
\texttt{push\_n\_shove} pushes neighboring tracks aside to dynamically clear space for the new path.
\texttt{walkaround} preserves the existing copper geometry and searches for an optimal detour around it.}
  \label{fig:appx-routing-modes}
  \Description{\textbf{Obstacle interaction policies.} \texttt{set\_routing\_mode} chooses one of three policies for interacting with existing copper.
\texttt{mark\_as\_obstacles} performs collision detection and aborts trace generation when a clearance constraint would be violated.
\texttt{push\_n\_shove} pushes neighboring tracks aside to dynamically clear space for the new path.
\texttt{walkaround} preserves the existing copper geometry and searches for an optimal detour around it.}
\end{figure}

\begin{figure}[!htbp]
  \centering
  \includegraphics[height=3.1cm]{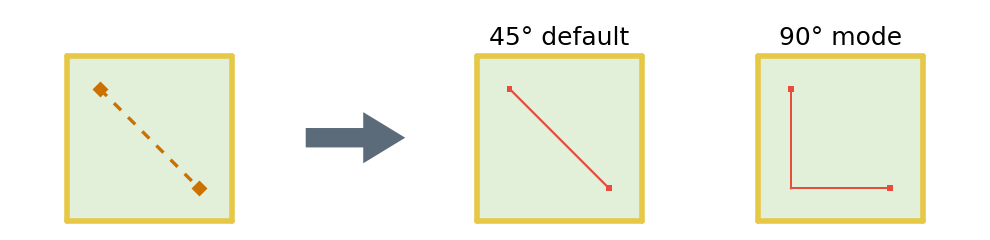}
  \caption{\textbf{Trace corner geometry.} \texttt{set\_corner\_mode} controls the angular format at trace corners.
The industry-standard $45^\circ$ miter, \texttt{MITERED\_45}, is optimized to minimize signal loss and avoid manufacturing defects.
The $90^\circ$ rectilinear corner, \texttt{MITERED\_90}, is reserved for cases that require special geometric alignment.}
  \label{fig:appx-corner-modes}
  \Description{\textbf{Trace corner geometry.} \texttt{set\_corner\_mode} controls the angular format at trace corners.
The industry-standard $45^\circ$ miter, \texttt{MITERED\_45}, is optimized to minimize signal loss and avoid manufacturing defects.
The $90^\circ$ rectilinear corner, \texttt{MITERED\_90}, is reserved for cases that require special geometric alignment.}
\end{figure}

\paragraph{Track and via geometry.}
Width and via geometry are exposed as session-time setters that override the design-rule defaults for the next committed segment or via.
\cref{fig:appx-track-width} sweeps \texttt{set\_track\_width} across three target widths on identical endpoints, the same control used in practice to differentiate power, clock, and signal nets.
\cref{fig:appx-via-geometry} factorizes via geometry into three independent calls, \texttt{set\_via\_diameter}, \texttt{set\_via\_drill}, and \texttt{reset\_via\_mode}, with a final reset variant verifying that the per-call override is not sticky.

\begin{figure}[!htbp]
  \centering
  \includegraphics[height=3.1cm]{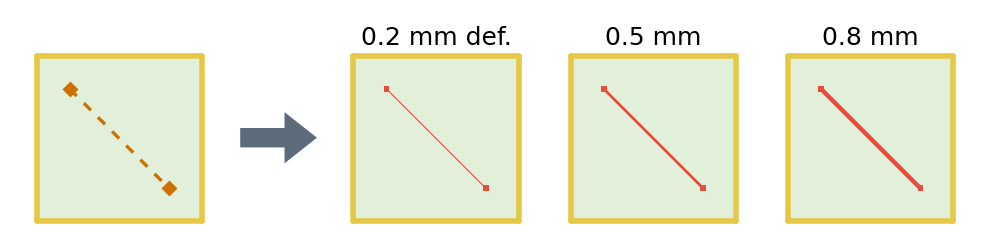}
  \caption{\textbf{Trace width control.} \texttt{set\_track\_width} explicitly sets the physical conductor width.
Width is the key parameter that determines characteristic impedance and current-carrying capacity, distinguishing wide power traces that carry large currents from thin signal traces that carry fine signals.}
  \label{fig:appx-track-width}
  \Description{\textbf{Trace width control.} \texttt{set\_track\_width} explicitly sets the physical conductor width.
Width is the key parameter that determines characteristic impedance and current-carrying capacity, distinguishing wide power traces that carry large currents from thin signal traces that carry fine signals.}
\end{figure}

\begin{figure}[!htbp]
  \centering
  \includegraphics[height=3.1cm]{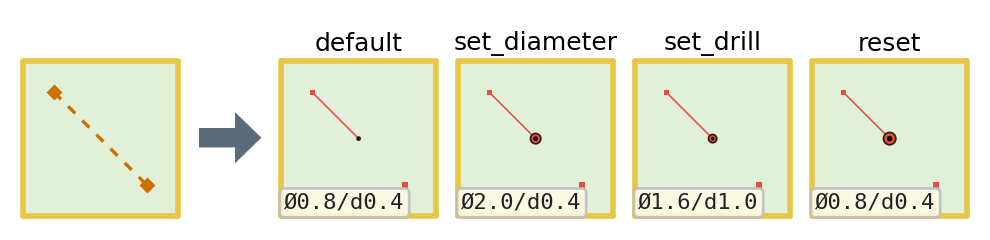}
  \caption{\textbf{Via geometry and reset.} The outer diameter and inner drill of a via are adjusted individually and independently of the design rules.
\texttt{reset\_via\_mode} clears such temporary overrides and restores the system's default design standard, preserving design consistency.}
  \label{fig:appx-via-geometry}
  \Description{\textbf{Via geometry and reset.} The outer diameter and inner drill of a via are adjusted individually and independently of the design rules.
\texttt{reset\_via\_mode} clears such temporary overrides and restores the system's default design standard, preserving design consistency.}
\end{figure}

\paragraph{Drag-to-modify.}
Once a track is committed, the engine permits non-destructive geometric modification through a drag handle on its midpoint.
\cref{fig:appx-drag} shows the three primitives \texttt{start\_drag}, \texttt{fix\_drag}, and \texttt{cancel\_drag} on a single pre-routed diagonal, contrasting the committed and aborted shapes side by side.
The drag family complements the deletion primitives below by providing an in-place modification path that does not require removal followed by re-routing.

\begin{figure}[!htbp]
  \centering
  \includegraphics[height=3.1cm]{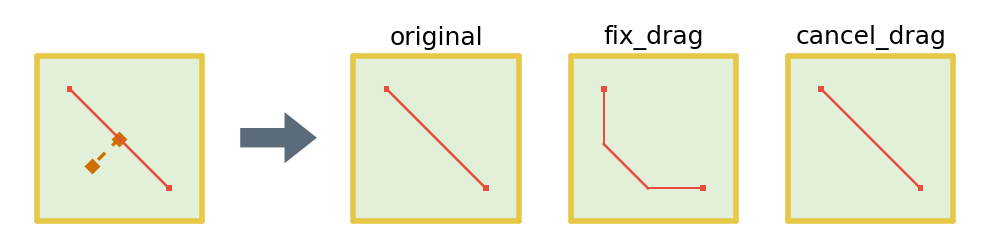}
  \caption{\textbf{Trace drag-to-modify.} Pulling the midpoint of an already committed trace allows its geometry to be reshaped flexibly.
This is a non-destructive editing path used when local route optimization is needed without disturbing the existing connectivity.}
  \label{fig:appx-drag}
  \Description{\textbf{Trace drag-to-modify.} Pulling the midpoint of an already committed trace allows its geometry to be reshaped flexibly.
This is a non-destructive editing path used when local route optimization is needed without disturbing the existing connectivity.}
\end{figure}

\paragraph{Track and via removal.}
Tracks and vias can be removed either by integer position in the engine's internal list or by spatial coordinate within a tolerance.
\cref{fig:appx-delete-tracks} contrasts \texttt{delete\_track\_by\_index} and \texttt{delete\_track\_near} on a chained three-segment L, and \cref{fig:appx-delete-vias} contrasts \texttt{delete\_via\_by\_index} and \texttt{delete\_via\_near} on a multi-via trace.
The two addressing modes are functionally equivalent at the engine level but differ in agent ergonomics: index addressing is convenient for replay or rollback patterns, while coordinate addressing is convenient when the agent identifies the target by its board coordinates.

\begin{figure}[!htbp]
  \centering
  \includegraphics[height=3.1cm]{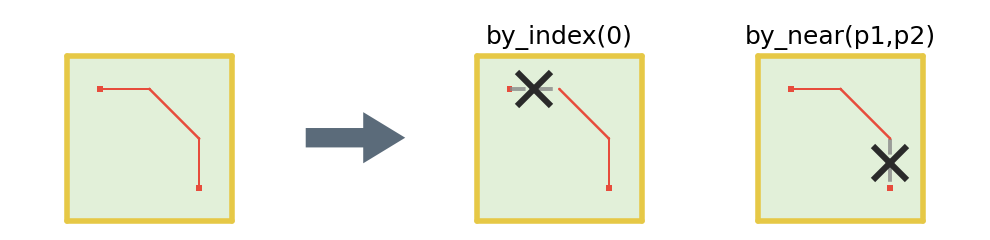}
  \caption{\textbf{Track deletion methods.} A track object can be identified and removed by two complementary logical approaches.
\texttt{delete\_track\_by\_index} addresses the engine's internal list by integer position to identify the target precisely.
\texttt{delete\_track\_near} addresses the same target by physical coordinate, providing an intuitive interface for an agent that reasons in board coordinates.}
  \label{fig:appx-delete-tracks}
  \Description{\textbf{Track deletion methods.} A track object can be identified and removed by two complementary logical approaches.
\texttt{delete\_track\_by\_index} addresses the engine's internal list by integer position to identify the target precisely.
\texttt{delete\_track\_near} addresses the same target by physical coordinate, providing an intuitive interface for an agent that reasons in board coordinates.}
\end{figure}

\begin{figure}[!htbp]
  \centering
  \includegraphics[height=3.1cm]{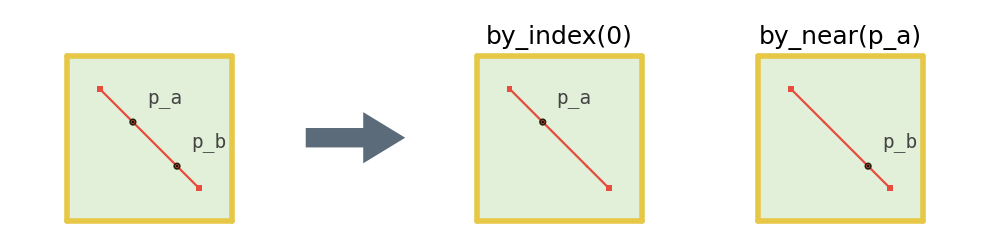}
  \caption{\textbf{Via deletion methods.} By the same logic as track removal, the vertical connection object is selectively removed either by engine index or by spatial coordinate.}
  \label{fig:appx-delete-vias}
  \Description{\textbf{Via deletion methods.} By the same logic as track removal, the vertical connection object is selectively removed either by engine index or by spatial coordinate.}
\end{figure}

\section{State-dictionary schema}\label{sec:state-schema}

\newcommand{\yes}{\textcolor{green!50!black}{\checkmark}}
\newcommand{\no}{\textcolor{red!70!black}{\textbf{\textsf{X}}}}
\newcommand{\partyes}{\textcolor{orange!80!black}{$\circ$}}

\begin{table}[ht]
\centering
\caption{Field-by-field schema across the projection layers.
\yes\ = present, \partyes\ = present in reduced form, \no\ = absent.}
\label{tab:schema}
\begin{adjustbox}{max width=\linewidth}
\begin{tabular}{@{}llcc>{\centering\arraybackslash}p{2.4cm}>{\centering\arraybackslash}p{3.2cm}@{}}
\toprule
\textbf{Field} & \textbf{\texttt{.kicad\_pcb} origin} & \textbf{Engine} & \textbf{JSON} & \textbf{RL} & \textbf{LLM}\\
\midrule
\multicolumn{6}{@{}l}{\emph{Geometry}}\\
Board bbox / copper layer count   & derived from outline                 & \yes & \yes        & \yes        & \yes \\
Board outline (Edge.Cuts)         & \texttt{(gr\_line/arc/...)}          & \partyes\textsuperscript{a} & \yes        & \yes        & \partyes\textsuperscript{b} \\
Pad geometry (xy, w, h, layer)    & \texttt{(pad ...)}                   & \yes & \yes        & \yes        & \partyes\textsuperscript{c} \\
Pad metadata (name, ref, rotation, shape) & \texttt{(pad/footprint ...)} & \yes & \partyes\textsuperscript{d} & \no & \partyes\textsuperscript{d} \\
Track segment                     & \texttt{(segment ...)}               & \yes & \yes        & \yes        & \partyes\textsuperscript{b} \\
Via outer diameter / layer span   & \texttt{(via ...)}                   & \yes & \yes        & \yes        & \partyes\textsuperscript{e} \\
Via drill diameter                & \texttt{(via ...)}                   & \yes & \no         & \no         & \no \\
Zone / copper pour                & \texttt{(zone ...)}                  & \no  & \no         & \no         & \no \\
\midrule
\multicolumn{6}{@{}l}{\emph{Net / connectivity}}\\
Net membership                    & \texttt{(net N)}                     & \yes & \yes        & \yes\textsuperscript{f} & \yes \\
Unconnected / NPTH pads           & \texttt{net 0}, \texttt{np\_thru\_hole} & \yes & \yes     & \no         & \partyes\textsuperscript{g} \\
Ratsnest (unrouted edges)         & --- (computed)                       & \yes & \yes        & \yes        & \yes \\
\midrule
\multicolumn{6}{@{}l}{\emph{Design rules}}\\
Per-netclass distribution         & \texttt{(setup)} / \texttt{.kicad\_pro} & \yes & \no      & \no         & \no \\
BDS minima, presets               & \texttt{(setup ...)}                 & \yes & \no         & \no         & \no \\
Strictest effective constraints   & --- (max over BDS + classes)         & \yes & \yes        & \no         & \yes\textsuperscript{h} \\
\midrule
\multicolumn{6}{@{}l}{\emph{Router state}}\\
Head xy / layer / net / phase / mode & in-memory                         & \yes & \yes        & \yes        & \yes \\
\texttt{is\_dragging}, \texttt{state\_code}, \texttt{routing\_target} & in-memory & \yes & \yes        & \no         & \no \\
\texttt{step}, \texttt{step\_ratio}, prev-action & env-synthesized       & ---  & \yes        & \yes        & \partyes\textsuperscript{i} \\
\midrule
\multicolumn{6}{@{}l}{\emph{DRC}}\\
Top-$k$ violation list            & --- (\texttt{run\_drc})              & \yes & \yes (top-32) & \yes      & \no\textsuperscript{j} \\
Per-net counts, error/warning split & ---                                & \yes & info only   & \no         & \no \\
\midrule
\multicolumn{6}{@{}l}{\emph{Action interface}}\\
6-way action mask                 & env-synthesized                      & ---  & info        & logit mask  & valid-action listing + grammar \\
Coordinate emission               & ---                                  & ---  & ---         & discrete pointer pool & continuous text + regex grammar \\
Self-feedback (no-effect, streak) & ---                                  & ---  & \no         & \no         & \yes\textsuperscript{k} \\
\bottomrule
\end{tabular}
\end{adjustbox}

\footnotesize
\textsuperscript{a}\,curves linearized once at the C++ binding; \textsuperscript{b}\,segment / edge widths dropped from the textual prompt; \textsuperscript{c}\,sexpr keeps \texttt{xy+layer} only, xml keeps full geometry; \textsuperscript{d}\,shape only; \textsuperscript{e}\,diameter dropped from the textual prompt; \textsuperscript{f}\,via slot id, name dropped; \textsuperscript{g}\,NPTH only as obstacles, unconnected pads omitted; \textsuperscript{h}\,emitted as \texttt{board\_constraints}, omitting unset fields; \textsuperscript{i}\,history rendered as multi-turn text; \textsuperscript{j}\,reward signal only, not echoed in the prompt; \textsuperscript{k}\,LLM-specific (no parameter-side memory).
\end{table}

The environment projects the raw \kicad{} \texttt{.kicad\_pcb} S-expression through four progressively narrower views: the C++ \emph{engine} (\texttt{KiCadEngine}, the authoritative live state), a JSON \emph{Gym observation} (\texttt{KiCadHLEnv}), an \emph{RL token stream} (\texttt{BatchedStateTokenizer}), and an \emph{LLM text prompt} (S-expression / XML rendering).
\cref{tab:schema} lists every logical field once and marks where it appears in the chain.
The mutability split of the main text's nested dictionary (\S\ref{sec:env-base}) maps onto these fields directly: \textbf{Board\_static} covers the pad and outline geometry and the design-rule fields, while \textbf{Routing\_geometry} covers the track, via, connectivity, router-state, and DRC fields that change during routing.
The chain is \emph{largely lossless in the sense that matters for routing}. Every field the agent does not see is one the engine still retains and enforces on the agent's behalf.

\paragraph{Lossless where it matters.}
Every row of \cref{tab:schema} that drops out below the engine column does so for one of three reasons, none of which compromises the agent's ability to plan a legal route.

\textit{(1) The engine still enforces it.} Via drill and pad rotation disappear from the agent's view but are used by the C++ DRC engine when it scores the agent's track attempts; the agent decides ``place a via here'', not ``place a via with this drill''.
Per-netclass design rules are collapsed for the agent into a single strictest envelope, but on \texttt{net\_select} the environment pushes the resolved class's track width, via diameter and drill into the PNS router, so the engine routes at the correct per-class dimensions even though the agent only sees the envelope.

\textit{(2) The field is derivable from what the agent does see.} The \texttt{state\_code} duplicates \texttt{is\_routing} plus the phase flag. \texttt{is\_dragging} is non-zero only in GUI drag sessions the agent never triggers. Track and via widths in the LLM prompt are recoverable from \texttt{board\_constraints} for routing-quality reasoning.

\textit{(3) The field is not relevant to routing decisions.} Pad names, footprint references, layer-stack-up dielectrics, 3D models, UUIDs and timestamps are dropped throughout; the schema synthesizes its own short, episode-local IDs (\texttt{P0}, \texttt{T0}, \dots) for addressing.

The genuine losses are limited to two: copper pours (\texttt{(zone ...)}), which the engine itself does not represent in the PNS routable space, and per-class rule differentiation as visible to the agent.
Both are recoverable upgrades rather than architectural barriers. The file's information already reaches the engine, and surfacing it would extend the projection rather than replace it.

\paragraph{Differences between the RL and LLM views are representational, not informational.}
The two agent-facing layers see the same underlying state but consume it differently.
The RL stream emits a discrete \texttt{(action\_type, pointer\_idx, routing\_mode)} triple over a candidate pool synthesized from the current net's pads, vias, track endpoints and a directional grid; coordinates are therefore selected from a finite snap-set.
The LLM emits free-form text constrained by a guided-decoding regex, allowing it to invent detour waypoints anywhere within the bounding box.
The RL stream additionally encodes the top-32 DRC violations and an augmentation context for coordinate symmetries; the LLM compensates with multi-turn history, a no-effect retrofit marker on \texttt{start\_route}+empty-effect pairs, and a rejection-streak slot that surfaces consecutive identical mask rejections.
In both cases the engine is the final filter. Mask rejection, pointer-out-of-pool (RL only) and parse failure (LLM only) all collapse to an \texttt{idle} fallback, so the PNS router never sees illegal input regardless of which consumer is driving.

\section{Design Rule Check Catalog}\label{sec:drc-catalog}

\paragraph{Source of truth.}
\kicad{} enumerates every Design Rule Check it can emit in \texttt{drc\_item.h}: 62 distinct \texttt{DRCE\_*} codes spanning copper-clearance, geometric (track width, annular ring, hole-to-hole), connectivity (unconnected, shorts, dangling), schematic-parity, courtyard, silkscreen, footprint-library and tuning (length / skew / diff-pair) checks.
The default severity for each code is set in \texttt{board\_design\_settings.cpp}: every code starts at \textsc{Error} and a small set of overrides demotes 22 codes to \textsc{Warning} (mostly silkscreen, schematic and library checks plus the dangling/colocated-hole pair) and 5 codes to \textsc{Ignore} (courtyard membership and footprint-filter/type checks), leaving 35 stock error-level checks.

\paragraph{Severity surface seen by our environment.}
Our pipeline never edits the per-code severity table and collapses the resulting violation list into a two-way counting surface:

\begin{itemize}[leftmargin=*]
\item \textbf{Error.} Any violation whose \kicad{} severity is \textsc{Error}, together with three stock-warning codes whose presence invalidates a routed board and which we therefore count at error level: \texttt{DANGLING\_VIA}, \texttt{DANGLING\_TRACK}, and \texttt{NET\_CONFLICT}.
\item \textbf{Other.} Every remaining warning or ignored code: silkscreen and text overlaps, courtyard membership, library/schematic parity, colocated holes, copper slivers, isolated copper, padstack questions, and so on.
These never enter the penalty or the headline DRV count.
\end{itemize}

Both the reward penalty and the evaluation DRV therefore count the same 38 checks, the 35 stock error-level checks plus the three warning-level codes above.
The full enumeration with our classification is in \crefrange{tab:drc-catalog-routing}{tab:drc-catalog-other}; the count-to-penalty mapping is in \cref{tab:drc-weights}.

The catalog splits naturally into routing-relevant codes (the agent's actions can produce or fix them) and non-routing codes (footprint / silk / text / library / schematic / courtyard / zone-fill issues that the agent has no actuators for).
\cref{tab:drc-catalog-routing} lists the routing-relevant subset, and \cref{tab:drc-catalog-other} the rest.
In both tables \emph{Default} is the \kicad{} stock severity and \emph{Env} is the bucket that drives the reward penalty and the DRV count (\textsc{E}~= Error; \textsc{W}~= Warning; \textsc{I}~= Ignore); Env coincides with Default for every code except \texttt{DANGLING\_VIA}, \texttt{DANGLING\_TRACK}, and \texttt{NET\_CONFLICT}, which carry Default \textsc{W} but are counted at Env \textsc{E}.

\begin{table}[h]
\centering\small
\caption{Routing-relevant DRC codes. The agent's actions can directly
produce or clean these.}
\label{tab:drc-catalog-routing}
\begin{adjustbox}{max width=\linewidth}
\begin{tabular}{@{}rlllcc@{}}
\toprule
\textbf{ID} & \textbf{\texttt{DRCE\_*} name} & \textbf{Settings key} & \textbf{Message} & \textbf{Default} & \textbf{Env}\\
\midrule
1  & \texttt{UNCONNECTED\_ITEMS}             & \texttt{unconnected\_items}             & Missing connection between items     & E & E \\
2  & \texttt{SHORTING\_ITEMS}                & \texttt{shorting\_items}                & Items shorting two nets              & E & E \\
3  & \texttt{ALLOWED\_ITEMS}                 & \texttt{items\_not\_allowed}            & Items not allowed (custom rule)      & E & E \\
5  & \texttt{CLEARANCE}                      & \texttt{clearance}                      & Clearance violation                  & E & E \\
6  & \texttt{CREEPAGE}                       & \texttt{creepage}                       & Creepage violation                   & E & E \\
7  & \texttt{TRACKS\_CROSSING}               & \texttt{tracks\_crossing}               & Tracks crossing                      & E & E \\
8  & \texttt{EDGE\_CLEARANCE}                & \texttt{copper\_edge\_clearance}        & Board edge clearance                 & E & E \\
12 & \texttt{DANGLING\_VIA}                  & \texttt{via\_dangling}                  & Via not / partly connected           & W & E \\
13 & \texttt{DANGLING\_TRACK}                & \texttt{track\_dangling}                & Track has unconnected end            & W & E \\
16 & \texttt{HOLE\_CLEARANCE}                & \texttt{hole\_clearance}                & Hole clearance violation             & E & E \\
17 & \texttt{TRACK\_WIDTH}                   & \texttt{track\_width}                   & Track width out of range             & E & E \\
18 & \texttt{TRACK\_ANGLE}                   & \texttt{track\_angle}                   & Track angle out of range             & E & E \\
19 & \texttt{TRACK\_SEGMENT\_LENGTH}         & \texttt{track\_segment\_length}         & Track segment length out of range    & E & E \\
20 & \texttt{ANNULAR\_WIDTH}                 & \texttt{annular\_width}                 & Annular ring too small               & E & E \\
22 & \texttt{DRILL\_OUT\_OF\_RANGE}          & \texttt{drill\_out\_of\_range}          & Hole size out of range               & E & E \\
23 & \texttt{VIA\_DIAMETER}                  & \texttt{via\_diameter}                  & Via diameter out of range            & E & E \\
26 & \texttt{MICROVIA\_DRILL\_OUT\_OF\_RANGE}& \texttt{microvia\_drill\_out\_of\_range}& Microvia hole out of range           & E & E \\
32 & \texttt{DISABLED\_LAYER\_ITEM}          & \texttt{item\_on\_disabled\_layer}      & Item on disabled copper layer        & E & E \\
37 & \texttt{NET\_CONFLICT}                  & \texttt{net\_conflict}                  & Pad net mismatches schematic         & W & E \\
46 & \texttt{ASSERTION\_FAILURE}             & \texttt{assertion\_failure}             & Custom-rule assertion                & E & E \\
47 & \texttt{GENERIC\_WARNING}               & \texttt{generic\_warning}               & Custom-rule warning                  & E & E \\
48 & \texttt{GENERIC\_ERROR}                 & \texttt{generic\_error}                 & Custom-rule error                    & E & E \\
50 & \texttt{SOLDERMASK\_BRIDGE}             & \texttt{solder\_mask\_bridge}           & Solder-mask bridges different nets   & E & E \\
56 & \texttt{LENGTH\_OUT\_OF\_RANGE}         & \texttt{length\_out\_of\_range}         & Track length out of range            & E & E \\
57 & \texttt{SKEW\_OUT\_OF\_RANGE}           & \texttt{skew\_out\_of\_range}           & Skew between tracks out of range     & E & E \\
58 & \texttt{VIA\_COUNT\_OUT\_OF\_RANGE}     & \texttt{too\_many\_vias}                & Too many / few vias on connection    & E & E \\
59 & \texttt{DIFF\_PAIR\_GAP\_OUT\_OF\_RANGE}& \texttt{diff\_pair\_gap\_out\_of\_range}& Diff-pair gap out of range           & E & E \\
60 & \texttt{DIFF\_PAIR\_UNCOUPLED\_LENGTH\_TOO\_LONG} & \texttt{diff\_pair\_uncoupled\_length\_too\_long}& Diff-pair uncoupled too long & E & E \\
\bottomrule
\end{tabular}
\end{adjustbox}

\end{table}

\begin{table}[h]
\centering\small
\caption{Non-routing DRC codes: footprint, silkscreen, text, library,
schematic-parity, courtyard and zone-fill checks.
Listed for completeness; none of these enter our reward penalty or DRV count.}
\label{tab:drc-catalog-other}
\begin{adjustbox}{max width=\linewidth}
\begin{tabular}{@{}rlllcc@{}}
\toprule
\textbf{ID} & \textbf{\texttt{DRCE\_*} name} & \textbf{Settings key} & \textbf{Message} & \textbf{Default} & \textbf{Env}\\
\midrule
4  & \texttt{TEXT\_ON\_EDGECUTS}             & \texttt{text\_on\_edge\_cuts}           & Text on Edge.Cuts layer              & E & E \\
9  & \texttt{ZONES\_INTERSECT}               & \texttt{zones\_intersect}               & Copper zones intersect               & E & E \\
10 & \texttt{ISOLATED\_COPPER}               & \texttt{isolated\_copper}               & Isolated copper fill                 & W & W \\
11 & \texttt{STARVED\_THERMAL}               & \texttt{starved\_thermal}               & Thermal relief incomplete            & E & E \\
14 & \texttt{DRILLED\_HOLES\_TOO\_CLOSE}     & \texttt{hole\_to\_hole}                 & Drilled hole too close to other      & W & W \\
15 & \texttt{DRILLED\_HOLES\_COLOCATED}      & \texttt{holes\_co\_located}             & Drilled holes co-located             & W & W \\
21 & \texttt{CONNECTION\_WIDTH}              & \texttt{connection\_width}              & Copper connection too narrow         & W & W \\
24 & \texttt{PADSTACK}                       & \texttt{padstack}                       & Padstack questionable                & W & W \\
25 & \texttt{PADSTACK\_INVALID}              & \texttt{padstack\_invalid}              & Padstack not valid                   & E & E \\
27 & \texttt{OVERLAPPING\_FOOTPRINTS}        & \texttt{courtyards\_overlap}            & Courtyards overlap                   & E & E \\
28 & \texttt{MISSING\_COURTYARD}             & \texttt{missing\_courtyard}             & Footprint has no courtyard           & I & I \\
29 & \texttt{MALFORMED\_COURTYARD}           & \texttt{malformed\_courtyard}           & Malformed courtyard                  & E & E \\
30 & \texttt{PTH\_IN\_COURTYARD}             & \texttt{pth\_inside\_courtyard}         & PTH inside courtyard                 & I & I \\
31 & \texttt{NPTH\_IN\_COURTYARD}            & \texttt{npth\_inside\_courtyard}        & NPTH inside courtyard                & I & I \\
33 & \texttt{INVALID\_OUTLINE}               & \texttt{invalid\_outline}               & Malformed board outline              & E & E \\
34 & \texttt{MISSING\_FOOTPRINT}             & \texttt{missing\_footprint}             & Missing footprint                    & W & W \\
35 & \texttt{DUPLICATE\_FOOTPRINT}           & \texttt{duplicate\_footprints}          & Duplicate footprints                 & W & W \\
36 & \texttt{EXTRA\_FOOTPRINT}               & \texttt{extra\_footprint}               & Extra footprint                      & W & W \\
38 & \texttt{SCHEMATIC\_PARITY}              & \texttt{footprint\_symbol\_mismatch}    & Footprint attrs don't match symbol   & W & W \\
39 & \texttt{FOOTPRINT\_FILTERS}             & \texttt{footprint\_filters\_mismatch}   & Footprint outside symbol filters     & I & I \\
40 & \texttt{FOOTPRINT\_TYPE\_MISMATCH}      & \texttt{footprint\_type\_mismatch}      & Footprint type vs. pads mismatch     & I & I \\
41 & \texttt{LIB\_FOOTPRINT\_ISSUES}         & \texttt{lib\_footprint\_issues}         & Footprint not in libraries           & W & W \\
42 & \texttt{LIB\_FOOTPRINT\_MISMATCH}       & \texttt{lib\_footprint\_mismatch}       & Doesn't match library copy           & W & W \\
43 & \texttt{PAD\_TH\_WITH\_NO\_HOLE}        & \texttt{through\_hole\_pad\_without\_hole}& Through-hole pad without hole      & E & E \\
44 & \texttt{FOOTPRINT}                      & \texttt{footprint}                      & Footprint not valid                  & E & E \\
45 & \texttt{UNRESOLVED\_VARIABLE}           & \texttt{unresolved\_variable}           & Unresolved text variable             & E & E \\
49 & \texttt{COPPER\_SLIVER}                 & \texttt{copper\_sliver}                 & Copper sliver                        & W & W \\
51 & \texttt{SILK\_CLEARANCE}                & \texttt{silk\_over\_copper}             & Silkscreen clipped by mask           & W & W \\
52 & \texttt{SILK\_EDGE\_CLEARANCE}          & \texttt{silk\_edge\_clearance}          & Silkscreen clipped by board edge     & W & W \\
53 & \texttt{TEXT\_HEIGHT}                   & \texttt{text\_height}                   & Text height out of range             & W & W \\
54 & \texttt{TEXT\_THICKNESS}                & \texttt{text\_thickness}                & Text thickness out of range          & W & W \\
55 & \texttt{OVERLAPPING\_SILK}              & \texttt{silk\_overlap}                  & Silkscreen overlap                   & W & W \\
61 & \texttt{MIRRORED\_TEXT\_ON\_FRONT\_LAYER} & \texttt{mirrored\_text\_on\_front\_layer} & Mirrored text on front layer    & W & W \\
62 & \texttt{NONMIRRORED\_TEXT\_ON\_BACK\_LAYER} & \texttt{nonmirrored\_text\_on\_back\_layer} & Non-mirrored text on back layer & W & W \\
\bottomrule
\end{tabular}
\end{adjustbox}
\end{table}

\paragraph{Count-to-penalty mapping.}
The reward potential and the evaluation DRV count both count error-level violations, as follows.

\begin{table}[h]
\small
\caption{How each violation bucket enters the DRC penalty $f_{\mathrm{drv}}$ used in
the potential $\Phi$ (Equation~\eqref{eq:phi-default}) and the evaluation count.
The training penalty $f_{\mathrm{drv}}$ takes the log-per-net shape of Equation~\eqref{eq:fd-default}; here $x$ is the number of distinct nets carrying at least one in-bucket violation (breadth) and $x_i$ is the number of in-bucket violations on net~$i$ (depth).
The $3{:}1$ ratio is the breadth-vs-depth weighting, not error-vs-warning.}
\label{tab:drc-weights}
{\centering
\begin{adjustbox}{max width=\linewidth}
\begin{tabular}{@{}llcc@{}}
\toprule
\textbf{Bucket} & \textbf{Source} & \textbf{Reward penalty $f_{\mathrm{drv}}$} & \textbf{Eval DRV}\\
\midrule
Error                 & \kicad{} \textsc{Error} severity + dangling/net-conflict trio & counted (see below) & $+1$ \\
Warning               & remaining \kicad{} \textsc{Warning} codes                  & $0$ & $0$ \\
Ignore                & \kicad{} \textsc{Ignore} severity                          & $0$ & $0$ \\
\bottomrule
\end{tabular}
\end{adjustbox}
\par}
\vspace{2pt}
{\footnotesize
\noindent Both sums are taken over the Error bucket (\texttt{drc\_errors}).
The CP metric reported in \oursbench{} is the joint indicator \(\mathbf{1}\{\text{routability}=1\}\cdot \mathbf{1}\{\texttt{drc\_errors}=0\}\).
The penalty $f_{\mathrm{drv}}$ enters $\Phi$ with a minus sign (Equation~\eqref{eq:phi-default}), so larger $f_{\mathrm{drv}}$ means smaller reward.\par}
\end{table}

\paragraph{Why this subset.}
Of the 62 cataloged codes, the 28 in \cref{tab:drc-catalog-routing} are routing-relevant in the sense that the agent's actions can directly produce or fix them; the rest (\cref{tab:drc-catalog-other}) concern schematic parity, library state, courtyards, silkscreen and text, which the agent has no actuators for.
Of the 28 routing-relevant codes, all but the dangling/net-conflict trio carry stock \textsc{Error} severity.
The trio are stock warnings, but a dangling track or via and a net conflict each invalidate a routed board, so we count them at error level alongside the stock errors; all 28 routing-relevant codes therefore enter the penalty and the DRV count.
The non-routing warnings are deliberately left out of the penalty so that training signal is concentrated on outcomes the policy can affect, and left out of the headline DRV count so that comparable boards routed by different agents are not penalized for upstream artifacts they share.

\paragraph{Mapping LLM failure modes to DRC buckets.} \cref{tab:llm-failure-to-drc} gives representative examples of how LLM routing failures surface in the DRC reports, or as non-DRC outcomes such as mask reject, parse fail, or no-effect steps.
Failures that never reach the engine carry no DRC code and are bucketed as \emph{none}.

\begin{table}[h]
\centering\small
\caption{Representative mapping from LLM failure modes to DRC buckets.
Rows whose failures never reach the engine carry no DRC code; their evidence appears in the Secondary / non-DRC column and their Bucket is \emph{none}.}
\label{tab:llm-failure-to-drc}
\begin{adjustbox}{max width=\linewidth}
\begin{tabular}{@{}lllc@{}}
\toprule
\textbf{Failure mode} & \textbf{Primary DRC code(s)} & \textbf{Secondary / non-DRC} & \textbf{Bucket}\\
\midrule
\emph{via on pad}                        & \texttt{ANNULAR\_WIDTH}, \texttt{HOLE\_CLEARANCE} & --- & E \\
\emph{wrong-layer endpoint}              & \texttt{DANGLING\_TRACK}                          & no-effect step & E \\
\emph{cross-net contact}                 & \texttt{SHORTING\_ITEMS}, \texttt{CLEARANCE}      & --- & E \\
\emph{stranded segment}                  & \texttt{DANGLING\_TRACK}, \texttt{DANGLING\_VIA}  & --- & E \\
\emph{detour over board edge}            & \texttt{EDGE\_CLEARANCE}                          & --- & E \\
\emph{invalid coordinate / phantom waypoint} & ---                                           & parse fail / mask reject & none \\
\emph{repeated identical attempt}        & ---                                                & rejection streak & none \\
\bottomrule
\end{tabular}
\end{adjustbox}
\end{table}

\section{Reward potential and DRC penalty}\label{sec:reward-detail}

This section instantiates the potential $\Phi$ defined in \S\ref{sec:env-reward}, which we restate here:
\begin{equation}
\Phi(s) \;=\; -\bigl(f_{\mathrm{drv}}(s) + \lambda_w\,\ell(s) + \lambda_v\,n_{\mathrm{via}}(s)\bigr),
\label{eq:phi-default}
\end{equation}
and specifies the concrete choices used by our learned baselines: the DRC counting convention $n_{\mathrm{drv}}$, the concave shape of $f_{\mathrm{drv}}$, and the quality-term weights $(\lambda_w,\lambda_v)$.

\paragraph{DRC counting convention.}
Let $\mathcal{V}(s)$ be the multiset of DRC violations we count at error level: the 35 stock error-level checks together with \texttt{DANGLING\_VIA}, \texttt{DANGLING\_TRACK}, and \texttt{NET\_CONFLICT}, the 38 checks cataloged in Appendix~\ref{sec:drc-catalog}.
We take $n_{\mathrm{drv}}(s)=|\mathcal{V}(s)|$, and additionally decompose it per offending net~$i$:
\[\adjustbox{max width=\linewidth}{$\displaystyle
\begin{aligned}
x_i(s) &= \bigl|\bigl\{v\in\mathcal{V}(s)\,:\,\mathrm{net}(v)=i\bigr\}\bigr|
       && \text{(\emph{depth}: per-net violation count)},\\
x(s)   &= \bigl|\bigl\{i\,:\,x_i(s)>0\bigr\}\bigr|
       && \text{(\emph{breadth}: number of dirtied nets)},
\end{aligned}
$}\]
so that $n_{\mathrm{drv}}(s) = \sum_i x_i(s)$.

\paragraph{Shape of the DRC penalty $f_{\mathrm{drv}}$.}
Real industrial quality requires producing boards with $\mathrm{DRV}=0$. A board with one residual violation is no more shippable than a board with ten, so reward improvements only matter insofar as they push the policy toward the clean-board boundary.
We therefore adopt a concave, logarithmic shape for $f_{\mathrm{drv}}$, which assigns a large marginal penalty to the \emph{first} violation on a clean net and rapidly diminishing marginal penalty to each additional violation.
This concentrates the learning signal on closing out the last violations, where industrial quality is decided, rather than on incremental reductions on already dirty boards.
Concretely, we instantiate the concave penalty in Equation~\eqref{eq:phi-default} as a log shape that is separately concave in breadth and depth and unbounded above:
\begin{equation}
\adjustbox{max width=\linewidth}{$\displaystyle
f_{\mathrm{drv}}(s)
\;=\;
s_{\mathrm{agg}}\;\ln\!\Bigl(1+\tfrac{x(s)}{o}\Bigr)
\;+\;
s_{\mathrm{pn}}\,\sum_{i:\,x_i(s)>0}
\ln\!\Bigl(1+\tfrac{x_i(s)}{o}\Bigr),
$}
\label{eq:fd-default}
\end{equation}
with hyperparameters $(s_{\mathrm{agg}},\,s_{\mathrm{pn}},\,o)=(3,\,1,\,2)$.
The first (aggregate) term penalizes \emph{breadth} (how many distinct nets the policy has dirtied), while the per-net sum penalizes \emph{depth} (how many violations pile up on each individual net).
The shared offset $o$ acts as the log-curve knee, so a single new violation on a clean net contributes a bounded per-net term of $\ln(3/2)$ even when many other nets are already dirty, preventing the gradient from collapsing on heavily violated boards.
The $3{:}1$ ratio biases the policy toward routing fewer dirty nets rather than minimizing violations on a single dirty net.

\paragraph{Quality-term weights during training.}
Unless otherwise noted, the reported policies are trained with the default quality weights $(\lambda_w,\lambda_v)=(0.002,0.1)$, so the full potential of Equation~\eqref{eq:phi-default}, including the wirelength and via terms, is optimized during training.
The controllability study (\S\ref{sec:exp-rl}) sweeps these weights over a $3{\times}3$ grid, $\lambda_w\in\{0,0.001,0.002\}$ and $\lambda_v\in\{0,0.05,0.1\}$. Its feasibility-only corner $(\lambda_w,\lambda_v)=(0,0)$ trains against $\Phi(s)=-f_{\mathrm{drv}}(s)$ alone.

\paragraph{Severity-mode coupling.}
The reward penalty $f_{\mathrm{drv}}$, the DRC tokens emitted into the agent's state, and the headline evaluation DRV all count the same error-level violations, so the agent is never trained against violations it cannot observe, nor evaluated on a different set than it was rewarded for. The state tokens are specified in Appendix~\ref{sec:state-schema} and the evaluation count in Appendix~\ref{sec:metrics}.

\section{Wrapper Details}\label{sec:impl-detail}
\subsection{LLM Wrapper}\label{sec:llm_wrapper}

The wrapper builds each prompt by instantiating a fixed template with runtime values.
We partition the prompt into a static \emph{system} message and a dynamic per-turn \emph{user} message. The full prompt templates are reproduced in Appendix~\ref{sec:prompt-agent}.
The system message contains episode-invariant content: the agent role and priority, routing guidelines, the response-format contract, the board-state schema description, and the \texttt{board\_static} block (footprints, pads, nets, and design rules).
By placing these invariant fields in the system message, the wrapper keeps the prompt prefix stable across steps, making the prompt layout compatible with prefix caching.

The user message is regenerated at every environment step from the current routing state and bookkeeping variables.
It contains the 1-indexed step counter, the freshly serialized observation obtained by concatenating the \texttt{routing\_geometry} and \texttt{router\_head} blocks, the number of accepted actions so far, a rolling window of the recent action history, an optional rejection-streak line, and the list of action verbs allowed by the current router-phase mask.
Coordinates in the dynamic observation are rounded to three decimal places in millimeters.
All numerical values are normalized to a uniform decimal precision, with special sentinels (e.g., the through-hole layer) rendered as distinct tokens to prevent confusion with integers.
The prompt's guidelines cover target identification, coordinate usage, obstacle avoidance, layer-change detours, and explicit net release.
The rejection slot is empty in the nominal case; after a parse failure or mask veto, consecutive identical rejections are collapsed into a single ``\texttt{(rejected $\times$N)}'' line and cleared once the next valid action is accepted.

Every model response is funneled through a strict parser before it can touch the PNS router.
The parser requires a single \texttt{<think></think>}\,\texttt{<action></action>} pair, a known verb, the right arity, and ASCII-only content; failures fall back to an explicit \texttt{idle} fallback at index $6$.
\texttt{idle} is an internal no-op index, not one of the six action types of \cref{tab:action-description}, and leaves the board unchanged.
After parsing, the action index is intersected with the current phase's mask (\texttt{net\_select} $\to$ \texttt{start\_route} $\to$ routing); a mask miss is also rerouted to \texttt{idle} and the offending body is pushed onto the rejection streak so the next turn's user prompt can surface it.
Successfully dispatched actions that nonetheless leave the unrouted-pin count unchanged are kept but tagged \texttt{[no effect]} in history, and any preceding \texttt{start\_route} that they exposed as wasted is retroactively re-tagged.
The reward consequences of each class are summarized in \cref{tab:parser-fallback}.

\begin{table}[h]
\centering\small
\caption{Parser and action-mask fallback policy.
The \emph{Action taken} column
is what the underlying \kicad{} PNS router actually executes.}
\label{tab:parser-fallback}
\begin{adjustbox}{max width=\linewidth}
\begin{tabular}{@{}lll@{}}
\toprule
\textbf{Failure mode} & \textbf{Trigger} & \textbf{Action taken}\\
\midrule
Missing \texttt{<action>} tag      & no/empty action block                       & \texttt{idle} (idx 6) \\
Malformed body                     & unknown verb, wrong arity, conv.\ error      & \texttt{idle} (idx 6)  \\
Bad \texttt{<think>} block         & missing/duplicated reasoning channel         & \texttt{idle} (idx 6)  \\
Non-ASCII content                  & e.g.\ CJK characters in output               & \texttt{idle} (idx 6)  \\
Mask reject                        & valid parse, disallowed in current phase    & \texttt{idle} (idx 6)  \\
No-effect step                     & dispatched but \texttt{unrouted} unchanged  & as emitted, tagged     \\
\bottomrule
\end{tabular}
\end{adjustbox}
\end{table}

\subsection{RL Wrapper}
\label{app:rl-wrapper}

The RL wrapper turns the state dictionary of \S\ref{sec:env-base} into a batched token sequence and exposes the action space of \S\ref{sec:env-action} as a typed autoregressive head over the engine-provided candidate set.

\paragraph{Tokenization.} Each geometric or structural object becomes a single token.
We support
thirteen entity types: \texttt{BOARD}, \texttt{EDGE}, \texttt{NET}, \texttt{PAD}, \texttt{TRACK}, \texttt{VIA}, \texttt{RAT}, \texttt{HEAD}, four \texttt{CAND\_*} variants for the action-time candidate pool, and \texttt{DRC\_VIOLATION}.
For a token $i$ of type $\tau(i)$ with raw feature vector $\mathbf{f}_i$, the token embedding is the sum of an entity-type embedding and a per-type linear projection of its features:
\[
\mathbf{x}_i \;=\; \mathbf{e}_{\tau(i)} \;+\; \mathbf{W}_{\tau(i)}\,\mathbf{f}_i.
\]
Coordinates and dimensions are first normalized by the board center and a reference scale quantized onto a predefined discrete set, so identical physical quantities yield identical tokens regardless of board size.
The normalized continuous fields (point coordinates, segment widths, via diameters) then pass through a sin/cos Fourier feature map with $n_{\mathrm{freq}}=32$ and geometric base $1.20$; copper layers are encoded as the $(\text{dist\_top},\text{dist\_bot})$ pair.
\texttt{TRACK} tokens use symmetric endpoint pooling so that swapping the two endpoints yields the same embedding.
Every geometric token also receives the head-relative distance to the current routing head as an inductive feature.
A learned slot embedding distinguishes net-internal sub-objects, and the full sequence is finally normalized with LayerNorm.
With $d_{\mathrm{model}}=128$ and $n_{\mathrm{layers}}=4$, the resulting flat sequence preserves the native object hierarchy of the state dictionary while remaining permutation-equivariant within each entity group.

\paragraph{Autoregressive action head.} At step $t$, the action $a_t$ of
\S\ref{sec:env-action} is decoded as a triple $(\alpha,\,k,\,m)$ comprising an action type $\alpha$, a pointer index $k$ into the candidate set, and a routing mode $m$.
We follow the slot table in \cref{tab:rl-slot-usage}: \texttt{net\_select} and \texttt{start\_route} use only a pointer; \texttt{make\_line} and \texttt{make\_via} use both pointer and mode; \texttt{finish} uses only mode; \texttt{net\_end} uses neither.
Decoding proceeds in two Transformer passes.
In pass one the state sequence is extended with a single \texttt{SOD} token; the action-type logits are produced by a tied \texttt{action\_type\_head} embedding, sampled, and re-injected as a typed token to obtain a hidden state $\mathbf{h}_\alpha$.
The pointer distribution scores $\mathbf{h}_\alpha$ against the per-token states of the eligible candidate slice (net tokens for \texttt{net\_select}, candidate tokens otherwise) by scaled dot-product.
In pass two the chosen candidate's state token is re-emitted as \texttt{point\_tok} to obtain $\mathbf{h}_k$; the routing-mode distribution then dot-products $\mathbf{h}_k$ against the routing-mode embedding table reused from the tokenizer vocabulary.

\paragraph{Candidate pool.} The candidate set is built per-step from the active net.
It contains the
net's pads (thru-hole pads expanded to one entry per copper layer), the endpoints of its already-drawn tracks, its via centers, and an eight-way directional grid of $0.5$\,mm offsets around the current head (a four-way grid snapped to the underlying lattice when running a D1 grid instance).
Candidates are deduplicated by $(x,y,\text{layer})$ and truncated to $64$ slots.
Pointer selection therefore operates on a finite pool emitted by the engine; out-of-pool indices are masked to $-\infty$ before sampling.

\paragraph{Masking and stability.} The same FSM-based action mask of \S\ref{sec:env-base}
(\texttt{net\_select}\, $\rightarrow$\,\texttt{start\_route}\, $\rightarrow$\,\texttt{routing}) is applied to action-type logits; per-net pointer masks block out-of-scope candidates, and an additional pointer mask excludes the freshly started route's own coordinate across all layers, preventing self-loops and the resulting performance collapse; routing-mode masks honor the configured rule set.
Action-type and pointer logits are passed through $10\cdot\tanh(\cdot)$ to prevent early-training logit blow-up.
Unused slots in the emitted action triple are set to $-\inf$, so every emitted action is syntactically complete and lies inside the engine's valid-action set by construction.

\begin{table}[h]
\centering
\small
\begin{tabular}{lcc}
\toprule
Action type & Needs pointer & Needs mode \\
\midrule
\texttt{net\_select}  & \checkmark & \\
\texttt{start\_route} & \checkmark & \\
\texttt{net\_end}     & & \\
\texttt{make\_line}   & \checkmark & \checkmark \\
\texttt{make\_via}    & \checkmark & \checkmark \\
\texttt{finish}       & & \checkmark \\
\texttt{idle}         & & \\
\bottomrule
\end{tabular}
\caption{Slot usage per action type.
Unused slots are emitted as $-1$ in
the action triple and contribute no log-prob term.}
\label{tab:rl-slot-usage}
\end{table}
\section{Evaluation Metrics}\label{sec:metrics}

We evaluate PCB routing quality from multiple complementary angles, combining board-level aggregates with rollout-level statistics.
For each board $b \in \mathcal{B}$, we generate $k$ independent rollouts under the same budget; we use $k=5$ throughout.
We denote the $i$-th rollout for board $b$ as $s_{b,i}$ for $i \in \{1,\dots,k\}$, and reserve $s_{b,0}$ for the bare-board state before any routing.
We use $\mathbf{1}[\cdot]$ for the indicator function.

\paragraph{Potential Gain (Pot.$\uparrow$).}
Our central routing-quality measure is the \emph{potential gain} from the bare-board state.
Let $\Phi(s)$ be the board-level potential function introduced in \S\ref{sec:env-reward}, which jointly penalizes design rule violations, wirelength, and via count (and thereby captures net-level routing quality; see Appendix~\ref{sec:reward-detail}).
For a rollout $s_{b,i}$, the potential gain is
\[
\Delta \Phi(s_{b,i}) = \Phi(s_{b,i}) - \Phi(s_{b,0}),
\]
which measures how much routing progress has been made relative to the bare-board state $s_{b,0}$: a larger $\Delta\Phi$ reflects a cleaner, shorter, and more complete routing solution.
As a reported metric, Pot. is the potential gain of the per-board selected rollout (defined next), averaged over boards:
\[
\text{Pot.}
=
\frac{1}{|\mathcal{B}^{\star}|}
\sum_{(b,s_b^{\star}) \in \mathcal{B}^{\star}}
\Delta \Phi(s_b^{\star}).
\]

\paragraph{Rollout Selection.}
Potential gain also defines how a single representative rollout is chosen per board, which all subsequent rollout-level metrics are computed on.
For each board $b$ we select the rollout with the largest potential gain,
\[
I_b = \arg\max_{i \in \{1,\dots,k\}} \Delta \Phi(s_{b,i}), \qquad s_b^{\star} = s_{b,I_b},
\]
and collect the selected pairs into
\[
\mathcal{B}^{\star} = \bigl\{\, (b,s_b^{\star}) : b \in \mathcal{B} \,\bigr\}.
\]
Every per-board diagnostic metric defined below is reported on this potential-selected rollout $s_b^{\star}$, so that all metrics describe the same representative rollout.
The exceptions are Time, which is averaged over all $k$ rollouts, and Parse-fail, which is computed from all $k$ (see their paragraphs below).

\paragraph{Routability (Rout.$\uparrow$).}
We measure connectivity using \kicad{}'s \emph{ratsnest} connections: the set of pad-to-pad connections that the netlist mandates but that have not yet been physically routed, rendered in \kicad{} as the residual ``rats'' lines.
For a rollout $s_{b,i}$, let $R(s_{b,i})$ denote the number of remaining ratsnest edges.
Since $s_{b,0}$ denotes the bare-board state before any routing, the ratsnest-based routability of $s_{b,i}$ is
\[
\text{rout}(s_{b,i}) = \frac{R(s_{b,0}) - R(s_{b,i})}{R(s_{b,0})} \in [0,1],
\]
i.e., the fraction of required ratsnest connections resolved relative to the bare-board state.
A rollout is \emph{fully routed} iff $R(s_{b,i}) = 0$, equivalently $\text{rout}(s_{b,i}) = 1$.
We report Rout. on the per-board rollouts selected by potential gain:
\[\adjustbox{max width=\linewidth}{$\displaystyle
\text{Rout.}
=
\frac{1}{|\mathcal{B}^{\star}|}
\sum_{(b,s_b^{\star}) \in \mathcal{B}^{\star}}
\text{rout}(s_{b}^{\star})
=
\frac{1}{|\mathcal{B}^{\star}|}
\sum_{(b,s_b^{\star}) \in \mathcal{B}^{\star}}
\frac{R(s_{b,0}) - R(s_{b}^{\star})}{R(s_{b,0})}.
$}\]

\paragraph{Clean Pass (CP$\uparrow$).}
Connectivity alone is insufficient as a measure of manufacturability.
We therefore define a stricter criterion: a rollout $s_{b,i}$ is \emph{clean} iff (i) the residual ratsnest is empty, $R(s_{b,i}) = 0$, and (ii) the number of \kicad{} error-level design rule violations is zero, $\text{drv}(s_{b,i}) = 0$.
Consistent with the other metrics, we evaluate cleanliness on the per-board selected rollout $s_b^{\star}$ (the rollout of largest potential gain):
\[
\text{CP}
\;=\;
\frac{1}{|\mathcal{B}|}
\sum_{b \in \mathcal{B}}
\mathbf{1}\!\left[\,
\text{rout}(s_b^{\star}) = 1 \,\wedge\, \text{drv}(s_b^{\star}) = 0
\,\right].
\]
This is the closest proxy in our evaluation to ``ready-to-fabricate'' output.
We draw $k=5$ rollouts per board and report CP on the selected rollout $s_b^{\star}$.

\paragraph{DRV count (Design Rule Violation, DRV$\downarrow$).}
We quantify the severity of rule violations by the \kicad{} error count $\text{drv}(s_{b,i}) = \texttt{drc\_errors}(s_{b,i})$.
Following the main-text convention in \S\ref{sec:bench-metrics}, we report DRV on the per-board rollouts selected by potential gain:
\[
\text{DRV}
\;=\;
\frac{1}{|\mathcal{B}^{\star}|}
\sum_{(b,s_b^{\star}) \in \mathcal{B}^{\star}}
\text{drv}(s_{b}^{\star}).
\]

\paragraph{Physical Cost Metrics (WL$\downarrow$, Via$\downarrow$).}
We report two physical cost metrics on the per-board rollouts selected by potential gain.
WL is the total routed wirelength in millimeters, summed over all nets, and Via is the number of vias in the routed board.
Both are common secondary objectives in PCB routing, as shorter traces and fewer vias generally reduce routing cost and manufacturing complexity.
They are aggregated over $\mathcal{B}^{\star}$ as
\[\adjustbox{max width=\linewidth}{$\displaystyle
\text{WL}
=
\frac{1}{|\mathcal{B}^{\star}|}
\sum_{(b,s_b^{\star}) \in \mathcal{B}^{\star}}
\text{wl}(s_b^{\star}),
\qquad
\text{Via}
=
\frac{1}{|\mathcal{B}^{\star}|}
\sum_{(b,s_b^{\star}) \in \mathcal{B}^{\star}}
\text{via}(s_b^{\star}).
$}\]

\paragraph{Wallclock time (Time$\downarrow$).}
We report wallclock routing time per rollout (in seconds), measured from the first API call of a rollout until the rollout completes or exhausts its step budget.
Unlike the other metrics, Time is computed over \emph{all} $k$ rollouts on \emph{every} board, regardless of routability: we take the rollout-mean per board and then the mean across boards,
\[
\text{Time}
\;=\;
\frac{1}{|\mathcal{B}|}
\sum_{b \in \mathcal{B}}
\frac{1}{k}
\sum_{i=1}^{k}
\text{time}(s_{b,i}).
\]
Restricting to fully routed rollouts would bias the cost in favor of methods that abort early on hard boards; the formulation above charges every rollout for the time it actually consumed.
For the deterministic baselines (OrthoRoute and KRT), every rollout is identical, so @5 and @1 coincide and one routing pass yields all reported metrics.
Freerouting and the RL agents report each metric as the mean over 4 seeds.
The LLM agents run the five-rollout @5 protocol without seed repetition.

\paragraph{Parse failure rate (Parse-fail$\downarrow$).}
Parse-fail measures whether an LLM agent's output respects the basic \kicad{} board-file syntax.
Let $\mathrm{fail}(s_{b,i}) \in \{0,1\}$ indicate that the board produced by rollout $s_{b,i}$ fails to load as a valid \texttt{.kicad\_pcb} and therefore cannot be evaluated.
Under the default @5 protocol a board counts as a parse failure only when none of its $k$ rollouts loads, leaving selection no candidate, whereas Parse-fail@1 is the fraction of individual rollouts that fail,
\[\adjustbox{max width=\linewidth}{$\displaystyle
\text{Parse-fail}
=
\frac{1}{|\mathcal{B}|}
\sum_{b \in \mathcal{B}}
\prod_{i=1}^{k}
\mathrm{fail}(s_{b,i}),
\qquad
\text{Parse-fail@1}
=
\frac{1}{|\mathcal{B}|\,k}
\sum_{b \in \mathcal{B}}
\sum_{i=1}^{k}
\mathrm{fail}(s_{b,i}).
$}\]
\cref{tab:llm-ablation} reports Parse-fail under the default @5 protocol.
The panels of \cref{fig:llm_ablation} plot Parse-fail@1 in percent.
The metric is most informative for the engine-free mode. It emits the routing directly in \texttt{.kicad\_pcb} syntax, so any violation of the file grammar makes the assembled board unloadable.
In the interactive and plan-and-execute modes the accepted actions execute inside the engine, which serializes the board, so the produced file is well formed by construction.
We still report the metric for all three modes for comparability.
Model responses that fail the per-step action parser during a rollout fall back to \texttt{idle} and leave the board unchanged (\cref{tab:parser-fallback}).
These per-step rejections degrade CP and Rout.\ but are not counted by Parse-fail.
Parse-fail is a diagnostic of model-generated text, so we report it only for the LLM agents.
\section{Experimental Setup and Model Serving}\label{sec:model-serving}
\subsection{Proprietary LLM and Open-Source LLM}
\label{sec:llm-serving}

We benchmark the LLM agent on three OpenAI cloud-API models and one open-weights Qwen model served through Together AI.
\cref{tab:llm-setup} summarizes the deployment configuration for each: serving infrastructure, hardware when exposed by the provider, weight quantization as served, reasoning mode (``hidden'' for closed models with internal chain-of-thought, \texttt{think-off} for the Qwen variant where the explicit reasoning channel is disabled), the per-turn token budget for generation, and the datasets each model is evaluated on (D2, D3-A, and D3-B).

\begin{table}[h]
\centering\small
\caption{Deployment configuration of every LLM in \oursbench{}.
``--'' = not applicable / not exposed by the provider.}
\label{tab:llm-setup}
\begin{adjustbox}{max width=\linewidth}
\begin{tabular}{@{}lllllcl@{}}
\toprule
\textbf{Model} & \textbf{Infrastructure} & \textbf{Hardware} & \textbf{Quant.} & \textbf{Thinking} & \textbf{Max tok.} & \textbf{Datasets}\\
\midrule
\texttt{gpt-5.4}              & API (OpenAI)         & cloud              & --   & hidden        & 256 & D2, D3-A, D3-B \\
\texttt{gpt-5.4-mini}         & API (OpenAI)         & cloud              & --   & hidden        & 256 & D2, D3-A, D3-B \\
\texttt{gpt-5.4-nano}         & API (OpenAI)         & cloud              & --   & hidden        & 256 & D2, D3-A, D3-B \\
\midrule
\texttt{Qwen3.5-397B-A17B}    & API (Together)  & --                & FP4  & think-off     & 512     & D2, D3-A, D3-B \\
\bottomrule
\end{tabular}
\end{adjustbox}
\end{table}

\section{Hyperparameters and compute}\label{sec:hyperparams}
This appendix records the compute infrastructure, the configuration of the main-result RL policies (\cref{tab:hp-rl-main}), and the optimization and architecture settings of the D1 grid-size/action-abstraction scalability experiment (\cref{tab:rq1-compute}).

\subsection{Compute infrastructure}\label{sec:compute-infra}
All RL agents were trained on a single NVIDIA L40 48\,GB GPU per run.
Each training node carries two AMD EPYC 9354 32-core CPUs and 1.5\,TiB of RAM and runs Ubuntu 20.04 or 22.04.
Evaluation rollouts execute headless \kicad{} engine calls and are CPU-bound.
The environment and the PPO and GRPO agents run in Python 3.12.13 with PyTorch 2.8.0 (CUDA 12.8, cuDNN 9.10.2), NumPy 2.2.6, and Gymnasium 1.2.3.
The PPO and GRPO training loops are our own implementation.
The engine is \kicad{} 9.0.8 with the \texttt{kicad-python} 0.6.0 Python$\leftrightarrow$C++ binding.
Freerouting 2.1.0 runs under OpenJDK 21.0.6, and OrthoRoute is pinned to commit \texttt{f45dc68}.
The A2C (Jumanji) and Sable baselines run in a separate JAX 0.5.3 environment with Jumanji 1.1.1, Mava 0.2.0, Flax 0.10.3, and Optax 0.2.8.

\subsection{Main-result RL policy hyperparameters}\label{sec:hp-rl-main}
The PPO, PPO (terminal), GRPO, and PPO (w/o \texttt{finish}) entries of \cref{tab:rq2} all train the same decoder-only Transformer policy from scratch on D2-train, selecting checkpoints on D2-valid; the policy is then evaluated by rollout on D2-test and the zero-shot D3 open-source set.
The variants share the architecture, optimizer, and reward weights (\cref{tab:hp-rl-main}), and differ only in the reward form (per-step dense vs.\ terminal sparse), the number of training iterations, and, for PPO (w/o \texttt{finish}), the removal of \texttt{finish} from the agent's action space.
In practice we train with $\gamma = 0.995$ (\cref{tab:hp-rl-main}), for which the telescoping equivalence between the per-step and terminal rewards (\S\ref{sec:env-reward}) holds approximately.
\begin{table}[t]\centering\small
\caption{Configuration of the main-result RL policy (trained on D2-train; evaluated
on D2-test and zero-shot D3 test boards), shared by the PPO / PPO (terminal) / GRPO / PPO (w/o \texttt{finish}) entries of \cref{tab:rq2}.}
\label{tab:hp-rl-main}
\begin{adjustbox}{max width=\linewidth}
\begin{tabular}{ll}
\toprule
\multicolumn{2}{l}{\emph{Policy network (shared)}}\\ \midrule
Architecture            & decoder-only Transformer \\
$d_{\text{model}}$ / layers / heads / FFN & 128 / 4 / 8 / 512 \\
Coordinate encoding     & Fourier ($n_{\text{freq}}{=}32$), MLP width 128 \\
Activation / residual   & GELU / ReZero \\
Action / value head     & pointer network (logit clip $10$) / MLP critic (PPO only) \\
\midrule
\multicolumn{2}{l}{\emph{Optimization (shared)}}\\ \midrule
Optimizer               & AdamW ($\epsilon{=}10^{-5}$, weight decay $10^{-4}$) \\
Learning rate           & $10^{-4}$ (20-iteration warmup) \\
Discount $\gamma$ / GAE $\lambda$ & $0.995$ / $0.95$ \\
PPO clip $\epsilon$     & $0.2$ \\
Entropy / value coef.\  & $0.01$ / $0.5$ \\
Max gradient norm       & $0.5$ \\
Batch size / update epochs & $256$ / $4$ \\
Parallel envs / rollout horizon & $32$ / $512$ \\
Episode step limit      & $256$ \\
\midrule
\multicolumn{2}{l}{\emph{Reward (shared)}}\\ \midrule
Wirelength weight $\lambda_w$ / via weight $\lambda_v$ & $0.002$ / $0.1$ \\
DRC penalty $f_{\mathrm{drv}}$       & log-per-net, error-level violations \\
\midrule
\multicolumn{2}{l}{\emph{Per-variant}}\\ \midrule
PPO            & dense per-step reward, $300$ iterations \\
PPO (terminal) & sparse terminal reward, $300$ iterations \\
GRPO           & sparse reward, group size $16$, $1800$ iterations \\
PPO (w/o \texttt{finish}) & as PPO, with \texttt{finish} removed from the agent's action space \\
\midrule
Seeds / hardware & $42$--$45$ / $1\times$ NVIDIA L40 48\,GB \\
\bottomrule
\end{tabular}
\end{adjustbox}
\end{table}

\subsection{D1 grid-size scalability learned-policy hyperparameters}

The D1 scalability experiment compares PPO against grid-action A2C (Jumanji) and Sable on matched Jumanji-Connector v2 (Connector-v2) tasks: PPO consumes \kicad{} board files, while the grid-action baselines consume fixed Connector-v2 NPZ instances exported from the same splits.
\cref{tab:rq1-compute} lists the training settings that materially affect the RL results, with one hyperparameter family per row.
Final reporting evaluates each selected checkpoint on the exact 128 test boards per grid and seed.
The main-text figure (\cref{fig:rq1-instances-rout}) reports the single-rollout Rout.@1; the @5 diagnostics of \cref{tab:scale-std-rout} draw five rollouts per board and retain the selected rollout before aggregation.
For the grid-action baselines, training uses \texttt{train.npz}, training-time validation uses \texttt{val.npz}, and the final reported metric is recomputed on \texttt{test.npz}.
This split policy is important because the Connector-v2 NPZ files are exported from the same generated \kicad{} board splits used by the PPO agent.
The main figure reports the routability trend; \cref{tab:scale-std-rout} retains the per-seed routability breakdown.
For those diagnostics, A2C (Jumanji) and Sable wirelength is converted from grid steps to millimeters using the grid pitch, $100/G$ mm per cell.

\begin{table}[!htbp]
\centering
\caption{\textbf{D1 learned-policy hyperparameters.}
RL settings for the learned methods.
Each row isolates one training, architecture, or reward setting; ``--'' denotes a setting that is not used by that method.}
\label{tab:rq1-compute}
\small
\setlength{\tabcolsep}{2.5pt}
\renewcommand{\arraystretch}{1.05}
\begin{adjustbox}{max width=\linewidth}
\begin{tabular}{@{}
>{\raggedright\arraybackslash}p{0.19\linewidth}
>{\raggedright\arraybackslash}p{0.26\linewidth}
>{\raggedright\arraybackslash}p{0.26\linewidth}
>{\raggedright\arraybackslash}p{0.25\linewidth}@{}}
\toprule
\textbf{Setting} & \textbf{PPO} & \textbf{A2C (Jumanji)} & \textbf{Sable} \\
\midrule
\multicolumn{4}{@{}l}{\emph{Data and budget}} \\
Grids & 10, 50, 100, 200, 500 & 10, 50, 100, 200; 500 OOM & 10, 50, 100, 200, 500 \\
Seeds & 42--45 & 42--45 & 42--45 \\
Train split & 10K \kicad{} boards/grid & 10K Connector-v2 NPZ instances/grid & 10K Connector-v2 NPZ instances/grid \\
Eval protocol & 128 exact test boards/grid $\times$ 5 rollouts; selected rollout/board & 128 exact test NPZ instances/grid $\times$ 5 rollouts; selected rollout/board & 128 exact test NPZ instances/grid $\times$ 5 rollouts; selected rollout/board \\
Episode horizon & 256 & 256 & 256 \\
Native endpoint & 300 PPO iterations & D1-10: 8300 epochs; D1-50: 1600 epochs; D1-100: 5000 epochs; D1-200: 2200 epochs & D1-10: 1.58M updates; D1-50: 1.76M; D1-100: 1.46M; D1-200: 0.8125M; D1-500: 0.18M \\
\midrule
\multicolumn{4}{@{}l}{\emph{Optimization}} \\
Parallelism & 32 envs & total batch 256 & 16 envs \\
Rollout length & 512 & 10 & 128 \\
Update epochs & 4 PPO epochs & 100 learner steps/epoch & 4 PPO epochs \\
Minibatch/update batch & minibatch 256 & total batch 256 & update batch 2; 2 minibatches \\
Learning rate & $10^{-4}$ with 20-iter warmup & $2{\times}10^{-4}$ for D1-10/D1-50; $1.25{\times}10^{-5}$ for D1-100; $6.25{\times}10^{-6}$ for D1-200 & actor lr $2.5{\times}10^{-4}$ \\
Discount $\gamma$ & 0.995 & 1.0 & 0.99 \\
GAE / trace & GAE $\lambda=0.95$ & bootstrap factor 0.95 & GAE $\lambda=0.95$ \\
Policy clip & 0.2 & -- & 0.2 \\
Entropy coefficient & 0.01 & 0.01 & 0.01 \\
Value coefficient & 0.5 & TD loss weight 1.0 & 0.5 \\
Gradient clip & 0.5 & -- & 0.5 \\
\midrule
\multicolumn{4}{@{}l}{\emph{Architecture}} \\
Observation & \kicad{} token stream & Connector grid tensor & vector Connector observation \\
Backbone & token Transformer & Jumanji Connector A2C & feed-forward Sable with retention memory \\
Width & $d_{\rm model}=128$, FFN 512 & conv channels 32, MLP 512 & embedding 64 \\
Depth / heads & 4 layers, 8 heads & 4 encoder blocks, 8 heads, key size 16 & 1 block, 1 head \\
Action constraint & action masking with same-net bias & grid-action mask & grid-action policy \\
\midrule
\multicolumn{4}{@{}l}{\emph{Reward and accounting}} \\
Reward form & newly connected nets $-$ movement cost & newly connected nets $-$ movement cost & newly connected nets $-$ movement cost \\
Movement unit & wirelength increment in mm & grid-action step & grid-action step \\
Movement cost & $0.003 \times \Delta$WL(mm) & D1-10: 0.03; D1-50: 0.006; D1-100: 0.003; D1-200: 0.0015 & D1-10: 0.03; D1-50: 0.006; D1-100: 0.003; D1-200: 0.0015; D1-500: 0.0006 \\
Extra step cost & 0 & -- & -- \\
WL accounting unit & native mm & grid steps $\times\,100/G$ mm & grid steps $\times\,100/G$ mm \\
\bottomrule
\end{tabular}
\end{adjustbox}
\end{table}

\subsection{D1 training-time validation budgets}

The D1 learning curves are diagnostic W\&B validation logs, not the test-board rollout protocol used for the main results.
The PPO runs use the 300-iteration checkpoint batch and complete in 11.6--22.0 hours per seed/grid in W\&B runtime (\cref{tab:rq1-training-log-endpoints}).
The A2C (Jumanji) and Sable sweeps use fixed native endpoints chosen to match roughly one day of active training for the corresponding grid scale.
\cref{tab:rq1-training-log-endpoints} records these endpoints so that the cropped axes in \cref{fig:rq1-wandb-validation} are auditable.

\begin{table}[!htbp]
\centering
\caption{\textbf{D1 W\&B validation-log endpoints.}
Runtime is W\&B \texttt{\_runtime}, summarized as mean [min,max] hours across the listed seeds.
The native endpoint is the largest logged training coordinate used by the validation curves.
The displayed endpoint is the plotted coordinate after the figure's display-only clipping/scaling; for Sable this is the logged update counter divided by $100$.}
\label{tab:rq1-training-log-endpoints}
\footnotesize
\setlength{\tabcolsep}{6pt}
\renewcommand{\arraystretch}{1.04}
\begin{adjustbox}{max width=\linewidth}
\begin{tabular}{llcllll}
\toprule
Method & Subset & Seeds & Runtime (h) & Native endpoint & Displayed endpoint & State / note \\
\midrule
PPO & D1-10  & 42--45 & 19.5 [17.7, 22.0] & 300 iter & 300 & finished \\
PPO & D1-50  & 42--45 & 14.0 [13.7, 14.3] & 300 iter & 300 & finished \\
PPO & D1-100 & 42--45 & 13.3 [12.3, 15.9] & 300 iter & 300 & finished \\
PPO & D1-200 & 42--45 & 13.0 [11.6, 15.7] & 300 iter & 300 & finished \\
PPO & D1-500 & 42--45 & 13.7 [12.0, 16.7] & 300 iter & 300 & finished \\
\midrule
A2C (Jumanji) & D1-10  & 42--45 & approx. 24 & 8300 epochs & 5000 (clipped) & finished \\
A2C (Jumanji) & D1-50  & 42--45 & approx. 24 & 1600 epochs & 1600 & finished \\
A2C (Jumanji) & D1-100 & 42--45 & approx. 24 & 5000 epochs & 5000 & finished \\
A2C (Jumanji) & D1-200 & 42--45 & approx. 24 & 2200 epochs & 2200 & finished \\
A2C (Jumanji) & D1-500 & --     & -- & -- & -- & OOM \\
\midrule
Sable & D1-10  & 42--45 & approx. 24 & 1.58M updates & 15.8k & finished \\
Sable & D1-50  & 42--45 & approx. 24 & 1.76M updates & 17.6k & finished \\
Sable & D1-100 & 42--45 & approx. 24 & 1.46M updates & 14.6k & finished \\
Sable & D1-200 & 42--45 & approx. 24 & 0.8125M updates & 8.125k & finished \\
Sable & D1-500 & 42--45 & approx. 24 & 0.18M updates & 1.8k & finished \\
\bottomrule
\end{tabular}
\end{adjustbox}
\end{table}

\subsection{Training-time validation curves for D1 grid-size scalability}

\cref{fig:rq1-wandb-validation} visualizes the validation diagnostics logged during training for the learned policies in \cref{fig:rq1-instances}.
These curves are not the final reporting protocol: the main figure evaluates the selected checkpoints on the exact 128 test boards per grid and seed.
The plots instead show optimization dynamics under each method's native training axis, with seed mean and standard-deviation bands.

\begin{figure}[!htbp]
\centering
\includegraphics[width=.9\linewidth]{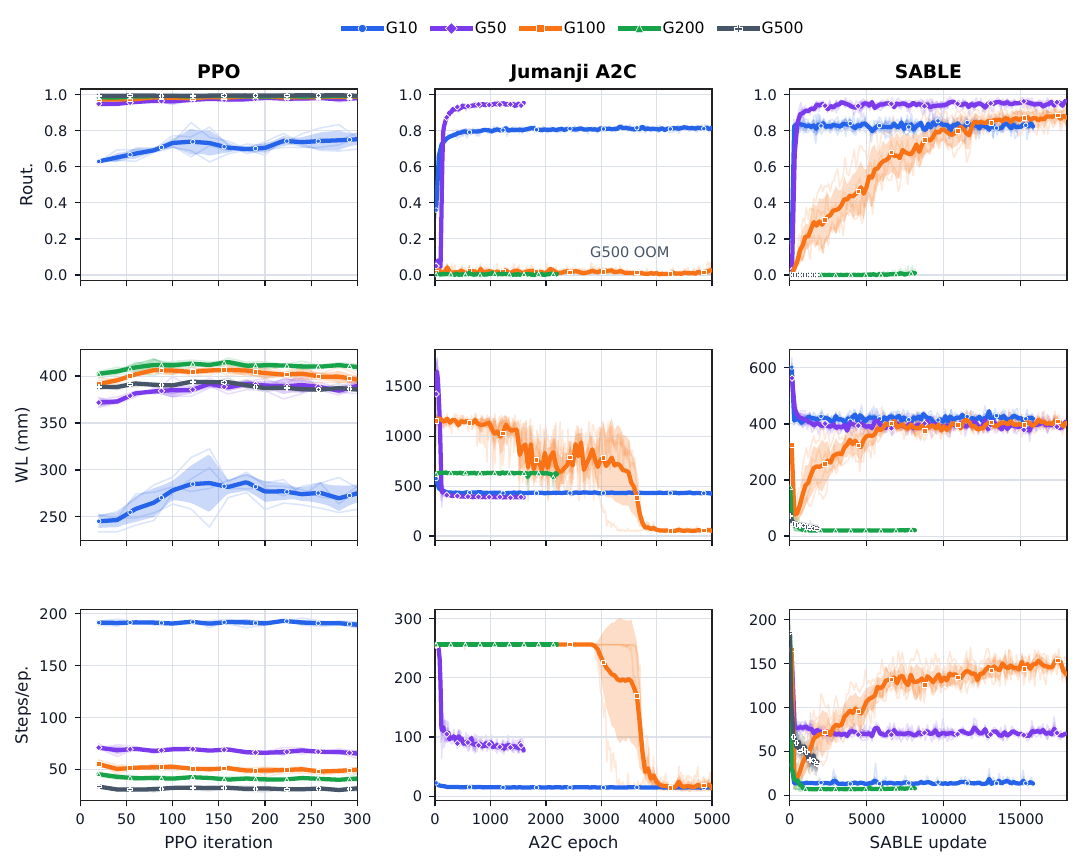}
\caption{\textbf{D1 grid-size scalability training-time validation diagnostics.} Columns separate
PPO, A2C (Jumanji), and Sable runs; rows show routability, wirelength, and episode length.
This four-seed plot uses the Connector-v2 NPZ split protocol and native endpoints; the A2C (Jumanji) axis is displayed up to 5000 epochs and the Sable update counter is divided by $100$ for readability.
Thick lines are seed means, translucent bands show one standard deviation, and faint traces show individual seeds.
These curves are diagnostic W\&B validation logs; the reported metrics are recomputed separately on the test-board rollouts.}
\label{fig:rq1-wandb-validation}
  \Description{\textbf{D1 grid-size scalability training-time validation diagnostics.} Columns separate PPO, A2C (Jumanji), and Sable runs; rows show routability, wirelength, and episode length.
This four-seed plot uses the Connector-v2 NPZ split protocol and native endpoints; the A2C (Jumanji) axis is displayed up to 5000 epochs and the Sable update counter is divided by $100$ for readability.
Thick lines are seed means, translucent bands show one standard deviation, and faint traces show individual seeds.
These curves are diagnostic W\&B validation logs; the reported metrics are recomputed separately on the test-board rollouts.}
\end{figure}
\subsection{Baseline Router Hyperparameters}
\label{app:baseline-hp}

We compare our learned policy against three rule-based routers: KiCadRoutingTools (KRT), OrthoRoute (a PathFinder-style negotiation router), and Freerouting~2.1.0.
Each router consumes its own preferred input format, so every input board is converted accordingly (Freerouting: DSN, OrthoRoute: ORP, KRT: native \kicad{} PCB) before the router is invoked, and the hyperparameters used for each board are recorded alongside the routing result.
All settings are taken from each router's \emph{default} configuration; no tuning is performed.
The \emph{Tunable} column indicates whether a parameter is freely user-configurable (\ding{51}) or hard-coded inside the router (\ding{55}).

\begin{table}[h]
\centering
\small
\caption{KiCadRoutingTools (KRT) hyperparameters.
Search-algorithm
parameters, cost-function weights, and geometry parameters are all exposed as CLI arguments of \texttt{route.py} and are therefore user-configurable.
Geometry parameters are set to match the design rules of each input board.}
\label{tab:hp-krt}
\begin{adjustbox}{max width=\linewidth}
\begin{tabular}{lllc}
\toprule
Group & Parameter & Value & Tunable \\
\midrule
\multirow{6}{*}{Search}
  & \texttt{grid\_step} (mm)             & 0.1            & \ding{51} \\
  & \texttt{max\_iterations}             & 200{,}000      & \ding{51} \\
  & \texttt{max\_probe\_iterations}      & 5{,}000        & \ding{51} \\
  & \texttt{max\_rip\_up\_count}         & 3              & \ding{51} \\
  & \texttt{heuristic\_weight}           & 1.9            & \ding{51} \\
  & \texttt{ordering\_strategy}          & \texttt{mps}   & \ding{51} \\
\midrule
\multirow{4}{*}{Cost}
  & \texttt{via\_cost}                   & 50             & \ding{51} \\
  & \texttt{via\_proximity\_cost}        & 10             & \ding{51} \\
  & \texttt{turn\_cost}                  & 1{,}000        & \ding{51} \\
  & \texttt{direction\_preference\_cost} & 50             & \ding{51} \\
\midrule
\multirow{5}{*}{Geometry}
  & \texttt{layers}                      & \{F.Cu, B.Cu\}                          & \ding{51} \\
  & \texttt{track\_width} (mm)           & matched to input board's design rules    & \ding{51} \\
  & \texttt{clearance} (mm)              & matched to input board's design rules    & \ding{51} \\
  & \texttt{via\_size} (mm)              & matched to input board's design rules    & \ding{51} \\
  & \texttt{via\_drill} (mm)             & matched to input board's design rules    & \ding{51} \\
\bottomrule
\end{tabular}
\end{adjustbox}
\end{table}

\begin{table}[h]
\centering
\small
\caption{OrthoRoute hyperparameters.
PathFinder-style negotiation
router~\citep{mcmurchie1995pathfinder}.
The first block (\emph{Exposed}) lists parameters that the user can change directly; the second block lists internal parameters that are fixed at the \texttt{PathFinderConfig} defaults.
We use the upstream defaults for all values.}
\label{tab:hp-ortho}
\begin{adjustbox}{max width=\linewidth}
\begin{tabular}{lllc}
\toprule
Group & Parameter & Value & Tunable \\
\midrule
\multirow{2}{*}{Exposed}
  & \texttt{max\_iterations}             & 250                  & \ding{51} \\
  & \texttt{use\_gpu}                    & \texttt{True}        & \ding{51} \\
\midrule
\multirow{9}{*}{PathFinder defaults}
  & \texttt{grid\_pitch} (mm)            & 0.4                  & \ding{55} \\
  & \texttt{pres\_fac\_init}             & 1.0                  & \ding{55} \\
  & \texttt{pres\_fac\_mult}             & 1.1                  & \ding{55} \\
  & \texttt{pres\_fac\_max}              & 64.0                 & \ding{55} \\
  & \texttt{hist\_gain}                  & 0.2                  & \ding{55} \\
  & \texttt{hist\_cost\_weight}          & 10.0                 & \ding{55} \\
  & \texttt{base\_cost\_weight}          & 0.3                  & \ding{55} \\
  & \texttt{portal\_discount}            & 0.4                  & \ding{55} \\
  & \texttt{span\_alpha}                 & 0.15                 & \ding{55} \\
\bottomrule
\end{tabular}
\end{adjustbox}
\end{table}

\paragraph{Grid resolution and design rules in OrthoRoute.}
PathFinder does not explicitly evaluate trace width or clearance; the grid spacing acts as an implicit minimum trace-to-trace distance.
We use the upstream default \texttt{grid\_pitch}\,$=0.4$\,mm, and we do \emph{not} enforce the board's design rule (\texttt{track\_width}\,$+$\,\texttt{clearance}) on \texttt{grid\_pitch}.
\texttt{grid\_pitch} is not exposed through the tool's interface, so changing it requires modifying the source code.
A single global pitch also cannot express boards whose netclass settings assign different clearances to different nets, as in the curated D3 boards.
Honoring per-net clearances in a uniform-grid router would require changing the routing algorithm itself rather than adjusting a parameter.
PathFinder also internally tunes a subset of its parameters (\texttt{pres\_fac\_*}, \texttt{hist\_*}, \texttt{max\_iterations}) on the fly based on per-board flexibility.

\begin{table}[h]
\centering
\small
\caption{Freerouting~2.1.0 hyperparameters.
We list only the CLI
options that we explicitly set.
Other options exposed by the JAR (e.g.\ \texttt{-mt} for the multi-threading level, \texttt{-pp} for the number of postroute passes) are left at their JAR defaults.}
\label{tab:hp-fr}
\begin{adjustbox}{max width=\linewidth}
\begin{tabular}{lllc}
\toprule
Group & Parameter & Value & Tunable \\
\midrule
\multirow{3}{*}{CLI}
  & jar version                            & \texttt{freerouting-2.1.0.jar} & \ding{51} \\
  & \texttt{-mp} (max passes)              & 10                              & \ding{51} \\
  & \texttt{-Xmx} (JVM heap)               & 4\,GiB                          & \ding{51} \\
\bottomrule
\end{tabular}
\end{adjustbox}
\end{table}

\paragraph{A note on iteration semantics.}
The three routers use mutually incomparable termination conditions: KRT's \texttt{max\_iterations} counts A* expansions across all rip-up attempts, OrthoRoute's \texttt{max\_iterations} counts PathFinder negotiation rounds, and Freerouting's \texttt{-mp} sets the maximum number of full ripup-and-reroute passes.
We therefore do not normalize these budgets and instead use each router's default termination condition.
\section{Additional results}\label{sec:add-results}

\paragraph{Per-seed std for D2/D3-A/D3-B routing quality.}
\cref{tab:rq2-std} reports mean$\pm$std for the evaluation that the main \cref{tab:rq2} summarizes with mean only, extended to the per-metric DRV/WL/Via breakdown.
\begin{table*}[!htbp]
\centering
\caption{\textbf{Routing-quality std breakdown (D2 / D3-A / D3-B).} Companion to the main \cref{tab:rq2}, adding DRV, WL (mm), and Via alongside CP, Pot., Rout., and Time.
Freerouting and the RL agents report mean$\pm$sample standard deviation ($n{-}1$) over 4 seeds.
Deterministic methods (OrthoRoute, KiCadRoutingTools) are run once, and the LLM agents run the five-rollout @5 protocol without seed repetition, so no seed-level std applies to either.
All metrics are computed on the potential-selected rollout, except Time, which is averaged over all five rollouts.}
\label{tab:rq2-std}
\small
\setlength{\tabcolsep}{4pt}
\begin{adjustbox}{width=\textwidth}
\begin{tabular}{ll ccccccc}
\toprule
Split & Method
& CP$\uparrow$
& Pot.$\uparrow$
& Rout.$\uparrow$
& DRV$\downarrow$
& WL$\downarrow$
& Via$\downarrow$
& Time$\downarrow$ \\
\midrule
\multirow{11}{*}{D2}
& Freerouting        & 1.00 $\pm$ 0.00 & 15.47 $\pm$ 0.07 & 1.00 $\pm$ 0.00 & 0.00 $\pm$ 0.00 & 393.9 $\pm$ 1.0 & 2.32 $\pm$ 0.03 & 2.69 $\pm$ 0.01 \\
& OrthoRoute         & 0.01 & 0.38 & 0.34 & 8.54 & 323.4 & 11.30 & 2.54 \\
& KiCadRoutingTools  & 1.00 & 15.52 & 1.00 & 0.00 & 373.9 & 8.09 & 0.82 \\
& GPT-5.4            & 0.96 & 16.05 & 1.00 & 0.05 & 401.4 & 1.61 & 94.04 \\
& GPT-5.4-mini       & 0.58 & 12.45 & 0.86 & 1.40 & 370.8 & 2.07 & 33.93 \\
& GPT-5.4-nano       & 0.55 & 12.23 & 0.85 & 1.45 & 364.6 & 1.91 & 63.39 \\
& Qwen3.5-397B       & 0.33 & 9.77 & 0.74 & 2.48 & 325.0 & 1.56 & 47.58 \\
& PPO                & 1.00 $\pm$ 0.00 & 16.24 $\pm$ 0.01 & 1.00 $\pm$ 0.00 & 0.00 $\pm$ 0.00 & 423.8 $\pm$ 2.8 & 1.70 $\pm$ 0.08 & 0.43 $\pm$ 0.04 \\
& GRPO               & 1.00 $\pm$ 0.00 & 14.03 $\pm$ 1.15 & 1.00 $\pm$ 0.00 & 0.00 $\pm$ 0.00 & 581.2 $\pm$ 52.4 & 6.31 $\pm$ 3.12 & 1.16 $\pm$ 0.13 \\
& PPO (terminal)     & 0.92 $\pm$ 0.03 & 15.61 $\pm$ 0.17 & 0.98 $\pm$ 0.01 & 0.16 $\pm$ 0.05 & 432.9 $\pm$ 4.5 & 1.63 $\pm$ 0.11 & 1.37 $\pm$ 0.09 \\
\cmidrule(lr){2-9}
& PPO (w/o \texttt{finish}) & 1.00 $\pm$ 0.00 & 16.21 $\pm$ 0.01 & 1.00 $\pm$ 0.00 & 0.00 $\pm$ 0.00 & 439.4 $\pm$ 6.1 & 1.48 $\pm$ 0.08 & 0.88 $\pm$ 0.06 \\
\midrule
\multirow{12}{*}{D3-A}
& Reference          & 1.00 & 23.16 & 1.00 & 0.00 & 170.2 & 1.18 & -- \\
& Freerouting        & 0.80 $\pm$ 0.01 & 22.71 $\pm$ 0.09 & 0.91 $\pm$ 0.00 & 0.86 $\pm$ 0.05 & 156.2 $\pm$ 0.2 & 0.74 $\pm$ 0.06 & 7.09 $\pm$ 0.26 \\
& OrthoRoute         & 0.02 & -6.16 & 0.53 & 99.70 & 129.8 & 26.97 & 2.20 \\
& KiCadRoutingTools  & 0.74 & 20.00 & 0.94 & 2.38 & 142.8 & 3.69 & 0.65 \\
& GPT-5.4            & 0.65 & 19.42 & 0.91 & 1.31 & 150.8 & 2.09 & 231.15 \\
& GPT-5.4-mini       & 0.28 & 14.22 & 0.72 & 3.88 & 127.0 & 0.39 & 56.85 \\
& GPT-5.4-nano       & 0.30 & 14.16 & 0.73 & 3.84 & 129.1 & 0.45 & 100.76 \\
& Qwen3.5-397B       & 0.34 & 14.88 & 0.75 & 3.63 & 121.0 & 0.47 & 81.82 \\
& PPO                & 0.86 $\pm$ 0.03 & 21.46 $\pm$ 0.20 & 0.95 $\pm$ 0.01 & 0.89 $\pm$ 0.09 & 165.4 $\pm$ 1.3 & 0.92 $\pm$ 0.29 & 1.83 $\pm$ 0.26 \\
& GRPO               & 0.85 $\pm$ 0.04 & 18.83 $\pm$ 1.50 & 0.98 $\pm$ 0.01 & 0.46 $\pm$ 0.15 & 279.0 $\pm$ 41.4 & 9.39 $\pm$ 4.52 & 3.54 $\pm$ 0.82 \\
& PPO (terminal)     & 0.82 $\pm$ 0.02 & 21.02 $\pm$ 0.41 & 0.95 $\pm$ 0.01 & 0.67 $\pm$ 0.09 & 171.3 $\pm$ 5.7 & 2.61 $\pm$ 1.62 & 2.58 $\pm$ 0.22 \\
\cmidrule(lr){2-9}
& PPO (w/o \texttt{finish}) & 0.94 $\pm$ 0.02 & 21.78 $\pm$ 0.10 & 0.99 $\pm$ 0.00 & 0.21 $\pm$ 0.07 & 224.1 $\pm$ 7.2 & 2.16 $\pm$ 0.65 & 3.05 $\pm$ 0.15 \\
\midrule
\multirow{12}{*}{D3-B}
& Reference          & 1.00 & 63.96 & 1.00 & 0.00 & 570.3 & 6.70 & -- \\
& Freerouting        & 0.78 $\pm$ 0.05 & 61.06 $\pm$ 0.57 & 1.00 $\pm$ 0.00 & 10.60 $\pm$ 0.35 & 535.4 $\pm$ 3.3 & 3.48 $\pm$ 0.22 & 9.94 $\pm$ 2.17 \\
& OrthoRoute         & 0.00 & -10.25 & 0.44 & 188.20 & 370.0 & 39.80 & 9.30 \\
& KiCadRoutingTools  & 0.20 & 40.94 & 0.86 & 44.70 & 472.3 & 11.80 & 3.27 \\
& GPT-5.4            & 0.00 & 34.72 & 0.62 & 17.30 & 347.1 & 3.60 & 865.83 \\
& GPT-5.4-mini       & 0.00 & 30.84 & 0.61 & 17.60 & 561.5 & 0.90 & 422.06 \\
& GPT-5.4-nano       & 0.00 & 30.66 & 0.59 & 18.60 & 460.0 & 0.30 & 510.25 \\
& Qwen3.5-397B       & 0.00 & 25.99 & 0.51 & 21.20 & 313.6 & 5.00 & 222.29 \\
& PPO                & 0.45 $\pm$ 0.10 & 50.20 $\pm$ 4.35 & 0.85 $\pm$ 0.06 & 8.28 $\pm$ 2.95 & 668.6 $\pm$ 64.1 & 6.17 $\pm$ 1.77 & 10.77 $\pm$ 3.51 \\
& GRPO               & 0.10 $\pm$ 0.00 & 30.00 $\pm$ 6.99 & 0.69 $\pm$ 0.09 & 15.82 $\pm$ 4.66 & 1147.6 $\pm$ 210.7 & 24.50 $\pm$ 9.42 & 11.20 $\pm$ 5.12 \\
& PPO (terminal)     & 0.38 $\pm$ 0.05 & 44.33 $\pm$ 4.28 & 0.81 $\pm$ 0.06 & 10.18 $\pm$ 2.90 & 667.4 $\pm$ 20.2 & 11.78 $\pm$ 3.40 & 12.60 $\pm$ 1.72 \\
\cmidrule(lr){2-9}
& PPO (w/o \texttt{finish}) & 0.42 $\pm$ 0.13 & 46.45 $\pm$ 1.36 & 0.87 $\pm$ 0.01 & 7.85 $\pm$ 0.81 & 1000.5 $\pm$ 126.4 & 10.20 $\pm$ 2.02 & 14.44 $\pm$ 1.69 \\
\bottomrule
\end{tabular}
\end{adjustbox}
\end{table*}

\paragraph{Single-rollout (@1) counterparts.}
\cref{tab:rq2-at1} rescores the same evaluation under the @1 protocol.
Every rollout is scored individually and each entry averages the per-rollout scores, so no best-of-five selection is applied.
Time is averaged over all five rollouts under both protocols, so it is unchanged from \cref{tab:rq2-std} and is omitted.
The method ordering of \cref{tab:rq2-std} is largely preserved, with only swaps among closely tied methods.
The margins widen, however, because the gain from selection varies widely across methods.
PPO and PPO (w/o \texttt{finish}) stay close to their @5 scores on D2, with CP moving from 1.00 to 0.98 and 1.00, whereas PPO (terminal) drops from 0.92 to 0.61 and GPT-5.4-mini falls from 0.58 to 0.20.
The same pattern holds zero-shot, where PPO (w/o \texttt{finish}) keeps a D3-A CP of 0.81 against 0.94 under @5 while the smaller LLM backbones lose about half of their clean passes.
Removing selection therefore separates methods that appear comparable under @5 without reordering them.
The @5 gains also show that the stochastic methods convert additional rollouts into better boards through selection, whereas the deterministic routers return a single solution that repeated runs cannot improve.
Since final board quality outweighs a constant factor of routing computation in EDA practice, spending a larger computation budget for a better solution is a practical advantage.

\begin{table*}[!htbp]
\centering
\caption{\textbf{Single-rollout (@1) routing quality (D2 / D3-A / D3-B).} Companion to \cref{tab:rq2-std} under the @1 protocol of \S\ref{sec:bench-metrics}.
Every rollout is scored without selection and each entry averages the per-rollout scores.
For the stochastic methods, $\pm$ is the sample standard deviation ($n{-}1$) across replicates, where a replicate is one rollout of one seed (20 replicates for Freerouting and the RL agents, 5 for the LLM agents, which run a single pass without seed repetition), whereas the $\pm$ of \cref{tab:rq2-std} is seed-level.
Deterministic methods (OrthoRoute, KiCadRoutingTools) are run once, so @1 and @5 coincide.
Time is averaged over all five rollouts under both protocols and is reported in \cref{tab:rq2-std}.
Best CP and Pot. per split (excluding the Reference row) are in \textbf{bold}.}
\label{tab:rq2-at1}
\small
\setlength{\tabcolsep}{4pt}
\begin{adjustbox}{max width=\textwidth}
\begin{tabular}{ll cccccc}
\toprule
Split & Method
& CP$\uparrow$
& Pot.$\uparrow$
& Rout.$\uparrow$
& DRV$\downarrow$
& WL$\downarrow$
& Via$\downarrow$ \\
\midrule
\multirow{11}{*}{D2}
& Freerouting        & \textbf{1.00} $\pm$ 0.00 & 14.42 $\pm$ 0.07 & 1.00 $\pm$ 0.00 & 0.00 $\pm$ 0.01 & 409.8 $\pm$ 2.1 & 2.48 $\pm$ 0.04 \\
& OrthoRoute         & 0.01 & 0.38 & 0.34 & 8.54 & 323.4 & 11.30 \\
& KiCadRoutingTools  & \textbf{1.00} & 15.52 & 1.00 & 0.00 & 373.9 & 8.09 \\
\cmidrule(lr){2-8}
& GPT-5.4            & 0.88 $\pm$ 0.02 & 15.24 $\pm$ 0.12 & 0.98 $\pm$ 0.00 & 0.27 $\pm$ 0.04 & 411.8 $\pm$ 2.5 & 2.00 $\pm$ 0.08 \\
& GPT-5.4-mini       & 0.20 $\pm$ 0.03 & 6.46 $\pm$ 0.45 & 0.49 $\pm$ 0.03 & 4.62 $\pm$ 0.27 & 208.6 $\pm$ 14.5 & 0.96 $\pm$ 0.10 \\
& GPT-5.4-nano       & 0.24 $\pm$ 0.03 & 6.97 $\pm$ 0.44 & 0.54 $\pm$ 0.03 & 4.21 $\pm$ 0.27 & 233.3 $\pm$ 11.3 & 1.25 $\pm$ 0.09 \\
& Qwen3.5-397B       & 0.14 $\pm$ 0.04 & 5.81 $\pm$ 0.40 & 0.48 $\pm$ 0.02 & 4.64 $\pm$ 0.19 & 229.6 $\pm$ 10.2 & 1.02 $\pm$ 0.06 \\
\cmidrule(lr){2-8}
& PPO                & 0.98 $\pm$ 0.02 & \textbf{15.84} $\pm$ 0.11 & 0.99 $\pm$ 0.01 & 0.06 $\pm$ 0.05 & 458.4 $\pm$ 9.0 & 2.22 $\pm$ 0.16 \\
& GRPO               & 0.95 $\pm$ 0.02 & 11.43 $\pm$ 1.15 & 0.98 $\pm$ 0.02 & 0.19 $\pm$ 0.14 & 744.3 $\pm$ 113.9 & 9.84 $\pm$ 3.57 \\
& PPO (terminal)     & 0.61 $\pm$ 0.07 & 12.06 $\pm$ 0.80 & 0.80 $\pm$ 0.06 & 1.60 $\pm$ 0.53 & 380.2 $\pm$ 33.8 & 1.62 $\pm$ 0.51 \\
\cmidrule(lr){2-8}
& PPO (w/o \texttt{finish}) & \textbf{1.00} $\pm$ 0.00 & 15.81 $\pm$ 0.09 & 1.00 $\pm$ 0.00 & 0.01 $\pm$ 0.01 & 497.1 $\pm$ 10.6 & 1.93 $\pm$ 0.20 \\
\midrule
\multirow{12}{*}{D3-A}
& Reference          & 1.00 & 23.16 & 1.00 & 0.00 & 170.2 & 1.18 \\
\cmidrule(lr){2-8}
& Freerouting        & 0.75 $\pm$ 0.01 & \textbf{22.04} $\pm$ 0.15 & 0.90 $\pm$ 0.00 & 1.08 $\pm$ 0.08 & 156.6 $\pm$ 1.0 & 0.86 $\pm$ 0.05 \\
& OrthoRoute         & 0.02 & -6.16 & 0.53 & 99.70 & 129.8 & 26.97 \\
& KiCadRoutingTools  & 0.74 & 20.00 & 0.94 & 2.38 & 142.8 & 3.69 \\
\cmidrule(lr){2-8}
& GPT-5.4            & 0.42 $\pm$ 0.03 & 16.15 $\pm$ 0.39 & 0.81 $\pm$ 0.02 & 2.71 $\pm$ 0.20 & 148.1 $\pm$ 5.7 & 2.36 $\pm$ 0.17 \\
& GPT-5.4-mini       & 0.13 $\pm$ 0.01 & 8.97 $\pm$ 0.20 & 0.49 $\pm$ 0.01 & 7.16 $\pm$ 0.18 & 81.9 $\pm$ 4.0 & 0.21 $\pm$ 0.05 \\
& GPT-5.4-nano       & 0.15 $\pm$ 0.01 & 8.35 $\pm$ 0.42 & 0.47 $\pm$ 0.02 & 7.46 $\pm$ 0.30 & 83.0 $\pm$ 6.1 & 0.31 $\pm$ 0.07 \\
& Qwen3.5-397B       & 0.19 $\pm$ 0.01 & 10.98 $\pm$ 0.43 & 0.59 $\pm$ 0.02 & 5.75 $\pm$ 0.31 & 102.7 $\pm$ 2.2 & 0.47 $\pm$ 0.08 \\
\cmidrule(lr){2-8}
& PPO                & 0.68 $\pm$ 0.03 & 19.22 $\pm$ 0.38 & 0.87 $\pm$ 0.02 & 1.91 $\pm$ 0.20 & 164.9 $\pm$ 3.9 & 1.09 $\pm$ 0.37 \\
& GRPO               & 0.64 $\pm$ 0.07 & 15.07 $\pm$ 1.82 & 0.90 $\pm$ 0.03 & 1.67 $\pm$ 0.52 & 337.1 $\pm$ 53.6 & 14.13 $\pm$ 5.17 \\
& PPO (terminal)     & 0.65 $\pm$ 0.03 & 18.23 $\pm$ 0.65 & 0.86 $\pm$ 0.02 & 1.95 $\pm$ 0.29 & 171.0 $\pm$ 7.5 & 3.50 $\pm$ 1.92 \\
\cmidrule(lr){2-8}
& PPO (w/o \texttt{finish}) & \textbf{0.81} $\pm$ 0.03 & 19.55 $\pm$ 0.40 & 0.94 $\pm$ 0.01 & 0.95 $\pm$ 0.26 & 262.7 $\pm$ 9.9 & 3.87 $\pm$ 1.03 \\
\midrule
\multirow{12}{*}{D3-B}
& Reference          & 1.00 & 63.96 & 1.00 & 0.00 & 570.3 & 6.70 \\
\cmidrule(lr){2-8}
& Freerouting        & \textbf{0.64} $\pm$ 0.09 & \textbf{58.86} $\pm$ 0.84 & 0.99 $\pm$ 0.01 & 11.94 $\pm$ 0.79 & 533.5 $\pm$ 7.3 & 3.88 $\pm$ 0.37 \\
& OrthoRoute         & 0.00 & -10.25 & 0.44 & 188.20 & 370.0 & 39.80 \\
& KiCadRoutingTools  & 0.20 & 40.94 & 0.86 & 44.70 & 472.3 & 11.80 \\
\cmidrule(lr){2-8}
& GPT-5.4            & 0.00 $\pm$ 0.00 & 25.65 $\pm$ 1.33 & 0.44 $\pm$ 0.02 & 24.88 $\pm$ 1.10 & 240.3 $\pm$ 21.8 & 2.06 $\pm$ 0.73 \\
& GPT-5.4-mini       & 0.00 $\pm$ 0.00 & 23.96 $\pm$ 0.85 & 0.49 $\pm$ 0.02 & 23.00 $\pm$ 1.14 & 408.6 $\pm$ 56.3 & 0.94 $\pm$ 0.27 \\
& GPT-5.4-nano       & 0.00 $\pm$ 0.00 & 23.62 $\pm$ 2.73 & 0.47 $\pm$ 0.06 & 24.00 $\pm$ 2.39 & 363.9 $\pm$ 45.7 & 0.36 $\pm$ 0.21 \\
& Qwen3.5-397B       & 0.00 $\pm$ 0.00 & 19.02 $\pm$ 0.93 & 0.41 $\pm$ 0.02 & 26.84 $\pm$ 0.74 & 249.6 $\pm$ 15.9 & 6.24 $\pm$ 3.77 \\
\cmidrule(lr){2-8}
& PPO                & 0.21 $\pm$ 0.14 & 37.60 $\pm$ 6.62 & 0.65 $\pm$ 0.11 & 17.33 $\pm$ 4.61 & 518.8 $\pm$ 94.9 & 5.06 $\pm$ 1.72 \\
& GRPO               & 0.07 $\pm$ 0.04 & 21.76 $\pm$ 6.35 & 0.55 $\pm$ 0.09 & 22.28 $\pm$ 4.40 & 1054.5 $\pm$ 169.8 & 27.26 $\pm$ 8.97 \\
& PPO (terminal)     & 0.15 $\pm$ 0.07 & 33.47 $\pm$ 3.82 & 0.64 $\pm$ 0.06 & 17.54 $\pm$ 2.75 & 550.0 $\pm$ 53.3 & 10.64 $\pm$ 4.01 \\
\cmidrule(lr){2-8}
& PPO (w/o \texttt{finish}) & 0.22 $\pm$ 0.08 & 35.99 $\pm$ 4.06 & 0.72 $\pm$ 0.07 & 15.32 $\pm$ 3.17 & 1008.6 $\pm$ 138.4 & 14.57 $\pm$ 4.71 \\
\bottomrule
\end{tabular}
\end{adjustbox}
\end{table*}

\paragraph{Grid-action baseline (Jumanji).}
Jumanji~\citep{jumanjibench} instantiates the simplest form of PCB routing under the grid-action abstraction: each net is an agent stepping on a neighbor grid in four directions from a source pad to a target pad, and any visited cell becomes unusable for all agents (single-occupancy).
The episode reward is the number of connected nets minus $\lambda$ times the total wirelength.
With every net configured as a 2-pad connection of unit-cell width, we adopt the grid-action agents trained in this environment, A2C (Jumanji) and Sable, as baselines and compare them against the PPO agent on identical instances.

\paragraph{\kicad{}-API treatment of grid scale.}
Under the grid-action abstraction, enlarging the grid preserves the local four-neighbor move set but expands both the state space and the routing horizon, so the same physical instance becomes a strictly harder MDP.
The \kicad{}-API abstraction sidesteps this by treating track width and clearance as \emph{environment parameters} rather than as state: the agent issues geometric net-segment commands whose action dimensionality is independent of the grid resolution, and grid scaling only changes the routing margin available to each net.
The D1 grid-size scalability experiment isolates exactly this design choice on otherwise identical instances.

\paragraph{Quantitative breakdown and horizon argument.}
Our \kicad{}-API agent's CP rises from 0.63 on D1-10 to 1.00 from D1-50 to D1-500 (selected-rollout routability 0.90 to 1.00; the single-rollout Rout.@1 in \cref{fig:rq1-instances-rout} is correspondingly lower, e.g.\ 0.77 on D1-10), tracking the geometric easing of the instance as each unit-cell-width net occupies a smaller fraction of the board.
The grid-action baselines display the opposite profile: A2C (Jumanji) and Sable are competitive on D1-10 and D1-50, but A2C collapses by D1-100 and runs out of memory at D1-500, while Sable remains viable at D1-100 (CP 0.73) but collapses at D1-200 and D1-500.
This is not a lack of wall-clock effort. The grid-action baselines consume far more environment steps through JAX/vectorized training.
Rather, the experiment isolates the core abstraction gap. Finer grids make the physical board easier, but they lengthen the local-move MDP horizon that grid-action agents must explore, whereas the \kicad{}-API action space keeps the decision horizon tied to routed segments rather than cells.

\paragraph{Per-seed std for D1 grid-size scalability.}
Companion to \cref{fig:rq1-instances} in the main text: per-seed mean$\pm$std for the learned D1 methods on the exact 128 test boards per grid and seed.
The main figure reports the single-rollout Rout.@1; \cref{tab:scale-std-rout} additionally keeps clean-pass, potential-gain, and wirelength diagnostics on the selected rollout.
Each board is evaluated with five rollouts, then one selected rollout (largest potential gain) is retained before aggregation.
CP requires full connectivity with zero error-level DRVs on this selected rollout; routability divides routed nets by the five two-pin nets in each Connector-v2 board.
WL is absolute routed wirelength in millimeters; A2C (Jumanji) and Sable wirelength is converted from grid steps using the grid pitch, $100/G$ mm per cell.
WL values with Rout.$<0.5$ are retained for auditability but are not comparable compactness estimates because incomplete episodes measure only produced traces.

\begin{table}[!htbp]
\centering
\caption{\textbf{D1 grid-size sweep across action abstractions.}
Mean$\pm$std across seeds (per-board rollout selected by Pot.) on the 128 Connector-v2 test boards per grid.
Metrics are clean pass (CP), potential gain (Pot.), routability (Rout.), and routed wirelength (WL, in mm; grid-action baselines converted from grid steps via the grid pitch $100/G$\,mm/cell).
A2C (Jumanji) on D1-500 exceeded memory (OOM).
WL at Rout.$<0.5$ reflects only partial traces and is not a comparable compactness estimate.}
\label{tab:scale-std-rout}
\setlength{\tabcolsep}{4pt}
\begin{adjustbox}{max width=\linewidth}
\begin{tabular}{ll cccc}
\toprule
Split & Method & CP$\uparrow$ & Pot.$\uparrow$ & Rout.$\uparrow$ & WL$\downarrow$ \\
\midrule
\multirow{3}{*}{D1-10}
& PPO  & 0.63 $\pm$ 0.01 & 4.75 $\pm$ 0.02 & 0.90 $\pm$ 0.00 & 344.2 $\pm$ 5.2 \\
& A2C (Jumanji) & 0.51 $\pm$ 0.02 & 4.22 $\pm$ 0.12 & 0.87 $\pm$ 0.01 & 407.1 $\pm$ 6.5 \\
& Sable & 0.46 $\pm$ 0.02 & 4.04 $\pm$ 0.11 & 0.86 $\pm$ 0.01 & 382.3 $\pm$ 2.7 \\
\midrule
\multirow{3}{*}{D1-50}
& PPO  & 1.00 $\pm$ 0.00 & 6.92 $\pm$ 0.03 & 1.00 $\pm$ 0.00 & 370.1 $\pm$ 2.6 \\
& A2C (Jumanji) & 0.91 $\pm$ 0.00 & 6.51 $\pm$ 0.03 & 0.98 $\pm$ 0.00 & 462.2 $\pm$ 6.3 \\
& Sable & 0.88 $\pm$ 0.02 & 6.38 $\pm$ 0.11 & 0.97 $\pm$ 0.01 & 370.9 $\pm$ 3.0 \\
\midrule
\multirow{3}{*}{D1-100}
& PPO  & 1.00 $\pm$ 0.00 & 6.95 $\pm$ 0.00 & 1.00 $\pm$ 0.00 & 363.0 $\pm$ 1.2 \\
& A2C (Jumanji) & 0.00 $\pm$ 0.00 & -5.47 $\pm$ 0.15 & 0.03 $\pm$ 0.02 & 76.7 $\pm$ 6.6 \\
& Sable & 0.73 $\pm$ 0.13 & 5.59 $\pm$ 0.71 & 0.94 $\pm$ 0.04 & 379.1 $\pm$ 7.2 \\
\midrule
\multirow{3}{*}{D1-200}
& PPO  & 1.00 $\pm$ 0.00 & 6.92 $\pm$ 0.02 & 1.00 $\pm$ 0.00 & 368.7 $\pm$ 8.5 \\
& A2C (Jumanji) & 0.00 $\pm$ 0.00 & -5.72 $\pm$ 0.02 & 0.01 $\pm$ 0.00 & 156.7 $\pm$ 1.8 \\
& Sable & 0.00 $\pm$ 0.00 & -5.69 $\pm$ 0.09 & 0.01 $\pm$ 0.01 & 25.7 $\pm$ 1.6 \\
\midrule
\multirow{3}{*}{D1-500}
& PPO  & 1.00 $\pm$ 0.00 & 6.95 $\pm$ 0.00 & 1.00 $\pm$ 0.00 & 358.1 $\pm$ 1.5 \\
& A2C (Jumanji) & \multicolumn{4}{c}{OOM} \\
& Sable & 0.00 $\pm$ 0.00 & -5.79 $\pm$ 0.00 & 0.00 $\pm$ 0.00 & 28.4 $\pm$ 6.9 \\
\bottomrule
\end{tabular}
\end{adjustbox}
\end{table}

\paragraph{Per-seed std for PPO/GRPO reward-training analysis.}
Per-seed mean$\pm$std for PPO, GRPO, and PPO (terminal) is part of \cref{tab:rq2-std}. \cref{tab:rq5-time-dist} reports the board-level timing distribution.

\paragraph{Checkpoint provenance for reward analysis.}
The reward sweep in \cref{fig:rq4-factorial} uses the existing 9-cell $\times$ 4-seed checkpoint set from the \oursbench{} experiments; it is not a new training sweep.
The default PPO (per-step) cell with wire penalty $0.002$, via penalty $0.1$, and seed 42 uses the best checkpoint from a completed same-configuration recovery run because the original symbolic-link target for that seed was unavailable.
All reward-analysis metrics are still recomputed with the same selected-rollout and \kicad{} evaluator protocol used for the other seeds.

\paragraph{Training-time validation curves for PPO/GRPO analysis.}
\cref{fig:rq5-wandb-validation} reports the W\&B diagnostics logged while training the three PPO/GRPO analysis policies.
As with the D1 scalability curves, these curves are included to show optimization behavior, not to define the reported metrics.
Main \cref{tab:rq2} and \cref{tab:rq2-std} use the selected @5 rollout artifacts, with quality metrics recomputed from saved PCBs using the \kicad{} evaluator.
We do not plot a training-time CP curve here because the logged validation scalars are not consistently the CP quantity used in the tables. CP requires the joint condition of full connectivity and zero \kicad{} error-level DRVs on the same selected route.

\begin{figure}[!htbp]
\centering
\includegraphics[width=\linewidth]{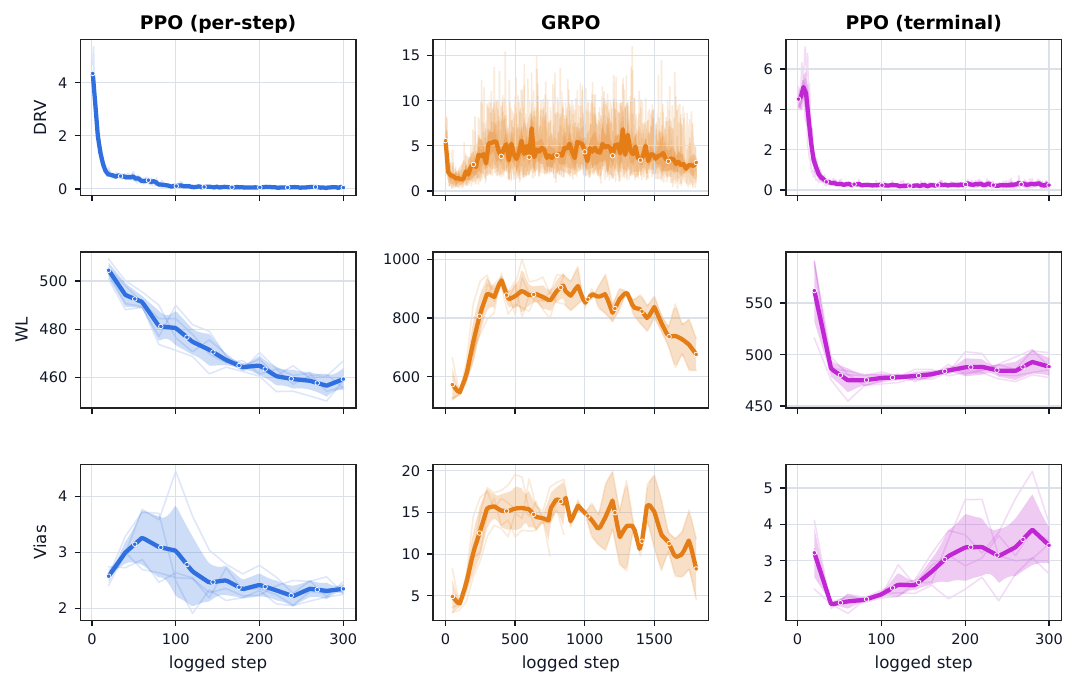}
\caption{\textbf{PPO/GRPO training-time validation diagnostics.} Columns correspond
to PPO (per-step), GRPO, and PPO (terminal). Rows show DRV, wirelength, and via diagnostics.
Curves are W\&B logged values over the logged training step or evaluation iteration; thick lines are seed means, translucent bands show one standard deviation, and faint traces show individual seeds.
Final \cref{tab:rq2} CP and quality values are computed separately from the selected routed PCBs.}

\label{fig:rq5-wandb-validation}
  \Description{\textbf{PPO/GRPO training-time validation diagnostics.} Columns correspond to PPO (per-step), GRPO, and PPO (terminal). Rows show DRV, wirelength, and via diagnostics.
Curves are W\&B logged values over the logged training step or evaluation iteration; thick lines are seed means, translucent bands show one standard deviation, and faint traces show individual seeds.
Final \cref{tab:rq2} CP and quality values are computed separately from the selected routed PCBs.}
\end{figure}

\begin{table}[!htbp]
\centering
\caption{\textbf{Episode timing distribution for the RL agents.} Board-level latency distribution for the Time entries in main \cref{tab:rq2}.
Timing uses the recorded per-board rollout time and includes board reload/reset, policy rollout, and routed-PCB serialization.
D3-A excludes \texttt{0096\_karabas-nano\_wifi\_revA}, the one board that cannot be loaded
through the engine interface (Appendix~\ref{sec:d3-splits}).}
\label{tab:rq5-time-dist}
\setlength{\tabcolsep}{3pt}
\begin{adjustbox}{max width=\linewidth}
\begin{tabular}{llccccc}
\toprule
Dataset & Configuration & Boards & Mean & Median & P95 & Max \\
\midrule
\multirow{4}{*}{D2}
  & PPO    & 128 & 0.43 & 0.37 & 0.92 & 1.42 \\
  & GRPO  & 128 & 1.16 & 0.94 & 2.94 & 3.67 \\
  & PPO (terminal) & 128 & 1.37 & 1.20 & 2.96 & 3.85 \\
  & PPO (w/o \texttt{finish}) & 128 & 0.88 & 0.74 & 1.82 & 3.16 \\
\midrule
\multirow{4}{*}{D3-A}
  & PPO & 99 & 1.83 & 1.04 & 5.39 & 15.00 \\
  & GRPO & 99 & 3.54 & 3.09 & 8.55 & 12.98 \\
  & PPO (terminal) & 99 & 2.58 & 1.88 & 5.91 & 8.78 \\
  & PPO (w/o \texttt{finish}) & 99 & 3.05 & 1.90 & 8.44 & 24.62 \\
\midrule
\multirow{4}{*}{D3-B}
  & PPO & 10 & 10.77 & 9.31 & 18.23 & 19.98 \\
  & GRPO & 10 & 11.20 & 10.44 & 14.76 & 15.15 \\
  & PPO (terminal) & 10 & 12.60 & 10.96 & 22.49 & 26.08 \\
  & PPO (w/o \texttt{finish}) & 10 & 14.44 & 15.23 & 19.48 & 20.82 \\
\bottomrule
\end{tabular}
\end{adjustbox}
\end{table}

PPO is usually fast on D3-A (median 1.04\,s, mean 1.83\,s), with the tail driven by a few hard real boards that run to the 256-step horizon and trigger expensive \kicad{} geometry updates during rollout and serialization.
On the larger D3-B boards all four configurations run several times slower (means 10.8--14.4\,s, three to six times their D3-A means).

\subsection{Throughput and Token Usage in LLM Experiments}\label{sec:llm-cost}

For the OpenAI LLM runs, we record wall-clock time and token consumption per episode.
\cref{tab:llm-cost-synth2L,tab:llm-cost-pcbench,tab:llm-cost-pcbench-medium} report the per-episode means: latency in seconds, and three token streams, \emph{System} (system-prompt tokens accumulated across all turns), \emph{User} (user-message tokens, dominated by the per-step state-of-board observation), and \emph{Response} (model-generated tokens, including any hidden reasoning the API counts on the input/output side).
\emph{Total} is the sum of the three.

\begin{table}[h]
\centering\small
\caption{Per-episode latency and token usage on D2
(128 boards $\times$ 5 rollouts = 640 episodes).
Tokens are means over completed episodes.}
\label{tab:llm-cost-synth2L}
\begin{adjustbox}{max width=\linewidth}
\begin{tabular}{@{}lrrrrr@{}}
\toprule
\textbf{Model} & \textbf{Sec/ep} & \textbf{System} & \textbf{User} & \textbf{Response} & \textbf{Total}\\
\midrule
GPT-5.4                   &  94.0 & 62,899 & 39,909 & 3,823 & 106,631 \\
GPT-5.4-mini              &  33.9 & 51,538 & 26,438 & 2,382 &  80,358 \\
GPT-5.4-nano              &  63.4 & 66,140 & 38,124 & 4,260 & 108,524 \\
\bottomrule
\end{tabular}
\end{adjustbox}
\end{table}

\begin{table}[h]
\centering\small
\caption{Per-episode latency and token usage on D3-A
(99 boards $\times$ 5 rollouts = 495 episodes).
Tokens are means over completed episodes.}
\label{tab:llm-cost-pcbench}
\begin{adjustbox}{max width=\linewidth}
\begin{tabular}{@{}lrrrrr@{}}
\toprule
\textbf{Model} & \textbf{Sec/ep} & \textbf{System} & \textbf{User} & \textbf{Response} & \textbf{Total}\\
\midrule
GPT-5.4                   & 231.1 & 263,612 & 152,583 & 8,683 & 424,879 \\
GPT-5.4-mini              &  56.9 & 156,258 &  57,796 & 3,525 & 217,578 \\
GPT-5.4-nano              & 100.8 & 199,424 &  92,200 & 6,667 & 298,291 \\
\bottomrule
\end{tabular}
\end{adjustbox}
\end{table}

\begin{table}[h]
\centering\small
\caption{Per-episode latency and token usage on D3-B
(10 boards $\times$ 5 rollouts = 50 episodes).
Tokens are means over completed episodes.}
\label{tab:llm-cost-pcbench-medium}
\begin{adjustbox}{max width=\linewidth}
\begin{tabular}{@{}lrrrrr@{}}
\toprule
\textbf{Model} & \textbf{Sec/ep} & \textbf{System} & \textbf{User} & \textbf{Response} & \textbf{Total}\\
\midrule
GPT-5.4                   & 865.8 & 1,119,161 & 775,289 & 26,717 & 1,921,167 \\
GPT-5.4-mini              & 422.1 & 1,147,573 & 918,588 & 20,134 & 2,086,296 \\
GPT-5.4-nano              & 510.3 & 1,117,479 & 861,073 & 29,945 & 2,008,496 \\
\bottomrule
\end{tabular}
\end{adjustbox}
\end{table}









\section{LLM prompts}\label{sec:prompts}

\subsection{\ours{} agent}\label{sec:prompt-agent}

Each step's prompt is assembled from the template in \cref{fig:kicadgym-agent-prompt}, partitioned into the static system message and the per-step user message as described in Appendix~\ref{sec:llm_wrapper}.
Placeholders with curly-braces are populated at runtime from the environment state.

\paragraph{Placeholders.} The runtime values used throughout the
experiments in this paper are:
\begin{itemize}\itemsep0pt
\item \texttt{\{StateFormatDesc\}} -- schema description of the board-state encoding.
We use the S-expression encoding (\texttt{state\_format=sexpr}); the alternative XML encoding is supported but not used in the reported runs.
\item \texttt{\{CurrentStep\}} -- 1-indexed step counter, ranging from $1$ to the per-episode budget $T$.
We set $T = 200$ for D2, D3-A, and D3-B.
\item \texttt{\{CurrentObservation\}} -- concatenation of the \texttt{routing\_geometry} and \texttt{router\_head} blocks, regenerated every step; all floating-point coordinates are rounded to $3$ decimal places (mm).
\item \texttt{\{ValidStepCount\}} -- number of accepted (parsed \emph{and} mask-valid) actions issued so far in the episode; bounded above by \texttt{\{CurrentStep\}}.
\item \texttt{\{ActionHistory\}} -- rolling window of the most recent \texttt{history\_length}$=2$ turns, each rendered as a \texttt{<think>}/\texttt{<action>} pair; entries that left the unrouted-pin count unchanged are prefixed with \texttt{[no effect]}.
\item \texttt{\{RejectedAttempts\}} -- conditional slot, empty string in the nominal case.
When the previous action was rejected (parse failure or mask veto) it expands to a single line of the form \texttt{Step F-L: <body> (rejected $\times$N)}; consecutive identical bodies are collapsed and the streak is cleared by the next valid action.
\item \texttt{\{ValidActions\}} -- the action verbs allowed by the current router phase mask (\texttt{net\_select} $\to$ \texttt{start\_route} $\to$ routing); the parser's index set has seven entries, the six action types plus the internal \texttt{idle} fallback at index $6$ (Appendix~\ref{sec:llm_wrapper}).
\end{itemize}

\begin{tcblisting}{promptbox}
You are a PCB routing agent for KiCad. Connect unconnected pins (points in routing_geometry) within each net. Never connect pins from different nets.

# Priority
1. Complete all connections (reduce unconnected points to zero).
2. Avoid DRC violations (no crossing other nets, respect clearance).
3. Minimize total wirelength.

# Routing Guidelines
- Targets are the (point ...) entries in routing_geometry; the final segment must land on the target pad's (x, y) while the cursor is on that pad's layer (points carry no layer -- look up the pad in board_static or the nearby tracks).
- Use exact coordinates from the observation (pad positions, point targets, track endpoints). Only invent coordinates for detour waypoints.
- Avoid static obstacles, board edges, and existing tracks on the same layer (any net, including the current one).
- When a path is blocked, detour: switch layers with `make_via` (offset >=1mm from any pad -- never place a via on a pad), route around on the opposite layer, then switch back.
- When the cursor is already near the remaining target with a clear path, prefer `finish <mode>` over a manual `make_line`.

## Layer-change detour pattern
Route from (0, 0) on layer 1 to (10, 10) on layer 1, bypassing an obstacle via layer 2:
  1. start_route 0.0 0.0 1
  2. make_via   0.0 0.5 w        # offset off the source pad, then cross to layer 2
  3. start_route 0.0 0.5 2       # re-enter routing on the new layer
  4. make_via   10.0 9.5 w       # STOP before the target's (x, y) and cross back
  5. start_route 10.0 9.5 1      # re-enter routing on the target pad's layer
  6. make_line  10.0 10.0 w      # final segment lands on the pad, on its own layer

## Do NOT draw or end on a pad from the wrong layer
Never let `make_line` or `make_via` reach the target pad's exact (x, y) while the cursor is on a different layer than that pad. If you do, you are trapped:
  - placing a via at the pad violates the "no via on pad" rule, and
  - the pad is unreachable from the wrong layer, so `finish` will fail repeatedly.
Always make the layer-return `make_via` at an offset waypoint (>=1mm away from the target pad) *before* the track arrives at the target's (x, y).

# Board State
{StateFormatDesc}

## Step {CurrentStep}
{CurrentObservation}

# History (step {CurrentStep}, {ValidStepCount} valid actions taken)
{ActionHistory}{RejectedAttempts}

# Valid Actions
Choose exactly one from:
{ValidActions}

# Response Format
Output exactly one <think>...</think> followed by one <action>...</action> with a valid action. Never repeat think-action pairs.
"[no effect]" in history = action made no progress; "Last rejected attempt" below = action was refused. Do not repeat either.
\end{tcblisting}
\captionof{figure}{Full prompt template for the \ours{} agent.}
\label{fig:kicadgym-agent-prompt}

\subsection{Baselines}\label{sec:llm_baselines}
\paragraph{Engine-free (open-loop) generation.} An LLM directly generates the routed board in \kicad{} board syntax: given the initial board state, it emits the wire segments and vias that are inserted into the \texttt{.kicad\_pcb} file.

\begin{tcblisting}{promptbox}
You are an expert PCB routing engineer using KiCad PCB format.

Your task is to generate valid PCB routing (tracks and vias) for a given KiCad PCB board.

## Instructions
- Analyze the given PCB layout, including components, pads, and nets.
- Generate routing (segments and vias) that correctly connects pads belonging to the same net.
- Ensure:
  - No short circuits between different nets
  - Minimal via usage unless necessary
  - Reasonable routing paths (avoid unnecessary detours)
  - Respect layer usage (F.Cu, B.Cu)

## Routing geometry: OCTILINEAR ROUTING with the 45-degree-ONLY constraint (mandatory)
Use **octilinear routing** for every `(segment ...)`. Octilinear routing is a wire/edge routing style where connections are restricted to **eight directions**: the four cardinal (horizontal, vertical) plus the four diagonals at 45 degrees. It sits between rectilinear routing (4 directions, Manhattan-style) and fully arbitrary Euclidean routing.

### The 45-degree-only constraint (sub-rule of octilinear)
Whenever a segment is *not* horizontal or vertical, it MUST be a **strict 45 (or 135) degree diagonal** -- i.e. exactly |dx| == |dy|. No other diagonal angle is permitted. The following are all VIOLATIONS:
    30, 60        (e.g. (0,0) -> (10, 5.77) or (0,0) -> (5, 8.66))
    22.5, 67.5    ("half-octilinear" angles)
    13.16, 26.6   (or any other arbitrary slope)
Even tiny rounding deviations break the rule: `(start 0 0) (end 10 9.9)` is NOT octilinear. If you intend a 45-degree diagonal, the rise must equal the run exactly.

### Allowed segment shapes (and only these)
Concretely, a segment from `(start sx sy)` to `(end ex ey)` is octilinear iff one of the following holds, with `dx = ex - sx` and `dy = ey - sy`:
    1. dy == 0                          (East / West   -- 0,   horizontal)
    2. dx == 0                          (North / South -- 90,  vertical)
    3. dx ==  dy  (and both nonzero)    (NE  / SW      -- 45,  diagonal)
    4. dx == -dy  (and both nonzero)    (NW  / SE      -- 135, diagonal)
Equivalently, every segment is horizontal, vertical, or a 45-degree diagonal where |dx| == |dy|. Anything else is forbidden.

### How to handle non-octilinear paths
If the natural route between two pads is at an angle that is not in {0, 45, 90, 135} degrees, decompose it into multiple octilinear segments joined at corners. For example, to go from (0,0) to (10,3) you might use:
    (segment ... (start 0 0) (end 3 3) ...)   ; 45-degree diagonal
    (segment ... (start 3 3) (end 10 3) ...)  ; horizontal
Use as many segments as needed; each one must individually satisfy the rule above.

This is exactly the constraint KiCad's interactive PNS router enforces with `corner_mode = MITERED_45`.

## Output Format
- Return ONLY the routing additions in valid KiCad PCB format.
- Do NOT repeat the full board.
- Only include `(segment ...)` and `(via ...)` entries.
\end{tcblisting}
\captionof{figure}{Full prompt template for the engine-free (open-loop) generation.}
\label{fig:cad-gen-prompt}

\paragraph{Plan-and-execute (open-loop) generation.} An LLM emits the full \ours{} action sequence ahead of execution, without observing any intermediate board state during generation.

\begin{tcblisting}{promptbox}
You are an expert PCB routing engineer driving a KiCad routing API.

You will be given the *initial* state of a PCB board (footprints, pads, nets) and must output the **complete sequence of routing API calls** that would route every net.

## API
The router exposes 6 high-level actions. Emit one per line.
    net_select <net_id>                      Select a net to start routing.
    start_route <x_mm> <y_mm> <layer>        Begin routing at a pad position.
                                             layer: 1 = F.Cu (top), 2 = B.Cu (bottom).
    make_line <x_mm> <y_mm> <mode>           Extend a track to (x,y) on the current layer.
    make_via <x_mm> <y_mm> <mode>            Extend track to (x,y), drop a via, switch layer.
                                             AFTER make_via, the current layer flips.
    finish <mode>                            Auto-complete the active route on the *CURRENT*
                                             layer to the nearest pending pad of the
                                             current net. CANNOT cross layers — if the
                                             remaining pad is on a different layer, finish
                                             will NOT reach it.
    net_end                                  Mark the active net as fully routed.
    mode is one of: m (MarkObstacles), p (PushAndShove), w (Walkaround). Use w by default.

## Reading <BOARD>
The input is KiCad-style sexpr. Inside `(nets ...)` each net lists its pads:
    (pad <id> <x> <y> <layer_tag>)
Layer tags:
    `1`  -> the pad lives on layer 1 (F.Cu, top).
    `2`  -> the pad lives on layer 2 (B.Cu, bottom).
    `th` -> through-hole, electrically present on BOTH layers; you
           may treat it as either 1 or 2 when choosing start_route's
           layer parameter or as the via-side of a make_via.

## Coordinates and layers (MUST follow exactly)
- Copy each pad's `<x>` and `<y>` token **verbatim** — same digits, same decimal places, no rounding. The router only accepts a pad's *exact* (x_mm, y_mm). Example: if the board has
    (pad D0 67.500 44.700 th)
  then output
    start_route 67.500 44.700 1
  not `start_route 67.5 44.7 1`.
- start_route MUST be issued at a pad position of the currently selected net.

## Layer-choice rule for the FIRST pad
You pick a `<layer>` argument for `start_route`. The choice determines which layer the router head sits on, which constrains every subsequent action.

Pick the start layer using THIS priority:
  1. If the first pad has tag `1` -> use 1.
  2. If the first pad has tag `2` -> use 2.
  3. If the first pad has tag `th` and at least one other pad of
     this net has tag `1` or `2` -> use that other pad's layer.
     (This avoids an unnecessary via.)
  4. If the first pad has tag `th` and all other pads also `th`
     -> use 1.

## Routing protocol (state machine — violations cause the entire net to fail)
After `net_select`, the router is in {has_net=True, is_routing=False}.
After a SUCCESSFUL `start_route`, it moves to {has_net=True, is_routing=True}.
`make_line`, `make_via`, `finish` are ONLY valid while is_routing=True.
`net_end` closes the net and returns to {has_net=False}.
Emitting any of make_line / make_via / finish / net_end before a
successful start_route causes the entire net to fail.

## Per-topology emission rules (CASE BY CASE — match the net's shape)
Resolve each pad's *effective* layer using the layer-choice rule (`th` resolves to either 1 or 2 by neighbour). Then dispatch:

### Case A — 0 or 1 pads : skip the net entirely.

### Case B — 2 pads on the SAME effective layer L
  net_select <id>
  start_route <P1.x> <P1.y> L
  make_line  <P2.x> <P2.y> <mode>   ; recommended over finish
  net_end

  (alternative: `finish <mode>` also works — auto-completes head to P2 on L)

### Case C — 2 pads on DIFFERENT layers (cross-layer)
  `finish` cannot cross layers. Pick an intermediate point P1' and go
  P1 -> P1' -> P2, using make_via at P1' to switch layers:
  net_select <id>
  start_route <P1.x> <P1.y> L1        ; L1 = P1's layer
  make_via   <P1'.x> <P1'.y> <mode>   ; switch L1 -> L2 at P1'
  start_route <P1'.x> <P1'.y> L2      ; L2 = P2's layer, restart at P1'
  make_line  <P2.x> <P2.y> <mode>     ; (or `finish <mode>`)
  net_end

### Case D — k pads (k >= 3) all on the SAME effective layer L
  net_select <id>
  start_route <P1.x> <P1.y> L
  make_line  <P2.x> <P2.y> <mode>
  start_route <P2.x> <P2.y> L
  make_line  <P3.x> <P3.y> <mode>
  ... up to P(k-1) ...
  make_line  <P(k).x> <P(k).y> <mode>   ; (or `finish <mode>`)
  net_end

## Anti-patterns (do NOT do these)
  ❌ `finish` immediately after start_route when the second pad is
     on a different layer. finish will route within the current
     layer and silently miss the cross-layer pad.
     ✓ Use make_via to land on the cross-layer pad (Case C).
  ❌ Picking a `<layer>` for start_route that no pad of this net
     lives on (e.g. layer 2 when both pads are on 1).
     ✓ Apply the layer-choice rule.
  ❌ Reformatting coordinates (5.7 instead of 5.700, dropping
     trailing zeros). The board's pad table has the exact form;
     copy it.
  ❌ Emitting `make_line <Pk.x> <Pk.y>` THEN `finish` for the
     final pad of a same-layer chain. `finish` already handles Pk
     — the explicit make_line creates a duplicate / overlapping
     segment and may break DRC.
  ❌ Forgetting `net_end` between nets. The next `net_select`
     fails if the previous net wasn't closed.
  ❌ Calling start_route at coordinates that are not a pad of the
     currently selected net (e.g. picking a pad of a different
     net by accident). The route fails silently.

## Output Format
Wrap the entire sequence in <actions>...</actions>. One action per line. No commentary inside the block. Do NOT repeat the board.
\end{tcblisting}
\captionof{figure}{Full prompt template for the plan-and-execute (open-loop) generation.}
\label{fig:api-seq-prompt}

\section{Visualization gallery and LLM failure traces}\label{sec:gallery}

\begin{figure*}[!htbp]
    \centering
    \includegraphics[width=1\linewidth]{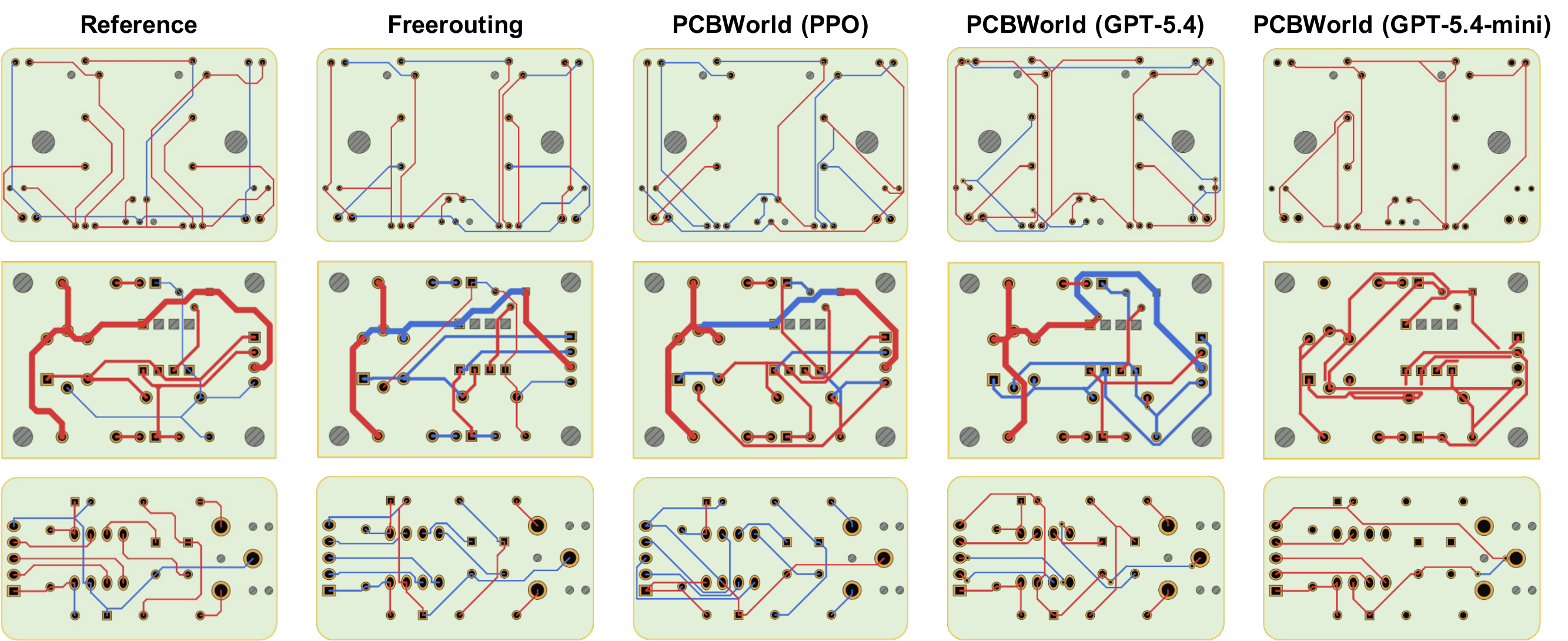}
    \caption{\textbf{Examples of routed boards (D3-A).} Full version of the main-text \cref{fig:qualitative-main}, showing all three boards. Red and blue traces mark wire segments on the two copper layers.}
    \label{fig:qualitative-full}
  \Description{\textbf{Examples of routed boards (D3-A).} Full version of the main-text figure, showing all three boards. Red and blue traces mark wire segments on the two copper layers.}
\end{figure*}

\paragraph{Importance of net selection order.} Across the studied rollouts, \texttt{net\_select}, nominally a simple routing-target selection action, emerges as a critical strategic decision whose first invocation often determines whether a board can be successfully routed (see \cref{fig:think-expl1}).
In the \texttt{0018\_hy\_adapter} case, successful episodes consistently begin with selecting a routing order that minimizes early routing difficulty, whereas unsuccessful episodes choose a more challenging net at the outset and subsequently fail to recover within the limited step budget.
Since the agent rarely performs corrective recovery behaviors after an unfavorable decision, the initial \texttt{net\_select} effectively constrains the future routing space and strongly influences the overall routing outcome.
These observations suggest that routing-target ordering is a primary factor governing downstream routing success.

\begin{figure}[h]
  \centering
  \includegraphics[width=0.8\linewidth]{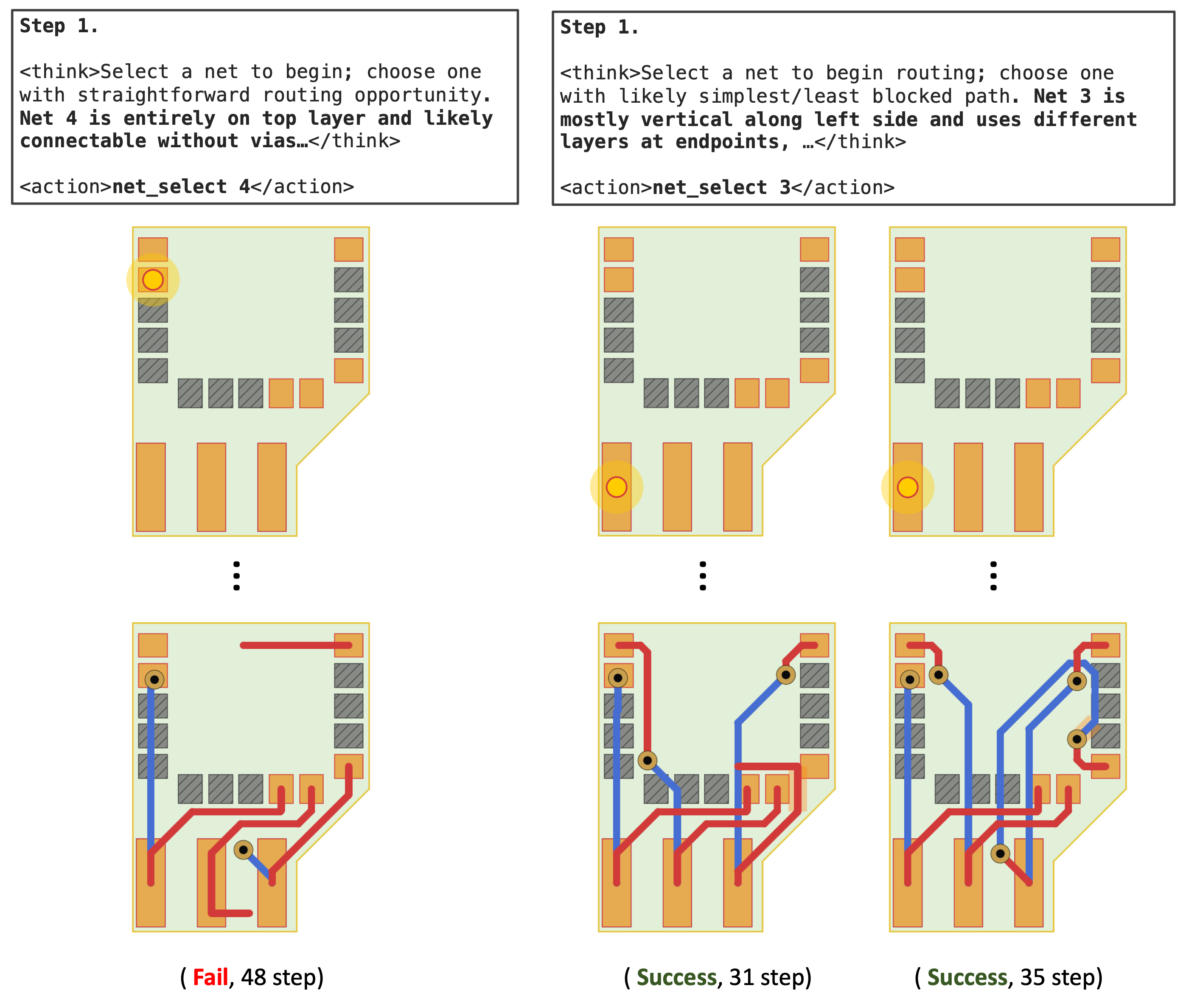}
  \caption{\textbf{Case Study 1 (\texttt{0018\_hy\_adapter}).} In this action design, determining the routing priority of nets is critically important.
Although different nets were selected for different reasons, we observe that routing succeeds when a particular net (net 3) is selected first, whereas it fails when another net (net 4) is selected first.}
  \label{fig:think-expl1}
  \Description{\textbf{Case Study 1 (\texttt{0018\_hy\_adapter}).} In this action design, determining the routing priority of nets is critically important.
Although different nets were selected for different reasons, we observe that routing succeeds when a particular net (net 3) is selected first, whereas it fails when another net (net 4) is selected first.}
\end{figure}

\paragraph{Importance of route editing.} The agent's action space (\cref{tab:action-description}) contains no operation that edits committed copper.
\ours{} does expose track-removal APIs (\texttt{delete\_track\_*}, \texttt{delete\_via\_*}) among the Action methods cataloged in Appendix~\ref{sec:action-apilist}, but we exclude them from the agent's action space to keep it compact and to scope the task to routing itself.
The agent can push existing traces aside through the \texttt{push\_n\_shove} routing mode, yet it cannot delete or redraw a trace it has already committed.
As a consequence, the agent tends to struggle when severe congestion occurs, since it has limited capability to resolve routing deadlocks after they emerge.
Nevertheless, we observe that successful routing remains achievable when the agent proactively avoids congestion through early-stage planning and anticipatory reasoning.

\cref{fig:think-expl2} analyzes successful and failed cases on (\texttt{0100\_smt-zvs-driver\_IH10-mc}).
Since the action space does not include direct editing of pre-existing tracks, once congestion is detected, the agent tends to repeatedly attempt rerouting through alternative paths.
In contrast, when the agent proactively anticipates potential congestion and initiates routing from a different region in advance, it can achieve more effective routing with fewer steps.

\begin{figure}[h]
  \centering
  \includegraphics[width=0.8\linewidth]{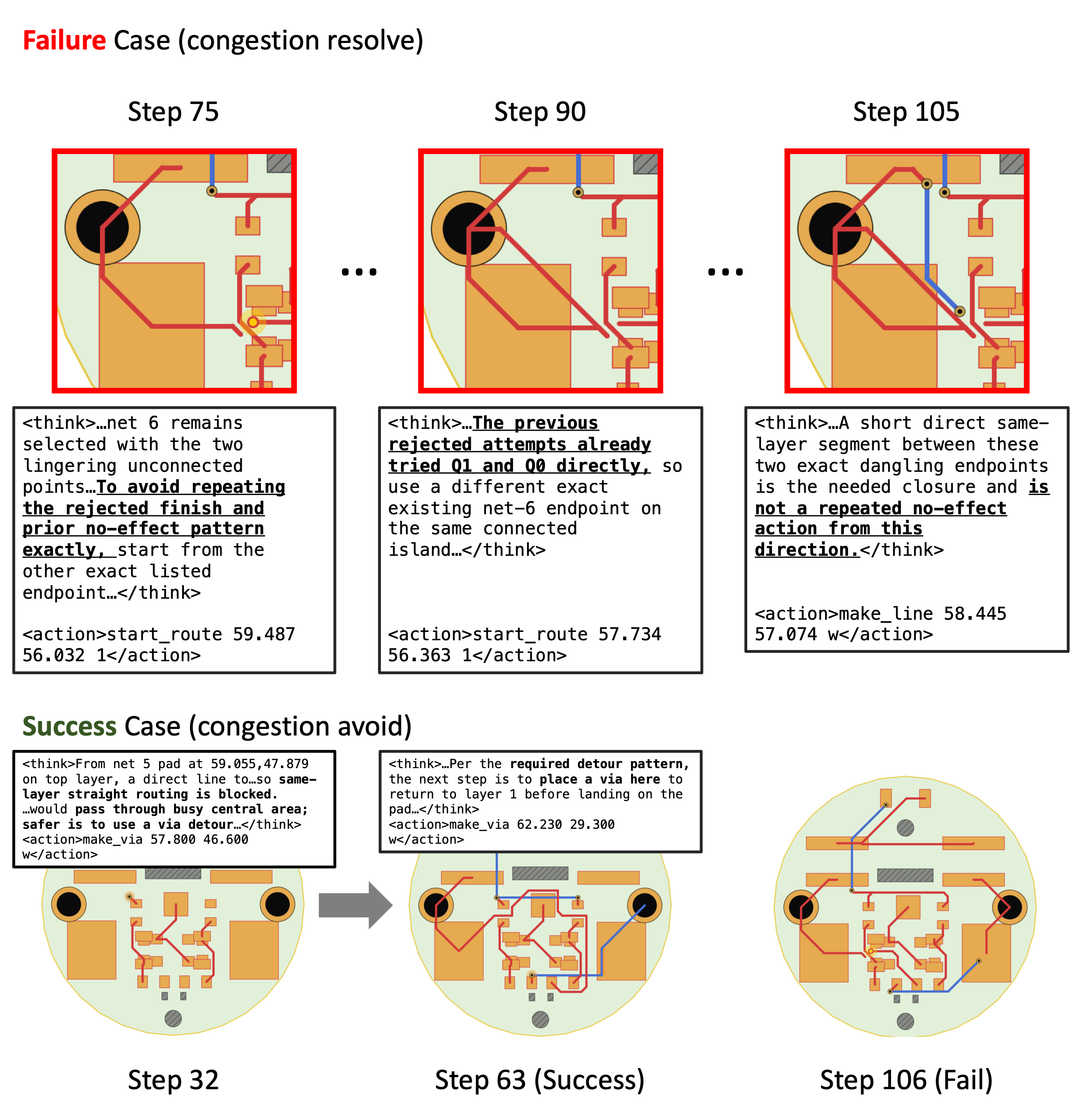}
  \caption{\textbf{Case Study 2 (\texttt{0100\_smt-zvs-driver\_IH10-mc}).} Failure case (top) and success case (bottom).
Since the action space does not include direct editing of pre-existing tracks, once congestion is encountered, the agent repeatedly attempts rerouting through alternative paths.
In contrast, proactively anticipating congestion and initiating routing from a different region enables more effective routing with fewer steps.}
  \label{fig:think-expl2}
  \Description{\textbf{Case Study 2 (\texttt{0100\_smt-zvs-driver\_IH10-mc}).} Failure case (top) and success case (bottom).
Since the action space does not include direct editing of pre-existing tracks, once congestion is encountered, the agent repeatedly attempts rerouting through alternative paths.
In contrast, proactively anticipating congestion and initiating routing from a different region enables more effective routing with fewer steps.}
\end{figure}

\clearpage

\section{Datasheet for Datasets}\label{sec:datasheet}

This appendix summarizes \oursbench{} using the main categories of the Datasheets for Datasets template~\citep{datasheet}.
The benchmark is a collection of \kicad{}-native PCB routing instances used with the engine-grounded evaluation protocol described in \S\ref{sec:bench}.

\subsection{Motivation}

\oursbench{} was created to evaluate PCB routing methods scored by the same engine-checked evaluator. DRC status, connectivity, wirelength, and via count are all computed by \kicad{}.
The dataset targets comparison among tool-using LLM agents, RL agents, grid-action baselines, and rule-based routers, rather than supervised imitation of a single reference solution.
It contains no human-subject records or personal data.

\subsection{Composition}

Each instance is stored as a \texttt{.kicad\_pcb} file containing the board geometry, placed pads/components, nets, design rules, and the initial routing state.
The benchmark has three task families: D1 is a 1-layer grid-routing family converted from Jumanji Connector-style instances~\citep{jumanjibench}; D2 is the Synthetic Gridless Board family, a 2-layer multi-pin family with configurable net and pad counts; and D3 is a 679-board curated set derived from PCBench's open-source PCB corpus~\citep{pcbench_github}, normalized to the \kicad{}~9 format, populated with each board's stored design rules, and retained only when fully connected and DRV-free.
D1 and D2 each provide 10{,}000/128/128 train/valid/test instances (D2-train/D2-valid/D2-test for D2), while D3 is reserved for zero-shot evaluation and stratified by pad count into D3-A (Small),
D3-B (Medium), and D3-C (Large). \cref{tab:bench-stats} summarizes the per-family statistics; the stratification
rule, the per-split sizes, and the boards used for evaluation are given in Appendix~\ref{sec:d3-splits}.

\subsection{Collection Process}

D1 is produced by converting grid instances into executable \kicad{} boards.
D2 is generated from a parametric synthetic board generator.
D3 starts from PCBench, a public PCB routing dataset with per-board \texttt{raw.kicad\_pcb} and \texttt{processed.kicad\_pcb} files, PCB-RDL \texttt{final.json} routing descriptions, metadata, data augmentation scripts, and an RL environment~\citep{pcbench_github}.
We use the \kicad{}-9-converted \texttt{PCBs\_version\_9} pool as the raw D3 source and apply the compatibility filtering below.
No crowdworkers or paid participants were involved.

\subsection{Preprocessing}\label{sec:datasheet-preprocessing}

The D3 preprocessing step addresses \kicad{}-version DRC compatibility rather than changing PCB geometry.
The \texttt{PCBs\_version\_9} source pool contains 1{,}182 boards converted to \kicad{}~9; many were originally authored in \kicad{}~4--7, so conversion assigns newer \kicad{}~9 default DRC constraints that can flag violations even when the physical board file is unchanged.
We therefore apply a project-file compatibility pass to \texttt{.kicad\_pro} files only: non-routing manufacturing/courtyard checks such as \texttt{solder\_mask\_bridge},   \texttt{drill\_out\_of\_range},  \texttt{malformed\_courtyard}, and \texttt{courtyards\_overlap} are ignored, and \kicad{}-9 floor constraints for annular width, copper-edge clearance, and hole clearance are relaxed to match earlier-version board compatibility.
Core routing constraints used for evaluation, including clearance, shorts/crossings, track width, via diameter, and the source-board netclass settings, are preserved.
Track width is always present in the source settings; when a board stores no explicit clearance value, the clearance is instead set to the minimum spacing observed among its own routed traces.
 
After this compatibility pass, 679 boards remain in the curated D3 set.
The pass does not edit the physical \texttt{.kicad\_pcb} content: a byte-level comparison confirmed that all 679 retained board files are unchanged before and after preprocessing.
The changes are limited to DRC rule interpretation in the project file, and the retained boards are revalidated through the same \kicad{} engine interface used for evaluation.

\begin{figure}[th]
\centering
\includegraphics[width=0.7\linewidth]{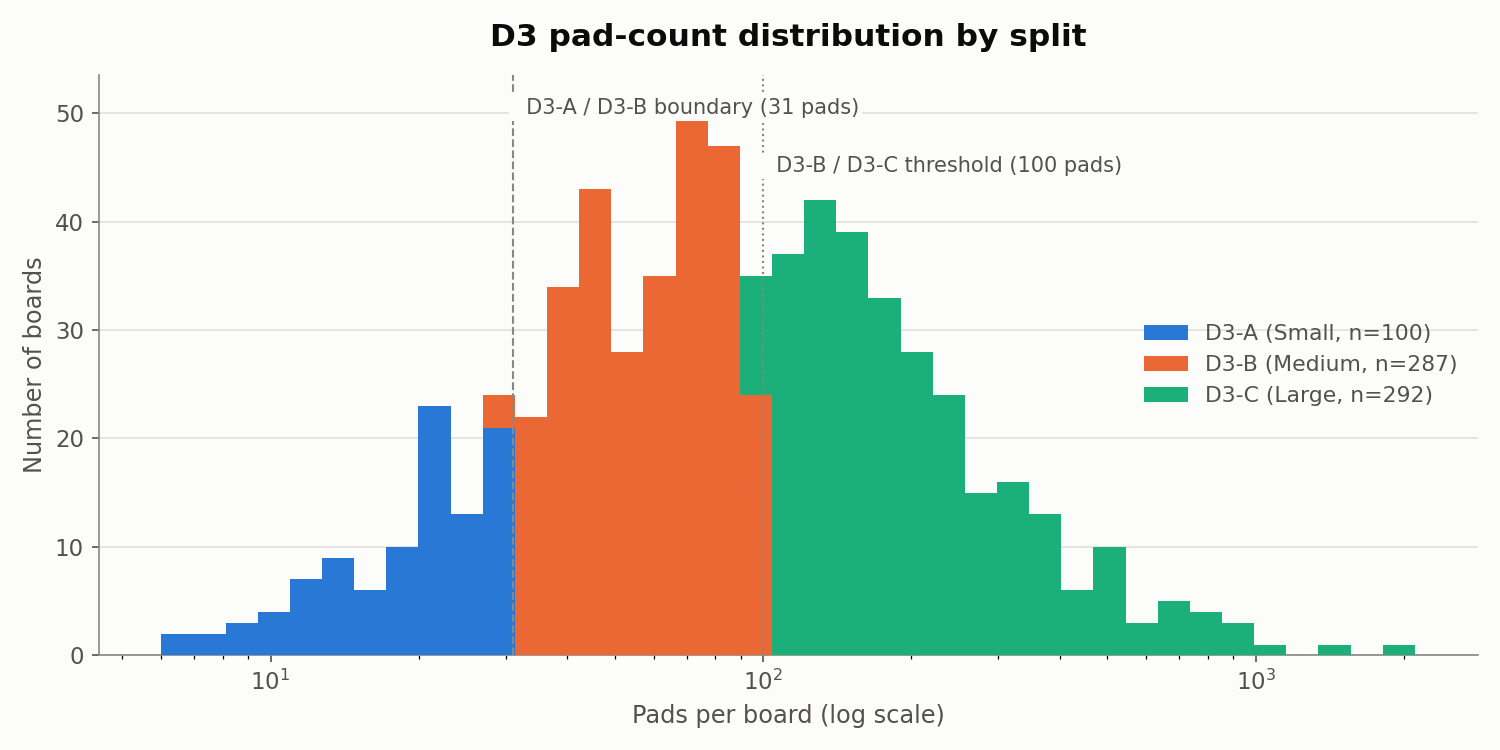}
\caption{Pad-count distribution of the 679 curated D3 boards, stacked by split (log-scaled
horizontal axis). The dashed line marks the D3-A/D3-B boundary induced by the 100-row prefix of the
pad-sorted characteristics table ($31$ pads); the dotted line marks the $100$-pad D3-B/D3-C
threshold.}
\label{fig:d3-pad-dist}
\Description{Pad-count distribution of the 679 curated D3 boards, stacked by split (log-scaled
horizontal axis). The dashed line marks the D3-A/D3-B boundary induced by the 100-row prefix of the
pad-sorted characteristics table (31 pads); the dotted line marks the 100-pad D3-B/D3-C threshold.}
\end{figure}

\subsection{Splits and Evaluation Subsets of D3 Boards}
\label{sec:d3-splits}

\paragraph{Ordering and stratification.}
The 679 curated boards that survive the compatibility pass of Appendix~\ref{sec:datasheet-preprocessing} are summarized by a
characteristics table with one row per board, recording its net, component, pad, and copper-layer
counts. Rows are sorted by pad count in non-decreasing order, so row order coincides with increasing
routing difficulty. The pad count bounds the number of terminals a routing method must connect and
correlates with both the net count and the total route length. We stratify this sorted table with two
parameters and no random component. D3-A (Small) is the 100-row prefix of the table, which
corresponds to the boards with at most 31 pads; among the remaining 579 boards, those with at most
100 pads form D3-B (Medium, 287 boards) and those with more than 100 pads form D3-C (Large, 292
boards). The three splits are pairwise disjoint and together cover all 679 boards.
\cref{fig:d3-pad-dist} shows the resulting pad-count distribution together with both
boundaries; \cref{tab:bench-stats} reports the per-split net, pad, and layer ranges.

\paragraph{Evaluation subsets.}
Zero-shot evaluation uses a fixed subset of each split rather than the full split, because per-board
cost is dominated by the LLM agents and grows steeply with board size. A single D3-A episode already
averages $231.1$\,s and $424{,}879$ tokens for GPT-5.4 (\cref{tab:llm-cost-pcbench}), and the RL
agents run three to six times slower per episode on D3-B than on D3-A (\cref{tab:rq5-time-dist}). For D3-A we evaluate
$99$ of the $100$ boards: \texttt{0096\_karabas-nano\_wifi\_revA} is excluded because it cannot be
loaded through the engine interface, and it is the only board in the split for which this occurs. For
D3-B and D3-C we select ten boards each with the following deterministic procedure, applied
identically to both splits:
\begin{enumerate}
  \item keep only the two-layer boards, so that every selected instance is handled by all baseline
        routers and stays comparable to the two-layer D2 boards on which the RL policies are trained;
  \item for D3-C only, discard the largest $10\%$ of the remaining boards by pad count, whose pad
        counts reach into the thousands (\cref{tab:bench-stats}) and would otherwise place a single extreme outlier
        in the top bin;
  \item sort the survivors by (pad count, net count, board identifier), which makes the order total
        and independent of file-system enumeration;
  \item partition the sorted sequence into ten equal-size quantile bins;
  \item take the lower-median board of each bin.
\end{enumerate}
The procedure involves no random component: the bins and their medians follow from the characteristics
table and the parameters above. The ten selected boards therefore span the pad-count range of their
split instead of concentrating at its mode.
\cref{fig:d3bc-picks} shows the pad-count distribution of the two pools the subsets are drawn
from, together with the selected boards, and \cref{tab:d3-eval-subsets} lists them.

\begin{table}[t]
\centering
\scriptsize
\setlength{\tabcolsep}{4pt}
\begin{tabular}{cll}
\toprule
Bin & D3-B (Medium) & D3-C (Large) \\
\midrule
1  & \texttt{0113\_maytal\_Maytal}                            & \texttt{0400\_laptimer58\_Chickadee} \\
2  & \texttt{0144\_ottawa-badges-2016\_ottawa-badge-tagger-2016} & \texttt{0423\_induction-hob\_temperature-sender} \\
3  & \texttt{0174\_sensorboard\_DiffIR}                       & \texttt{0448\_uedaino\_uedaino} \\
4  & \texttt{0203\_MOD-MPU9150\_mod-mpu9150}                  & \texttt{0473\_data-manager\_data-manager} \\
5  & \texttt{0232\_ATtiny461Breakout\_ATTiny461DevBoard}      & \texttt{0496\_kitspace\_f-91w} \\
6  & \texttt{0260\_NiMH-Charger\_NiMH Charger}                & \texttt{0520\_RGBMatrixPanelCPLD-PhotonBackpack\_RGBMatrixPanel\_CPLD\_negative} \\
7  & \texttt{0288\_raspberrypi-3-usb-hub\_usb\_hub}           & \texttt{0546\_ozinverter\_ozinverterkicad} \\
8  & \texttt{0316\_kicad-workshop\_fancyboard}                & \texttt{0571\_Brushless\_ESC\_Brushless\_ESC} \\
9  & \texttt{0344\_mavbridge\_mavbridge}                      & \texttt{0599\_RGBMatrixPanelCPLD-PhotonBackpack\_RGBMatrixPanel\_CPLD} \\
10 & \texttt{0376\_cat-trainer\_teensy\_base\_pcb} & \texttt{0626\_TOBS\_HybridChargeController} \\
\bottomrule
\end{tabular}
\caption{Ten-board evaluation subsets of D3-B and D3-C, one board per quantile bin, ordered by
increasing pad count. All twenty boards are two-layer by construction. D3-B is the subset used in the
main results; the D3-C subset is fixed for reference and is not evaluated in this paper.}
\label{tab:d3-eval-subsets}
\end{table}


\begin{figure}[ht]
\centering
\includegraphics[width=0.7\linewidth]{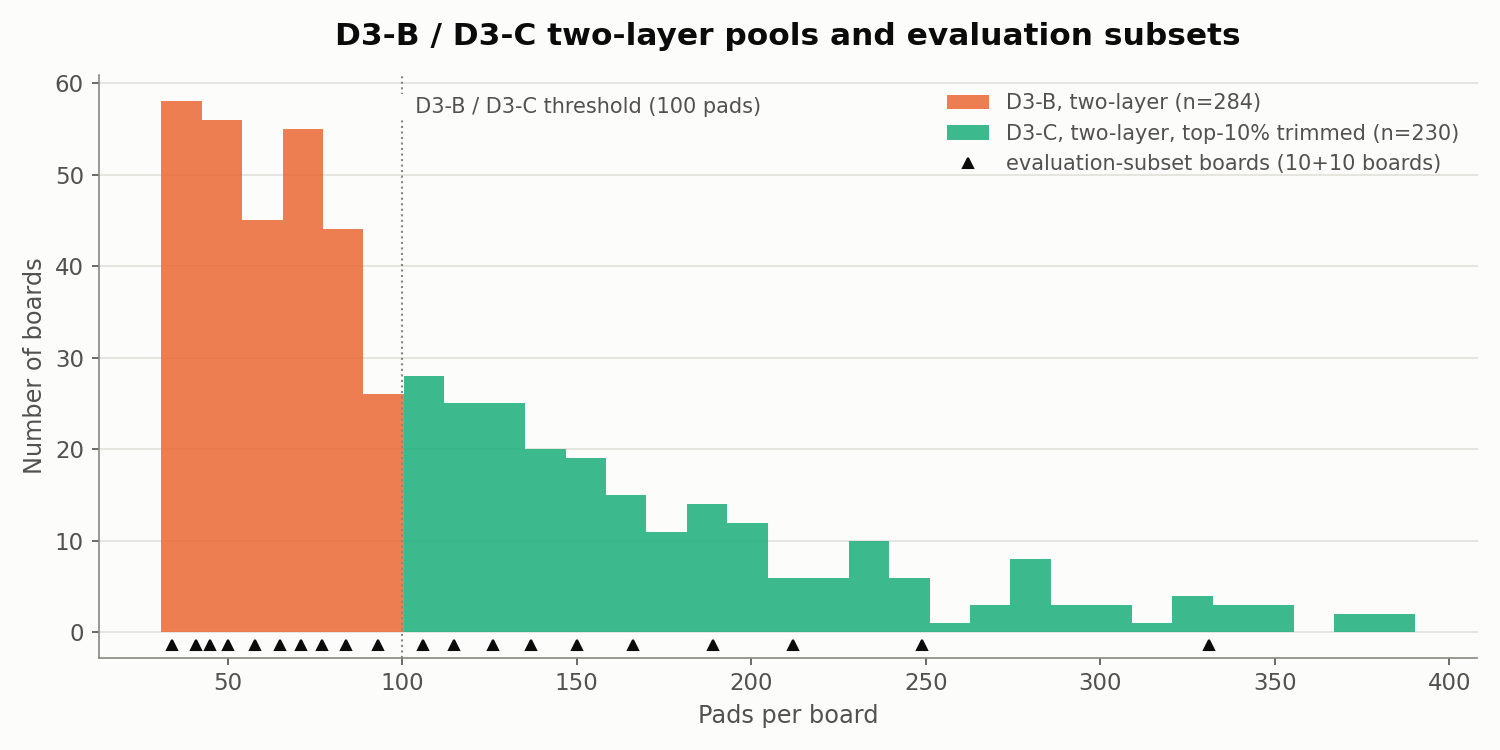}
\caption{Pad-count distribution of the two-layer D3-B and D3-C pools from which the evaluation
subsets are drawn, with the D3-C pool trimmed of its largest $10\%$. The dotted line is the $100$-pad
threshold separating the two pools. Triangles below the axis mark the twenty selected boards, one per
quantile bin, showing that they cover the pad-count range of each pool.}
\label{fig:d3bc-picks}
\Description{Pad-count distribution of the two-layer D3-B and D3-C pools from which the evaluation
subsets are drawn, with the D3-C pool trimmed of its largest 10 percent. The dotted line is the 100-pad
threshold separating the two pools. Triangles below the axis mark the twenty selected boards, one per
quantile bin, showing that they cover the pad-count range of each pool.}
\end{figure}

\subsection{Uses}

The intended use is benchmarking PCB routing methods under a shared engine-grounded protocol, including studies of closed-loop tool use, action masking, synthetic-to-real transfer, and constrained sequential decision-making.
The dataset should not be used as fabrication signoff: it does not replace professional checks for signal integrity, thermal behavior, EMI, or product-specific manufacturing requirements.

\subsection{Distribution}

The synthetic datasets are released under the BSD-3-Clause license together with their generator and evaluation code.
Curated open-source D3 boards retain the licenses of their source repositories and are distributed with attribution metadata where redistribution is permitted.
Each instance is versioned together with the corresponding \ours{} environment release.

\subsection{Maintenance}

Dataset and environment versions are maintained together so that results can be reproduced against the exact board files and evaluator used.
Maintainers will preserve attribution metadata, keep prior versions available when practical, and accept issue reports for corrupted files, license metadata problems, incompatible boards, and evaluator discrepancies.

\section{License, hosting, maintenance}\label{sec:license}
\paragraph{Licensing and versioning.}
The \ours{} repository is released under the BSD-3-Clause license, covering the environment, the synthetic datasets and their generators, the baselines, and the evaluation code.
The routing engine derived from \kicad{} is distributed separately as PCBWorld-Engine under the GPL-3.0 license, and the \ours{} repository pins it as a submodule.
Open-source boards retain the license of their source repository and ship with an attribution table for redistribution.
Every instance carries a dataset version tag in 1:1 correspondence with the environment version, and each release is pinned to \kicad{}~9.0.8 with \texttt{kicad-python}~0.6.0.
The datasheet is provided in Appendix~\ref{sec:datasheet}.

\paragraph{Hosting and maintenance.}
The environment, \oursbench{} datasets, baselines, and evaluation harness are publicly available at \url{https://github.com/LGAI-Research/PCBWorld}.
The authors will maintain the release, tracking \oursbench{} versions against engine versions as described above.

\end{document}